\pdfoutput=1
\documentclass[10pt,twocolumn,letterpaper]{article}
\usepackage[pagenumbers]{cvpr}

\usepackage{cuted} %

\usepackage{currfile} %

\usepackage{caption} %

\usepackage{amsmath,amssymb,amsthm,mathtools,bm}
\usepackage{booktabs,multirow,array,graphicx}
\usepackage{algorithm,algpseudocode}
\usepackage{placeins}%
\usepackage{multicol}%
\usepackage{needspace}%
\newcommand{\vect}[1]{\bm{#1}}
\newcommand{\vst}{\vect{s}}   %
\newcommand{\vz}{\vect{z}}   %
\newcommand{\vx}{\vect{x}}   %
\newcommand{\vg}{\vect{g}}   %
\newcommand{\vb}{\vect{b}}   %
\newcommand{\vu}{\vect{u}}   %
\newcommand{\vv}{\vect{v}}   %
\newcommand{\vk}{\vect{k}}   %
\newcommand{\vw}{\vect{w}}
\newcommand{\vh}{\vect{h}}
\newcommand{\vr}{\vect{r}}
\newcommand{\indic}{\mathbf{1}}
\newcommand{\cA}{\mathcal{A}}
\newcommand{\cB}{\mathcal{B}}
\newcommand{\cD}{\mathcal{D}}
\newcommand{\cE}{\mathcal{E}}
\newcommand{\cH}{\mathcal{H}}
\newcommand{\cP}{\mathcal{P}}
\newcommand{\cQ}{\mathcal{Q}}
\newcommand{\cS}{\mathcal{S}}
\newcommand{\E}{\mathbb{E}}
\newcommand{\R}{\mathbb{R}}

\newcommand{\KLb}{\overline{\mathrm{KL}}}
\newcommand{\pic}{\pi^{c}}
\newcommand{\T}{\top}
\DeclareMathOperator{\silu}{silu}

\DeclareMathOperator*{\argmin}{arg\,min}

\newcommand{\runin}[1]{\par\smallskip\noindent\textbf{#1}}
\newcommand{\name}{BaseRSI}

\definecolor{pinkbg}{HTML}{FBE7EA}%

\definecolor{cvprblue}{rgb}{0.21,0.49,0.74}
\usepackage[pagebackref,breaklinks,colorlinks,allcolors=cvprblue]{hyperref}
\hypersetup{pdftitle={BaseRSI: Recursive Self-Improvement of a Base Vision-Language Model via Training-Free Harness Internalization},pdfauthor={Mingda Zhang, Yuanhao Dong, Qiang Huang, Rui Qian, Yanjin Li, Zhenyu Liu, Xiao Huang, Zihao Yang, Jian Wang, Pinjia He, Xiaoying Tang}}
\def\confName{CVPR}
\def\confYear{2027}
\renewcommand{\dblfloatpagefraction}{0.6}

\title{BaseRSI: Recursive Self-Improvement of a Base Vision-Language Model\\via Training-Free Harness Internalization}
\author{Mingda Zhang$^{1*}$\quad Yuanhao Dong$^{2*}$\quad Qiang Huang$^{3}$\quad Rui Qian$^{3}$\quad Yanjin Li$^{4}$\quad Zhenyu Liu$^{5}$\\
Xiao Huang$^{6}$\quad Zihao Yang$^{1}$\quad Jian Wang$^{1}$\quad Pinjia He$^{1}$\quad Xiaoying Tang$^{1}$\\[\arxivsep]
{\footnotesize\begin{tabular}[t]{@{}c@{}}
$^{1}$The Chinese University of Hong Kong, Shenzhen, China\quad $^{2}$National University of Singapore, Singapore\quad $^{3}$Fudan University, China\\
$^{4}$University of Illinois Urbana-Champaign, USA\quad $^{5}$The University of Melbourne, Australia\quad $^{6}$Shanghai Jiao Tong University, China\\
$^{*}$Equal contribution\quad \texttt{mingdazhang@ieee.org}
\end{tabular}}}
\newlength{\arxivmid}\newlength{\arxivsep}
\makeatletter
\def\@maketitle{\newpage\null
  \begin{center}
    {\Large \bf \@title \par}
    \vspace*{\arxivmid}
    {\large\lineskip .5em
     \begin{tabular}[t]{c}\@author\end{tabular}\par}
  \end{center}}
\makeatother

\begin{document}
\maketitle
\begin{abstract}
Recursive self-improvement (RSI) is moving from the harness of prompts and tools around a base model toward the model itself.
However, existing RSI still faces key challenges, including \emph{harness-bound gains} that stay outside the base model, \emph{black-box supervision} that reads only the model's outputs, and \emph{imprecise edits} whose site and strength are fixed in advance.
To address these challenges, we propose \name{}, a new paradigm of \emph{Recursive Harness Internalization}: the base model improves its harness, then writes what the harness adds into its own weights through training-free edits, round after round on its own responses.
To support this paradigm, we introduce \emph{ReGap}, a mechanism that supervises every transformer block with the high-dimensional activation difference that the harness makes at that block alone, and updates the block's weights along this signal in a low-dimensional subspace.
Building on this mechanism, we further propose \emph{ReKey}, an algorithm that writes this signal by regression into a few key--value neurons of every block, and measures the site and strength of each edit on the model's own responses.
Moreover, the base model fills every role of this recursion with forward passes alone, and its edits, written at different sites, compose in one checkpoint without refitting.
Experimental results on ten visual reasoning benchmarks show that \name{} consistently improves a base vision-language model both in distribution and on held-out benchmarks.
Our code is available~at \url{https://github.com/beita6969/BaseRSI}.
\end{abstract}

\section{Introduction}\label{sec:intro}

\begin{figure}[t]
\centering
\includegraphics[width=\linewidth]{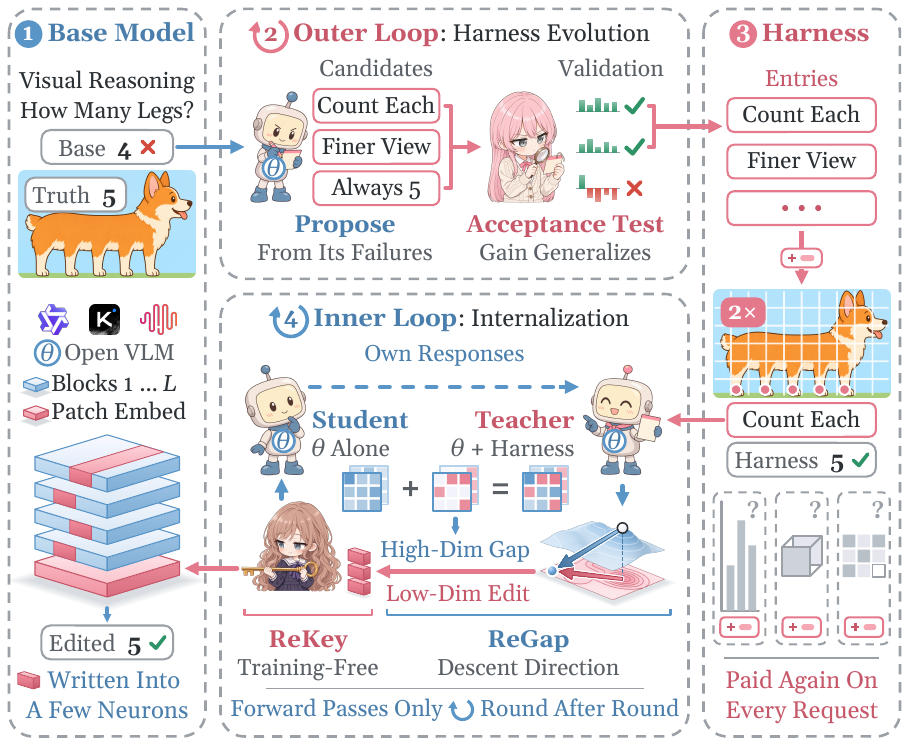}
\caption{\textbf{A harness gain that stays outside the base model is paid again on every request.} \name{} lets the base model write what its harness learns into its own weights, round by round, and answer the original request alone.}
\label{fig:teaser}
\end{figure}
\begin{figure*}[t]
\centering
\includegraphics[width=\textwidth]{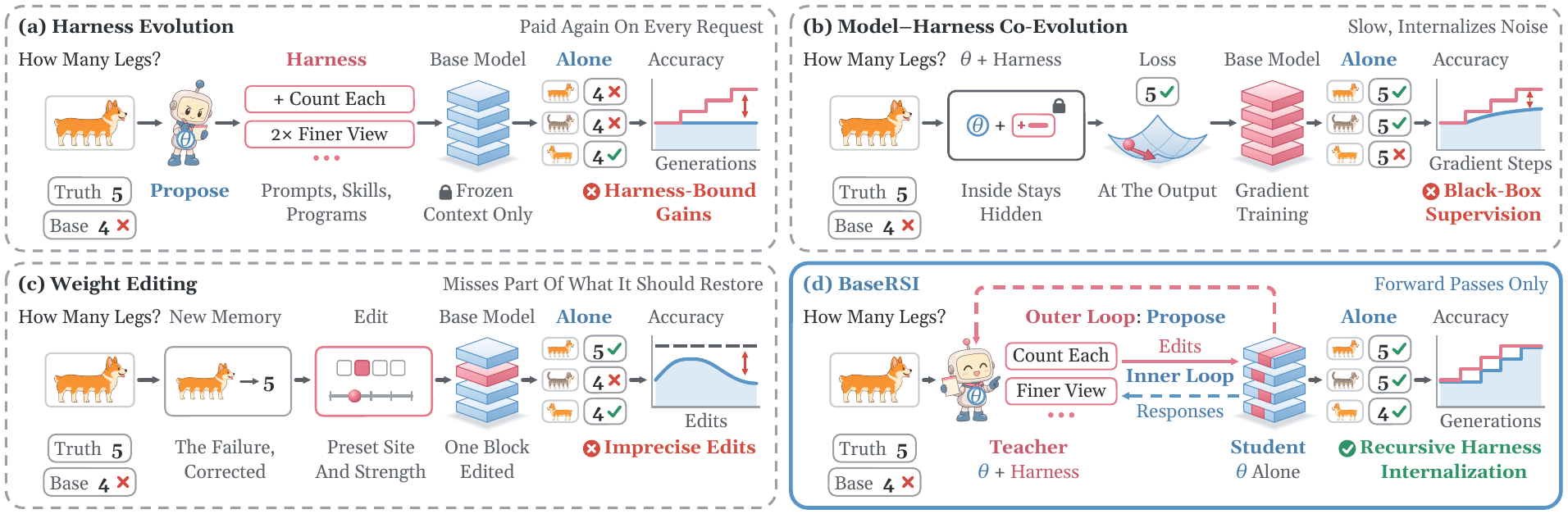}
\caption{\textbf{Three lines of work and \name{}.} (a) Harness evolution leaves \emph{harness-bound gains}. (b) Model--harness co-evolution learns from \emph{black-box supervision}. (c) Weight editing makes \emph{imprecise edits}. (d) \name{} evolves the harness and internalizes each accepted gain into the base model from its own forward passes.}
\label{fig:paradigm}
\end{figure*}

In recent years, a variety of recursive self-improvement (RSI) systems have let foundation models improve the prompts, tools, and context they work with \citep{lee2026meta,rhi,zhou2026hierarchical}, gradually moving from the harness around a frozen model toward improving the base model itself.

In this process, the harness has become a key bridge from a frozen base model to a stronger system: by adding verified procedures and finer views to the request, a vision-language model can answer visual questions with improved accuracy, controllability, and reuse (Figure~\ref{fig:teaser}).
However, in practice, these gains still live in the request rather than in the model \citep{tan2026metarsi,ye2026harness}, making them costly on every request and leaving the base model unable to build on what its harness learns.

Along this path from the harness to the model, three main lines of work have emerged, as shown in Figure~\ref{fig:paradigm}.
First, \emph{harness evolution} rewrites the prompts, tools, and programs around frozen weights \citep{lee2026meta,rhi,xia2026rrsi,xu2026videoharness}, treating the harness as the object of self-improvement.
Second, \emph{model--harness co-evolution} updates the weights together with the harness through fine-tuning, reinforcement learning, or distillation \citep{yu2026co,luo2026harness,gao2026experience}, moving harness gains into the model.
Third, \emph{weight editing} writes new behavior into the weights directly, as memories of facts and mistakes \citep{meng2022locating,meng2022mass,huang2023transformer} or as updates computed directly from forward passes \citep{merzin2026context,yang2026training}.

However, these methods still face three challenges (Figure~\ref{fig:paradigm}a--c).
\textbf{(i) Harness-bound gains.}
Harness evolution improves what surrounds a frozen model \citep{lee2026meta,rhi,xia2026rrsi}; every addition costs context again on each request and is lost when the harness is taken away \citep{tan2026metarsi,ye2026harness,gao2026experience}, so the gains accumulate outside the base model and its own capability stays where it started.
\textbf{(ii) Black-box supervision.}
Co-evolution and harness distillation supervise the weights only at the output of the model, with trajectories, rewards, or output distributions \citep{ye2026harness,su2026coevo,luo2026harness,zhao2026training}; what the harness changes inside the model stays hidden, so the signal travels back through every block by gradient training, which is slow and internalizes noise \citep{gao2026experience}.
\textbf{(iii) Imprecise edits.}
Weight editing sets the site and strength of an edit by a preset rule, in a few layers \citep{meng2022mass} or the last layer \citep{huang2023transformer}; localization does not tell where an edit works best \citep{hase2023does,pan2025precise}, so edits miss part of their target \citep{baghel2025resolving} and harm general abilities \citep{gu2024model}.

To address these challenges, we propose \textbf{\name{}} (Figure~\ref{fig:paradigm}d), a new paradigm of \emph{Recursive Harness Internalization} that runs in two loops: in the outer loop, the base model adds candidate entries to its harness, and an acceptance test keeps those whose gain generalizes.
In the inner loop, the base model with an accepted entry in the request is the teacher, the same model on the original request is the student, and the teacher and the unedited student first answer the development questions.
To support this paradigm, we introduce \emph{ReGap}, a mechanism that replays each response through the teacher and the student, takes the difference that the entry makes at every block alone, the teacher--student gap, as a high-dimensional target inside the model rather than at its output, and turns each gap into a low-dimensional update of that block's weights.
Building on this mechanism, we further propose \emph{ReKey}, an algorithm that first measures on the student's responses which weights the entry should be written into, then writes the gaps by regression into a few key--value neurons of every block, and finally measures the strength of the edit on held-out responses.
The edited student then answers again, and since the edit has changed its responses, the next round repeats these steps on the new ones. After the last round, the edits written at different sites add up in one checkpoint, which answers the original requests alone.
Throughout, the base model fills every role with its own forward passes, and \name{} joins harness evolution, weight editing, and internalization into one recursive loop that makes the base model itself stronger.

We evaluate on five in-distribution and five held-out visual reasoning benchmarks across perception, counting, charts, spatial reasoning, and visual logic.
Results show that \name{} outperforms prompting, visual zoom-in, harness evolution, and fine-tuning on every benchmark and carries over to other base models (RQ1, RQ2), that the edited checkpoint exceeds its own harness in the request while answering original requests on standard inference engines (RQ3), and that supervision inside the model and a measured site and strength each add to the gain (RQ4, RQ5).
These results mark a step toward base models that recursively improve their own weights.

\section{Related Work}\label{sec:related}

\runin{Self-Improving Harnesses and Models.}
Harness evolution searches and rewrites prompts, tools, and programs around frozen weights \citep{lee2026meta,rhi,zhou2026hierarchical,xia2026rrsi,xu2026videoharness,luo2026video}, and co-evolution, harness distillation, and self-edits also update the weights, by fine-tuning, reinforcement learning, or distillation from a teacher that holds the context \citep{ye2026harness,yu2026co,su2026coevo,luo2026harness,xu2026evo,zhao2026training,harnessgoes,jiang2026distill,gao2026experience,zweiger2026self,behrouz2026language}, at times with a scheduler choosing between harness and weights \citep{hebbar2026sia,tan2026metarsi}.
The former keeps its gains in the request and the latter supervises the weights at the output, and both often run an external frontier model as proposer or corrector \citep{lee2026meta,ye2026harness,yu2026co,hebbar2026sia}.
\name{} keeps every role in one base model and writes each accepted entry into its weights from the gap measured inside the model.

\runin{Weight Editing and Parametric Memory.}
Weight editing writes facts, mistakes, and experiences into one or a few layers, the last layer, or adapters \citep{meng2022locating,meng2022mass,huang2023transformer,ren2026scaling,zhao2026rpmem}, and more precise or neighbor-aware edits raise locality \citep{pan2025precise,baghel2025resolving}, while edits still miss related knowledge and can harm general abilities \citep{zhang2025uncovering,cohen2024evaluating,gu2024model}.
A context or behavior is also condensed into weights, as steering directions \citep{arditi2024refusal}, implicit or least-squares block updates \citep{dherin2025learning,goldwaser2025equivalence,mazzawi2025transmuting}, edits from contrastive forward passes \citep{merzin2026context,ebrahimpour2026resisting}, or dormant neurons \citep{zeng2026lazare}, and specialists of one base are fused from forward passes \citep{yang2026training}; concurrently, a frontier model rewrites the weights of a small model \citep{takehi2026sparks}.
In contrast, \name{} fits each edit to the gap of every block alone, measures its site and strength, and refits it round by round on the model's own responses.

\section{Preliminaries}\label{sec:prelim}

\noindent\textbf{Definition 1: Base model and harness.}
A base model $f_\theta$ is a vision-language model with weights $\theta$ and $L$ residual blocks, each applying an attention layer (linear or full attention) and a SwiGLU MLP \citep{shazeer2020glu} to a residual stream of width $H$.
A harness \citep{rhi} $\cH=\{c_1,\dots,c_m\}$ wraps the model with entries $c=(\kappa_c,a_c)$, where the scope $\kappa_c(q)\in\{0,1\}$, answered by the base model as a router, tells whether the entry applies to a request $q$, and the action $a_c$ rewrites the request by adding a procedure to the prompt or a finer view of the image; $c(q)$ is the rewritten request when the entry applies and $q$ otherwise.
The entries are the parameters of the harness, as the weights $\theta$ are those of the model.

\noindent\textbf{Definition 2: Response trajectory.}
A model with a harness responds to a request $q$ with a token sequence $y_{1:T}$, its reasoning followed by the final answer, in which every token is drawn on the request that the harness has rewritten:
\begin{equation}\label{eq:traj}
\begin{gathered}
y_t\sim p_\theta\big(\cdot\mid\cH(q),y_{<t}\big),\quad \cH(q)=(c_m\circ\cdots\circ c_1)(q),\\
\vst_l=F_l\big(\vst_{l-1};\vz_{l-1}\big),\quad l=1,\dots,L,
\end{gathered}
\end{equation}
where $p_\theta$ is the next-token distribution, $\vz_l$ and $\vst_l$ are the activations after block $l$ at the prompt and response positions (the tokens of $y$), the prompt states and the response states, and $F_l$ is the output of block $l$ at the response positions.
Blocks are causal, so the prompt states depend on the request alone, and an empty harness leaves it unchanged.

\noindent\textbf{Problem Statement.}
Given base weights $\theta_0$, an empty harness $\cH_0$, a pool $\cQ$ of development requests, and a family $\cE$ of edits that rewrite a bounded number of MLP neurons per block or the patch kernel, we seek a base model that answers original requests better by recursive self-improvement:
\begin{equation}\label{eq:problem}
\max_{e_0,\dots,e_{N-1}\in\cE}\ \E_{(q,y^\star)\sim\mathcal{T}}\,\indic\big[f_{\theta_N}(q)=y^\star\big],
\end{equation}
where $\mathcal{T}$ are test tasks in distribution and held out, $f_{\theta_N}(q)$ is the final answer of the base model to the original request, and $\theta_{n+1}=\theta_n\oplus e_n$ are the weights after edit $e_n$.
Every generation may evolve a harness $\cH_n$ to find what to internalize, and only the weights enter Eq.~\eqref{eq:problem}.
Equation~\eqref{eq:problem} is the deployed target; Section~\ref{sec:method} shows how an outer loop over the harness and an inner loop over the weights approach it.

\section{Method}\label{sec:method}

\begin{figure*}[t]
\centering
\includegraphics[width=\textwidth]{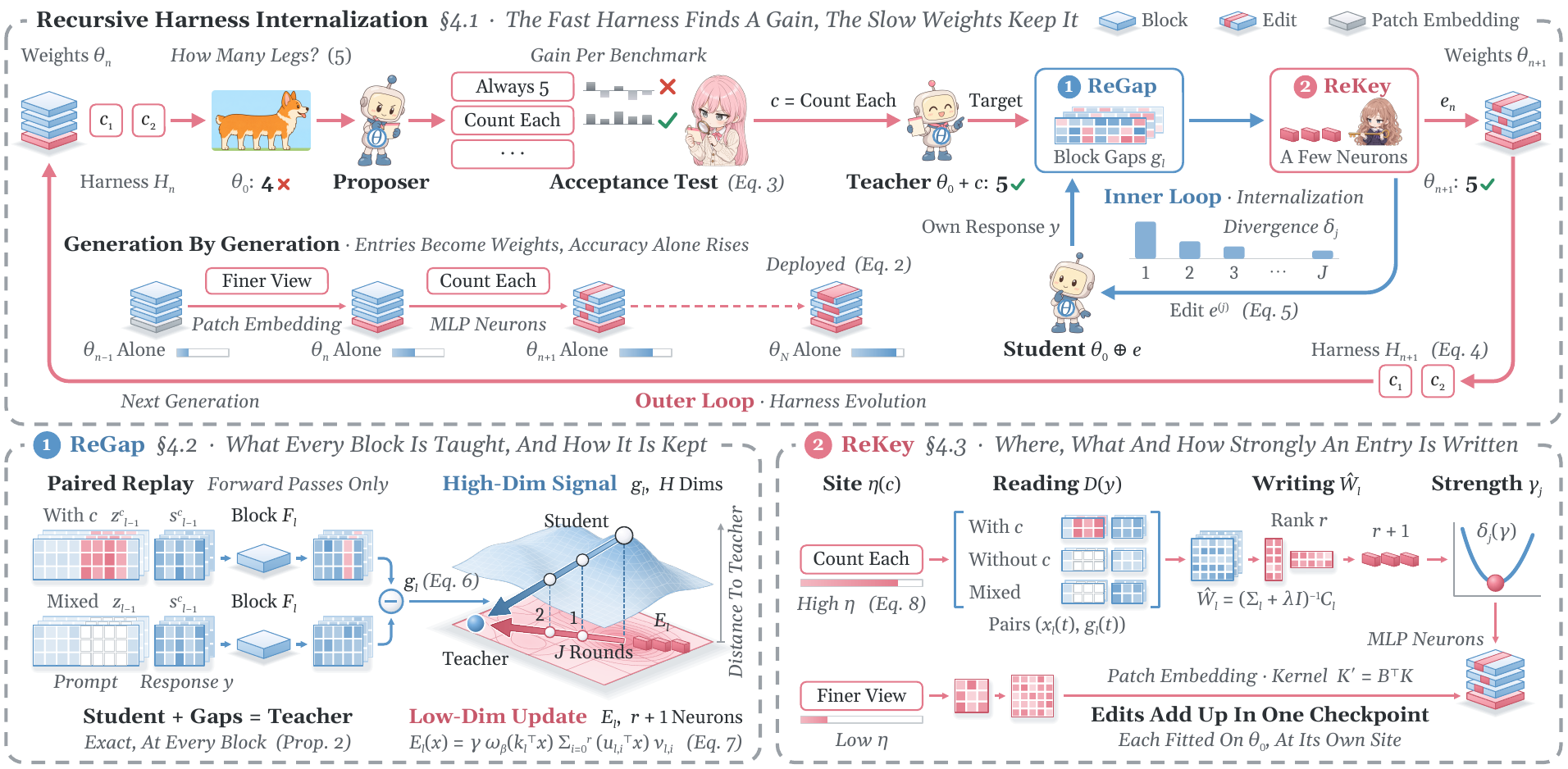}
\caption{\textbf{Overview of \name{}.} \emph{Top:} Recursive Harness Internalization runs an outer loop that evolves the harness and an inner loop in which the model with its harness teaches the model alone, round after round. \emph{Left:} ReGap supervises every block with the teacher--student gap and keeps it as a low-dimensional update. \emph{Right:} ReKey writes the gap into key--value neurons at a measured site and strength.}
\label{fig:framework}
\end{figure*}

As illustrated in Figure~\ref{fig:framework}, this section introduces the \name{} framework from the system down to the weights, including Recursive Harness Internalization (Section~\ref{sec:rhc}), the paradigm in which the base model teaches itself, ReGap (Section~\ref{sec:regap}), the mechanism that states what is taught in every block, and ReKey (Section~\ref{sec:rekey}), the algorithm that writes it into the weights at a measured site and strength.

\subsection{Recursive Harness Internalization}\label{sec:rhc}

The paradigm lets the base model keep what its harness learns.
\name{} treats the harness as a fast system that records new experience and the weights as a slow system that consolidates it through replay \citep{mcclelland1995there,kumaran2016learning}, and joins them in one cycle (Figure~\ref{fig:framework}, top): an outer loop finds a gain in the harness, an inner loop moves it into the weights, and the next cycle looks for a further gain.

\runin{Outer loop: harness evolution.}
The fast system has to supply gains that are worth keeping, so the base model $\theta_0$ proposes entries from its own failures and an acceptance test keeps those whose gain generalizes:
\begin{equation}\label{eq:judge}
\cA_n=\big\{c:\ p_{\mathrm{gen}}(c)=\max_{b\in\cB}\,p\big(\{d_{b'}(c)\}_{b'\neq b}\big)\le\alpha_c\big\},
\end{equation}
where $d_b(c)$ is the paired accuracy gain of adding $c$ on the validation questions of benchmark $b$ at an equal token budget, $p(\cdot)$ is the one-sided sign-flip $p$-value of the macro mean of its arguments, and $\alpha_c$ is the Holm threshold over the candidates of generation $n$ \citep{holm1979simple}.
A general entry is thus kept only when its gain is significant with any one benchmark left out, and an entry for one benchmark only when its gain there is significant and no other benchmark is significantly worse (Appendix~\ref{app:judge}).
Each accepted entry is tested on questions disjoint from those it was written from, so it hands the inner loop a verified target.

\runin{Inner loop: internalization.}
A verified entry $c\in\cA_n$ still acts through the request, so the inner loop moves it into the weights.
The base model with $c$ in the request is the teacher $\pic_{\theta_0}=p_{\theta_0}(\cdot\mid c(q),y_{<t})$, the same model on the original request is the student $\pi_{\theta_0}=p_{\theta_0}(\cdot\mid q,y_{<t})$, and the edit makes the student match its own teacher on the responses that it generates by itself \citep{vapnik2009new,lopez2015unifying}:
\begin{equation}\label{eq:consol}
\begin{gathered}
e_n\in\argmin_{e\in\cE}\ \E_{q\sim\cQ}\,\E_{y\sim\pi_{e}(\cdot\mid q)}\,\KLb\big(\pic_{\theta_0}\,\big\|\,\pi_{e}\big),\\
\theta_{n+1}=\theta_n\oplus e_n,\qquad
\cH_{n+1}=(\cH_n\cup\cA_n)\setminus\{c\},
\end{gathered}
\end{equation}
where $\pi_e=\pi_{\theta_0\oplus e}$ is the student that carries edit $e$, teacher and student both run under the entries internalized before $c$, and $\KLb$ averages the divergence between the two next-token distributions over the response positions of $(q,y)$.
The expectation runs over responses of the edited student, and since $c(q)=q$ outside the scope of the entry, the same objective there asks the edit to keep the base model as it is.
Every edit is fitted on the base weights and takes its own neurons or the patch kernel, so the edits add up in one checkpoint without refitting, and each can be removed alone.
The internalized entries leave the request, and the checkpoint $\theta_N$ with every edit is deployed alone (Eq.~\eqref{eq:problem}).

\runin{Rounds on its own responses.}
The target of Eq.~\eqref{eq:consol} is a fixed point, since an edit changes the responses on which it is fitted.
An edit fitted once on the teacher's responses therefore loses part of its effect once the edited student generates its own, the covariate shift of imitation learning \citep{ross2010efficient,li2026revisiting}. So the inner loop runs in rounds on the student's own responses \citep{ross2011reduction}: the edited student responds, the teacher labels that response, and the edit is refitted on the pooled responses:
\begin{equation}\label{eq:aggregate}
\begin{gathered}
\cD^{(j)}=\cD^{(j-1)}\cup\cD(y),\qquad y\sim\pi_{e^{(j-1)}}(\cdot\mid q),\\
e^{(j)}=\mathrm{ReKey}\big(\cD^{(j)}\big),
\end{gathered}
\end{equation}
where $\cD(y)$ are the labels of a response $y$, the gaps of Section~\ref{sec:regap}, and $\mathrm{ReKey}(\cdot)$ is the edit that Section~\ref{sec:rekey} fits on a set of labels.
The first two rounds replay the responses of the teacher and of the unedited student, and the later rounds those of the edited student (Appendix~\ref{app:algo}).
Every round computes its labels with the unedited base model, so the labels of all rounds describe one target and can be pooled.
The refit replaces the previous edit and minimizes the summed loss of all rounds so far \citep{ross2011reduction}, and the last one is $e_n$ of Eq.~\eqref{eq:consol}.

\noindent\textbf{Proposition 1.} \emph{Some round's fit leaves a gap on its own student's responses no larger than the smallest regularized gap any single linear readout leaves on the responses of all rounds, plus a margin that shrinks as rounds accumulate.}
\emph{Proof.} Appendix~\ref{app:dagger}, via Lemmas~\ref{lem:ftl} and~\ref{lem:own} \citep{ross2011reduction}.

\subsection{ReGap}\label{sec:regap}

ReGap is the mechanism of the inner loop: it states what the teacher teaches in every block, a high-dimensional signal, and in what form the student keeps it, a low-dimensional update of the block's weights (Figure~\ref{fig:framework}, bottom left).

\runin{High-dimensional signal.}
The whole effect of an entry on a response is a set of additions, one per block.
We define the gap of block $l$ by running the block on the teacher's states with and without the entry in the request, and treat it as an implicit update of the block's MLP weights \citep{dherin2025learning,goldwaser2025equivalence,mazzawi2025transmuting}:
\begin{equation}\label{eq:gap}
\vg_l=F_l\big(\vst^{c}_{l-1};\,\vz^{c}_{l-1}\big)-F_l\big(\vst^{c}_{l-1};\,\vz_{l-1}\big),
\end{equation}
where $F_l$ is the block map of Eq.~\eqref{eq:traj} and the superscript $c$ marks the states of the teacher.
Both terms take the same response states, so $\vg_l$ is what the entry adds at block $l$ alone, its direct effect \citep{pearl2022direct}, and gaps of different blocks share no part.
For a procedure entry, the teacher's attention layers read procedure tokens that the student's request lacks, and $\vg_l$ is the part of the block's output that they account for.

\runin{Descent direction.}
The gap takes the place that the gradient has in training.
It is the exact descent direction of the distance between block $l$ and its teacher \citep{mason1999boosting,friedman2001greedy}: a unit step along $\vg_l$ closes that distance, and it comes from forward passes, with no gradient propagated from the output.
Every term of the gap is a forward pass of the base model itself, which also proposes and routes the entries, so one model fills every role of the cycle and edits itself.

\runin{Low-dimensional update.}
The student has to produce $\vg_l$ from its own state, so the edit is a function of the input $\vx$ of the MLP of block $l$, kept in a few directions.
The update is a small set of key--value neurons, each storing one key--value pair in the MLP weights \citep{kohonen1972correlation,anderson1972simple,geva2021transformer,bau2020rewriting,meng2022locating}:
\begin{equation}\label{eq:unit}
E_l(\vx)=\gamma\,\omega_\beta\big(\vk_l^{\T}\vx\big)\sum\nolimits_{i=0}^{r}\big(\vu_{l,i}^{\T}\vx\big)\vv_{l,i},
\end{equation}
where each term reads the input along a readout $\vu_{l,i}$ and writes a value $\vv_{l,i}$, the key $\vk_l$ carries the entry's scope through $\omega_\beta(z)=z\,\sigma(\beta z)$, $\gamma$ is the step, and $r$ is far below the width $H$.
Each term is one SwiGLU neuron, which passes its readout scaled by the key (Lemma~\ref{lem:neuron}), so an edit rewrites $r{+}1$ neurons per block, every other weight stays bit-identical, and the MLP keeps its architecture.

\noindent\textbf{Proposition 2.} \emph{Adding every block's gap to the student's activations makes its output identical to the teacher's.}
\emph{Proof.} Appendix~\ref{app:gap}, via Lemma~\ref{lem:hybrid} and induction.

\begin{table*}[t]
\centering
\footnotesize
\setlength{\tabcolsep}{0pt}
\setlength{\aboverulesep}{0.25ex}
\setlength{\belowrulesep}{0.4ex}
\newcommand{\hd}[2]{\shortstack{#1\\#2}}
\begin{tabular*}{\textwidth}{@{}l@{\extracolsep{\fill}}cccccccccccc@{}}
\toprule
 & \multicolumn{6}{c}{In-distribution (ID)} & \multicolumn{6}{c}{Held-out (OOD)} \\
\cmidrule{2-7}\cmidrule{8-13}
Method & \hd{Baby}{Vision} & BLINK & CharXiv & \hd{Visu}{Logic} & \hd{VLMs Are}{Biased} & Avg. & \hd{ChartQA}{Pro} & \hd{Hallusion}{Bench} & \hd{Perception}{Bench} & \hd{Spatial}{Viz} & \hd{Visual}{Puzzles} & Avg. \\
\midrule
\multicolumn{13}{@{}l}{\textit{Closed-source models}}\\
Claude Haiku 5.5 & 37.0\smash{$_{\pm2.7}$} & 72.9\smash{$_{\pm2.0}$} & 78.6\smash{$_{\pm3.0}$} & 35.9\smash{$_{\pm2.7}$} & 62.8\smash{$_{\pm0.5}$} & 57.4\smash{$_{\pm0.3}$} & 68.8\smash{$_{\pm0.8}$} & 79.2\smash{$_{\pm1.2}$} & 55.5\smash{$_{\pm2.1}$} & 57.8\smash{$_{\pm0.8}$} & 66.9\smash{$_{\pm2.7}$} & 65.6\smash{$_{\pm0.5}$}\\
GPT-6 Luna & 35.7\smash{$_{\pm0.9}$} & 76.6\smash{$_{\pm1.4}$} & 74.7\smash{$_{\pm0.5}$} & 34.4\smash{$_{\pm1.6}$} & 50.8\smash{$_{\pm1.4}$} & 54.4\smash{$_{\pm0.5}$} & 63.3\smash{$_{\pm0.8}$} & 74.0\smash{$_{\pm2.0}$} & 33.1\smash{$_{\pm4.0}$} & 49.5\smash{$_{\pm3.5}$} & 67.2\smash{$_{\pm3.4}$} & 57.4\smash{$_{\pm0.9}$}\\
\midrule
Qwen3.8-27B (base) & 39.1\smash{$_{\pm1.6}$} & 74.0\smash{$_{\pm1.2}$} & 79.9\smash{$_{\pm3.3}$} & 34.4\smash{$_{\pm2.7}$} & 74.7\smash{$_{\pm1.6}$} & 60.4\smash{$_{\pm1.3}$} & 68.0\smash{$_{\pm0.8}$} & 74.2\smash{$_{\pm0.8}$} & 50.3\smash{$_{\pm2.7}$} & 62.8\smash{$_{\pm0.9}$} & 67.4\smash{$_{\pm1.8}$} & 64.5\smash{$_{\pm0.5}$}\\
\midrule
\multicolumn{13}{@{}l}{\textit{Prompting and sampling}}\\
CoT~\citep{wei2022chain} & 45.8\smash{$_{\pm2.3}$} & 72.1\smash{$_{\pm2.0}$} & 81.0\smash{$_{\pm3.7}$} & 38.0\smash{$_{\pm3.0}$} & 75.3\smash{$_{\pm2.5}$} & 62.4\smash{$_{\pm0.9}$} & 70.1\smash{$_{\pm0.5}$} & 76.8\smash{$_{\pm1.2}$} & 55.2\smash{$_{\pm0.9}$} & 65.1\smash{$_{\pm3.0}$} & 68.5\smash{$_{\pm3.7}$} & 67.1\smash{$_{\pm0.8}$}\\
4-shot examples & 38.5\smash{$_{\pm3.7}$} & 69.5\smash{$_{\pm2.8}$} & 76.0\smash{$_{\pm1.2}$} & 31.8\smash{$_{\pm0.9}$} & 68.0\smash{$_{\pm1.4}$} & 56.8\smash{$_{\pm1.5}$} & 65.6\smash{$_{\pm1.6}$} & 79.2\smash{$_{\pm0.9}$} & 52.1\smash{$_{\pm1.6}$} & 62.2\smash{$_{\pm3.3}$} & 69.8\smash{$_{\pm1.2}$} & 65.8\smash{$_{\pm1.4}$}\\
Self-consistency~\citep{wang2022self} & 42.2\smash{$_{\pm2.1}$} & 73.7\smash{$_{\pm2.0}$} & 84.4\smash{$_{\pm1.4}$} & 35.9\smash{$_{\pm2.1}$} & 74.7\smash{$_{\pm3.5}$} & 62.2\smash{$_{\pm0.5}$} & 68.5\smash{$_{\pm1.2}$} & 75.8\smash{$_{\pm1.4}$} & 52.3\smash{$_{\pm1.4}$} & 65.1\smash{$_{\pm1.2}$} & 66.4\smash{$_{\pm2.7}$} & 65.6\smash{$_{\pm0.5}$}\\
Self-Refine~\citep{madaan2023self} & 39.8\smash{$_{\pm6.2}$} & 72.9\smash{$_{\pm2.0}$} & 82.3\smash{$_{\pm1.2}$} & 42.4\smash{$_{\pm5.2}$} & 73.4\smash{$_{\pm3.1}$} & 62.2\smash{$_{\pm3.2}$} & 68.2\smash{$_{\pm1.2}$} & 80.5\smash{$_{\pm3.9}$} & 57.3\smash{$_{\pm3.5}$} & 65.1\smash{$_{\pm1.6}$} & 71.1\smash{$_{\pm2.1}$} & 68.4\smash{$_{\pm1.6}$}\\
\midrule
\multicolumn{13}{@{}l}{\textit{Visual zoom-in}}\\
ZoomEye~\citep{shen2025zoomeye} & 31.3\smash{$_{\pm2.1}$} & 74.2\smash{$_{\pm3.6}$} & 76.8\smash{$_{\pm3.0}$} & 37.5\smash{$_{\pm0.8}$} & 60.4\smash{$_{\pm2.0}$} & 56.0\smash{$_{\pm1.7}$} & 65.9\smash{$_{\pm0.9}$} & 74.2\smash{$_{\pm1.6}$} & 48.7\smash{$_{\pm2.3}$} & 62.5\smash{$_{\pm2.1}$} & 68.5\smash{$_{\pm2.4}$} & 64.0\smash{$_{\pm0.4}$}\\
Two-stage zoom-in & 47.9\smash{$_{\pm3.9}$} & 73.4\smash{$_{\pm4.7}$} & 84.9\smash{$_{\pm2.7}$} & 38.5\smash{$_{\pm5.5}$} & 72.4\smash{$_{\pm4.8}$} & 63.4\smash{$_{\pm0.4}$} & 69.8\smash{$_{\pm1.2}$} & 79.9\smash{$_{\pm0.9}$} & 57.8\smash{$_{\pm2.3}$} & 64.6\smash{$_{\pm0.9}$} & 70.1\smash{$_{\pm0.5}$} & 68.4\smash{$_{\pm0.5}$}\\
\midrule
\multicolumn{13}{@{}l}{\textit{Harness evolution}}\\
GEPA~\citep{agrawal2026gepa} & 41.9\smash{$_{\pm3.0}$} & 73.7\smash{$_{\pm4.0}$} & 81.8\smash{$_{\pm1.6}$} & 40.4\smash{$_{\pm4.0}$} & 74.7\smash{$_{\pm1.6}$} & 62.5\smash{$_{\pm0.3}$} & 70.8\smash{$_{\pm1.2}$} & 79.4\smash{$_{\pm2.0}$} & 52.6\smash{$_{\pm1.2}$} & 62.0\smash{$_{\pm2.7}$} & 68.8\smash{$_{\pm1.6}$} & 66.7\smash{$_{\pm0.2}$}\\
ACE~\citep{zhang2026agentic} & 35.7\smash{$_{\pm0.5}$} & 70.8\smash{$_{\pm1.2}$} & 81.3\smash{$_{\pm1.4}$} & 36.5\smash{$_{\pm1.8}$} & 72.1\smash{$_{\pm3.2}$} & 59.3\smash{$_{\pm1.2}$} & 68.0\smash{$_{\pm1.4}$} & 79.2\smash{$_{\pm4.3}$} & 50.3\smash{$_{\pm2.0}$} & 61.2\smash{$_{\pm2.5}$} & 66.4\smash{$_{\pm5.4}$} & 65.0\smash{$_{\pm0.9}$}\\
Meta-Harness~\citep{lee2026meta} & 45.1\smash{$_{\pm3.7}$} & 73.4\smash{$_{\pm0.8}$} & 84.1\smash{$_{\pm1.6}$} & 40.9\smash{$_{\pm3.5}$} & 71.9\smash{$_{\pm0.8}$} & 63.1\smash{$_{\pm1.1}$} & 70.1\smash{$_{\pm2.3}$} & 77.3\smash{$_{\pm1.6}$} & 54.9\smash{$_{\pm3.9}$} & 65.6\smash{$_{\pm2.8}$} & 70.3\smash{$_{\pm1.6}$} & 67.7\smash{$_{\pm1.4}$}\\
SkillOpt~\citep{yang2026skillopt} & 41.7\smash{$_{\pm3.3}$} & 71.6\smash{$_{\pm1.2}$} & 81.8\smash{$_{\pm0.9}$} & 37.8\smash{$_{\pm1.2}$} & 75.8\smash{$_{\pm2.8}$} & 61.7\smash{$_{\pm0.2}$} & 70.8\smash{$_{\pm0.5}$} & 76.8\smash{$_{\pm1.6}$} & 54.2\smash{$_{\pm3.0}$} & 65.9\smash{$_{\pm3.9}$} & 69.0\smash{$_{\pm3.5}$} & 67.3\smash{$_{\pm1.2}$}\\
VideoHarness-RSI~\citep{xu2026videoharness} & 40.9\smash{$_{\pm0.5}$} & 75.0\smash{$_{\pm1.6}$} & 80.7\smash{$_{\pm0.5}$} & 37.8\smash{$_{\pm5.2}$} & 70.1\smash{$_{\pm3.5}$} & 60.9\smash{$_{\pm1.3}$} & 69.0\smash{$_{\pm0.5}$} & 76.0\smash{$_{\pm1.6}$} & 52.1\smash{$_{\pm3.2}$} & 64.1\smash{$_{\pm1.4}$} & 70.1\smash{$_{\pm2.3}$} & 66.3\smash{$_{\pm0.4}$}\\
RRSI~\citep{xia2026rrsi} & 42.4\smash{$_{\pm3.0}$} & 75.5\smash{$_{\pm0.5}$} & 81.3\smash{$_{\pm2.1}$} & 42.7\smash{$_{\pm2.5}$} & 83.6\smash{$_{\pm3.6}$} & 65.1\smash{$_{\pm1.3}$} & 69.8\smash{$_{\pm0.9}$} & 79.9\smash{$_{\pm1.6}$} & 54.9\smash{$_{\pm5.3}$} & 65.4\smash{$_{\pm4.4}$} & 71.4\smash{$_{\pm4.0}$} & 68.3\smash{$_{\pm2.0}$}\\
\midrule
\multicolumn{13}{@{}l}{\textit{Fine-tuning}}\\
SFT & 39.3\smash{$_{\pm1.6}$} & 72.4\smash{$_{\pm1.6}$} & 79.7\smash{$_{\pm2.3}$} & 37.0\smash{$_{\pm1.2}$} & 76.3\smash{$_{\pm3.0}$} & 60.9\smash{$_{\pm0.7}$} & 68.8\smash{$_{\pm2.7}$} & 75.3\smash{$_{\pm2.3}$} & 57.0\smash{$_{\pm3.6}$} & 63.3\smash{$_{\pm0.8}$} & 61.2\smash{$_{\pm2.0}$} & 65.1\smash{$_{\pm1.0}$}\\
GRPO~\citep{shao2024deepseekmath} & 45.6\smash{$_{\pm3.5}$} & 68.8\smash{$_{\pm1.4}$} & 79.2\smash{$_{\pm1.2}$} & 35.7\smash{$_{\pm2.5}$} & 76.8\smash{$_{\pm1.6}$} & 61.2\smash{$_{\pm0.8}$} & 67.7\smash{$_{\pm1.2}$} & 75.5\smash{$_{\pm2.7}$} & 52.6\smash{$_{\pm2.7}$} & 54.2\smash{$_{\pm2.7}$} & 69.3\smash{$_{\pm3.9}$} & 63.9\smash{$_{\pm0.7}$}\\
PixelProof~\citep{pixelproof2026} & 39.6\smash{$_{\pm3.2}$} & 75.0\smash{$_{\pm2.7}$} & 81.8\smash{$_{\pm2.0}$} & 39.3\smash{$_{\pm2.0}$} & 72.7\smash{$_{\pm2.1}$} & 61.7\smash{$_{\pm0.2}$} & 68.2\smash{$_{\pm2.4}$} & 75.3\smash{$_{\pm1.8}$} & 54.7\smash{$_{\pm2.1}$} & 60.9\smash{$_{\pm0.8}$} & 67.4\smash{$_{\pm2.7}$} & 65.3\smash{$_{\pm0.8}$}\\
LBS-27B~\citep{yao2026learning} & 49.0\smash{$_{\pm2.0}$} & 71.9\smash{$_{\pm1.4}$} & 80.5\smash{$_{\pm5.4}$} & 41.9\smash{$_{\pm3.5}$} & 70.1\smash{$_{\pm4.8}$} & 62.7\smash{$_{\pm2.6}$} & 67.7\smash{$_{\pm3.5}$} & 75.8\smash{$_{\pm1.4}$} & 52.3\smash{$_{\pm4.1}$} & 61.5\smash{$_{\pm7.5}$} & 64.1\smash{$_{\pm2.7}$} & 64.3\smash{$_{\pm0.2}$}\\
\midrule
\multicolumn{13}{@{}l}{\textit{Harness internalization}}\\
\smash{\rlap{\color{pinkbg}\rule[-12.5pt]{\textwidth}{19.1pt}}}\textbf{\name{} (ours)} & \textbf{60.7}\smash{$_{\pm2.4}$} & \textbf{76.6}\smash{$_{\pm1.4}$} & \textbf{87.5}\smash{$_{\pm1.4}$} & \textbf{44.8}\smash{$_{\pm2.0}$} & \textbf{91.1}\smash{$_{\pm1.2}$} & \textbf{72.1}\smash{$_{\pm0.4}$} & \textbf{74.0}\smash{$_{\pm1.6}$} & \textbf{81.3}\smash{$_{\pm2.1}$} & \textbf{59.1}\smash{$_{\pm2.7}$} & \textbf{66.1}\smash{$_{\pm2.5}$} & \textbf{72.9}\smash{$_{\pm1.6}$} & \textbf{70.7}\smash{$_{\pm0.4}$}\\
$\Delta$ over base model & $+$21.6 & $+$2.6 & $+$7.6 & $+$10.4 & $+$16.4 & $+$11.7 & $+$6.0 & $+$7.0 & $+$8.9 & $+$3.4 & $+$5.5 & $+$6.1\\
\bottomrule
\end{tabular*}
\caption{Main results on five in-distribution (ID) and five held-out (OOD) benchmarks: accuracy (\%), mean$_{\pm\mathrm{std}}$ over three runs. All methods but the two closed-source reference models use Qwen3.8-27B with the same reasoning effort, token budget, and evaluator; \name{} (shaded) is the edited base model answering the original requests alone. $\Delta$: gain over the base model, from unrounded means.}
\label{tab:main}
\vspace{-0.5em}
\end{table*}

\subsection{ReKey}\label{sec:rekey}

ReKey is the algorithm that computes the update of Eq.~\eqref{eq:unit} from forward passes alone: it first measures the site of an edit, then writes the gap by regression into a few key--value neurons of every block, and finally measures its strength, all on the model's own responses (Figure~\ref{fig:framework}, bottom right).

\runin{Site.}
Before any fit, ReKey measures the site of an edit, the weights that an entry is written into.
It replays the student's own responses with the outputs of its linear-attention layers set to those of the teacher and measures the fraction of the divergence from the teacher that this patch removes \citep{pearl2022direct}:
\begin{equation}\label{eq:share}
\eta(c)=1-\KLb\big(\pic_\theta\,\big\|\,\pi^{\mathrm{lin}\leftarrow c}_\theta\big)\,\big/\,\KLb\big(\pic_\theta\,\big\|\,\pi_\theta\big),
\end{equation}
where $\pi^{\mathrm{lin}\leftarrow c}_\theta$ is the patched student.
Linear attention carries the request as a running state, which converts exactly into bias terms \citep{chen2024exact}, whereas full attention retrieves single image tokens. A high fraction therefore sends the entry to the MLP neurons, which read the gap from the activations, and a low fraction sends it to the patch embedding, since such an entry changes what the model sees.
For such a view entry and an up-sampling $B$ of the image, resampling the patch kernel $K$ \citep{beyer2023flexivit} to $K'=B^{\T}K$ makes the student embed every patch of the original image as the teacher embeds it after up-sampling (Lemma~\ref{lem:adjoint}, Appendix~\ref{app:interface}).

\runin{Reading.}
ReKey then reads the gap of Eq.~\eqref{eq:gap} on a response $y$ by teacher-forced forward passes of the base model over its tokens, which label every block:
\begin{equation}\label{eq:pairs}
\cD(y)=\big\{\big(\vx_l(t),\vg_l(t)\big)\big\}_{l,\;t\in\cS(y)},
\end{equation}
where $\vx_l(t)$ is the input of the MLP of block $l$ at response position $t$ in the second term of Eq.~\eqref{eq:gap} and $\cS(y)$ are the sampled response positions.
The pairs are exact by Proposition~2 and supervise every block directly.

\runin{Writing.}
For every block, the fit is a linear readout of the MLP input by ridge regression on the pairs:
\begin{equation}\label{eq:ridge}
\hat W_l=\argmin_{W}\ \E_{\cD_l}\big\|\vg-\vb_l-W^{\T}P_l\vx\big\|^2+\lambda\|W\|_F^2,
\end{equation}
where $\cD_l$ are the pairs of block $l$ over the replayed responses, $\vb_l$ is the mean gap, $P_l$ removes the mean direction of the inputs (Lemma~\ref{lem:linear}), and $\lambda$ is the ridge weight.
The fit is then reduced to the $r$ directions that carry most of its predictions over all responses \citep{anderson1951estimating,izenman1975reduced}, so that it occupies only the few neurons a block can spare (Appendix~\ref{app:rekey}).
Each direction fills one key--value neuron of Eq.~\eqref{eq:unit} with a readout $\vu_{l,i}$ and a value $\vv_{l,i}$, and one more neuron carries the mean gap $\vb_l$.
Their key $\vk_l$ is fitted by ridge regression to tell the response positions where the gap is present from all other positions (Appendix~\ref{app:algo}); as a function of the block's input, it carries the scope of an entry to unseen requests.

\runin{Strength.}
Finally, the strength of an edit, its step $\gamma$, is measured by a line search on held-out student responses \citep{friedman2001greedy}:
\begin{equation}\label{eq:stop}
\begin{gathered}
\gamma_j=\argmin_{\gamma\in\Gamma}\ \delta_j(\gamma),\qquad
\delta_j(\gamma)=\KLb\big(\pic_\theta\,\big\|\,\pi_{\gamma e^{(j)}}\big),
\end{gathered}
\end{equation}
where $\gamma e^{(j)}$ is the refitted edit with step $\gamma$ in Eq.~\eqref{eq:unit}, $\Gamma$ is a small set of steps, and $\delta_j$ is the divergence that remains.

\noindent\textbf{Proposition 3.} \emph{On the fitted responses, the $r$ readout--value pairs of a block reproduce the predictions of the ridge fit as closely as any linear readout through $r$ dimensions can.}
\emph{Proof.} Appendix~\ref{app:rekey}, via Lemmas~\ref{lem:linear}--\ref{lem:neuron} \citep{eckart1936approximation}.

\section{Experiments}\label{sec:exp}

We evaluate \name{} through the following research questions (RQs).
\textbf{RQ1:} How does \name{} compare with prompting, visual zoom-in, harness evolution, and fine-tuning in distribution?
\textbf{RQ2:} Does its gain generalize to held-out (OOD) benchmarks and to other base models?
\textbf{RQ3:} Does the base model alone keep the gain of its harness after internalization, and do rounds on its own responses add to a one-shot edit (Section~\ref{sec:rhc})?
\textbf{RQ4:} Does supervision inside the model carry more of this gain into the weights than supervision at its output (Section~\ref{sec:regap})?
\textbf{RQ5:} Do a measured site and strength make the edit more effective than preset ones, and at what cost (Section~\ref{sec:rekey})?

\begin{table*}[p]
\centering
\setlength{\abovecaptionskip}{5pt}
\footnotesize
\setlength{\tabcolsep}{0pt}
\setlength{\aboverulesep}{0.25ex}
\setlength{\belowrulesep}{0.4ex}
\providecommand{\hd}[2]{\shortstack{#1\\#2}}
\begin{tabular*}{\textwidth}{@{}l@{\extracolsep{\fill}}cccccccccccc@{}}
\toprule
 & \multicolumn{6}{c}{In-distribution (ID)} & \multicolumn{6}{c}{Held-out (OOD)} \\
\cmidrule{2-7}\cmidrule{8-13}
Variant & \hd{Baby}{Vision} & BLINK & CharXiv & \hd{Visu}{Logic} & \hd{VLMs Are}{Biased} & Avg. & \hd{ChartQA}{Pro} & \hd{Hallusion}{Bench} & \hd{Perception}{Bench} & \hd{Spatial}{Viz} & \hd{Visual}{Puzzles} & Avg. \\
\midrule
Qwen3.8-27B (base) & 39.1 & 74.0 & 79.9 & 34.4 & 74.7 & 60.4 & 68.0 & 74.2 & 50.3 & 62.8 & 67.4 & 64.5\\
view edit only & 58.9 & 75.8 & 84.1 & 43.0 & 78.4 & 68.0 & 73.7 & 79.7 & 54.9 & 63.3 & 72.1 & 68.8\\
\midrule
\multicolumn{13}{@{}l}{\textit{Recursive Harness Internalization (Section~\ref{sec:rhc})}}\\
\ \ harness kept in the request & 58.6 & 76.3 & 85.4 & 40.6 & 87.0 & 69.6 & 73.2 & 77.1 & 57.8 & 61.7 & 72.1 & 68.4\\
\ \ one round & 59.6 & 75.5 & 85.9 & 44.5 & 88.0 & 70.7 & 73.4 & 78.6 & 58.9 & 62.2 & 71.4 & 68.9\\
\ \ teacher's responses only & 60.2 & 74.7 & 85.9 & 44.0 & 89.3 & 70.8 & 72.7 & 79.9 & 58.1 & 63.0 & 68.0 & 68.3\\
\midrule
\multicolumn{13}{@{}l}{\textit{ReGap (Section~\ref{sec:regap})}}\\
\ \ output only (distillation) & 62.2 & 75.8 & 83.3 & 42.4 & 80.7 & 68.9 & 72.9 & 78.1 & 55.2 & 65.1 & 68.0 & 67.9\\
\ \ last block only & 59.4 & 75.8 & 84.6 & 43.8 & 78.9 & 68.5 & 72.9 & 77.6 & 55.5 & 61.2 & 70.6 & 67.6\\
\ \ direct difference as signal & 59.6 & 74.7 & 82.3 & 40.9 & 89.3 & 69.4 & 73.4 & 80.5 & 54.7 & 64.3 & 71.6 & 68.9\\
\midrule
\multicolumn{13}{@{}l}{\textit{ReKey (Section~\ref{sec:rekey})}}\\
\ \ mean gap, no regression & 60.2 & 76.0 & 84.4 & 42.4 & 79.7 & 68.5 & 74.0 & 78.4 & 57.0 & 63.5 & 72.7 & 69.1\\
\ \ site preset (view in MLP) & 38.3 & 73.4 & 81.3 & 38.0 & 81.8 & 62.6 & 67.7 & 76.8 & 52.6 & 61.2 & 67.7 & 65.2\\
\ \ strength preset (1.0) & 59.4 & 74.0 & 84.1 & 42.4 & 82.3 & 68.4 & 72.9 & 78.6 & 56.3 & 63.8 & 71.4 & 68.6\\
\ \ constant key & 60.4 & 76.3 & 85.4 & 43.2 & 89.8 & 71.0 & 71.9 & 80.7 & 56.5 & 63.3 & 72.7 & 69.0\\
\midrule
\multicolumn{13}{@{}l}{\textit{Null edits on top of the view edit}}\\
\ \ random readouts & 60.7 & 75.8 & 86.7 & 41.4 & 75.8 & 68.1 & 72.7 & 76.6 & 56.8 & 61.7 & 72.9 & 68.1\\
\ \ random values & 60.2 & 75.8 & 84.1 & 41.4 & 77.9 & 67.9 & 73.4 & 77.3 & 58.3 & 63.5 & 72.9 & 69.1\\
\midrule
\smash{\rlap{\color{pinkbg}\rule[-3.1pt]{\textwidth}{10.0pt}}}\textbf{\name{}} & \textbf{60.7} & \textbf{76.6} & \textbf{87.5} & \textbf{44.8} & \textbf{91.1} & \textbf{72.1} & \textbf{74.0} & \textbf{81.3} & \textbf{59.1} & \textbf{66.1} & \textbf{72.9} & \textbf{70.7}\\
\bottomrule
\end{tabular*}
\caption{Ablations of the three components of \name{}, accuracy (\%), mean of three runs. Each group replaces one choice of the section it names; procedure-edit variants are deployed on the view edit (row 2). \emph{Direct difference} uses the state difference of two runs as the signal; null edits keep the procedure edit's neurons, key, and norms but randomize its readouts or values (Appendix~\ref{app:exp}).}
\label{tab:ablation}
\end{table*}

\begin{figure*}[p]
\centering
\setlength{\abovecaptionskip}{5pt}
\includegraphics[width=0.245\textwidth]{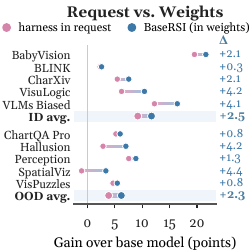}\hfill\includegraphics[width=0.245\textwidth]{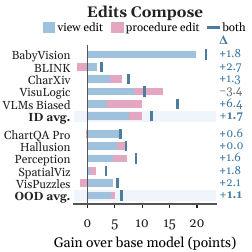}\hfill\includegraphics[width=0.245\textwidth]{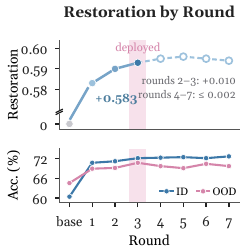}\hfill\includegraphics[width=0.245\textwidth]{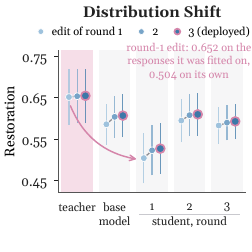}\\[-1pt]
{\small\makebox[0.245\textwidth]{\hspace*{0.0368\textwidth}(a) Harness vs. BaseRSI}\hfill\makebox[0.245\textwidth]{\hspace*{0.0368\textwidth}(b) Composition of edits}\hfill\makebox[0.245\textwidth]{\hspace*{0.0451\textwidth}(c) Inner-loop rounds}\hfill\makebox[0.245\textwidth]{\hspace*{0.0507\textwidth}(d) Shift of responses}}\\[3pt]
\includegraphics[width=0.245\textwidth]{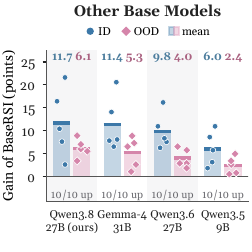}\hfill\includegraphics[width=0.245\textwidth]{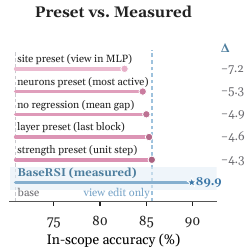}\hfill\includegraphics[width=0.245\textwidth]{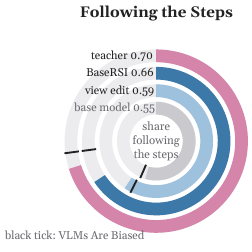}\hfill\includegraphics[width=0.245\textwidth]{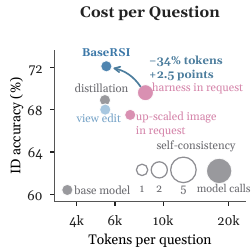}\\[-1pt]
{\small\makebox[0.245\textwidth]{\hspace*{0.0417\textwidth}(e) Cross-model gain}\hfill\makebox[0.245\textwidth]{\hspace*{-0.0245\textwidth}(f) Measured choices}\hfill\makebox[0.245\textwidth]{\hspace*{0.0588\textwidth}(g) Step adherence}\hfill\makebox[0.245\textwidth]{\hspace*{0.0466\textwidth}(h) Inference cost}}
\caption{\textbf{Internalization, generalization, and process.} (a) Gain over the base model with the harness in the request and with \name{}; $\Delta$: their difference. (b) Gains of each edit alone (stacked) and of both; $\Delta$: both minus the sum. (c) Restoration (share of the teacher divergence removed) on 28 held-out responses of the round-3 student, and test accuracy, from the base model to round 7. (d) Restoration by each round's edit on responses of the teacher, the base model, and each round's student; error bars: standard error. (e) Gain of \name{} per benchmark on four base models; bars: means. (f) In-scope accuracy with one choice of \name{} preset instead of measured; $\Delta$: against \name{}. (g) Share of responses that follow the steps; tick: VLMs Are Biased. (h) Tokens against ID accuracy; area: model calls.}
\label{fig:analysis}
\end{figure*}

\begin{figure*}[t]
\setlength{\abovecaptionskip}{5pt}%
\begin{minipage}[t]{0.69\textwidth}\vspace{0pt}
\centering
\scriptsize
\setlength{\tabcolsep}{2.0pt}
\renewcommand{\arraystretch}{1.15}
\begin{tabular*}{\linewidth}{@{}l@{\extracolsep{\fill}}cccc@{\hspace{7pt}}lcccc@{}}
\toprule
Setting & Restor. & In sc. & ID & OOD & Setting & Restor. & In sc. & ID & OOD\\
\midrule
base model & -- & -- & 60.4 & 64.5 & view edit only & -- & $+$14.7 & 68.0 & 68.8\\
\multicolumn{5}{@{}l}{\textit{Replayed responses (ours: three rounds, own responses)}} & \multicolumn{5}{@{}l}{\textit{\rlap{Signal and positions (ours: per-block gap, up to 512 positions)}\phantom{Signal and positions (ours: per-block gap, tail + 256 positions)}}}\\
\ \ one round & 0.583 & $+$16.3 & 70.7 & 68.9 & \ \ last-block state & 0.271 & $+$12.1 & 68.6 & 67.7\\
\ \ two rounds & 0.590 & $+$17.6 & 71.2 & 69.1 & \ \ tail + 64 positions & 0.575 & $+$16.3 & 70.7 & 68.4\\
\ \ base model's only & 0.515 & $+$17.3 & 70.5 & 67.9 & \ \ 16 responses & 0.541 & $+$17.3 & 70.8 & 68.0\\
\multicolumn{5}{@{}l}{\textit{Acceptance (ours: leave one benchmark out)}} & \ \ 64 responses & 0.578 & $+$19.0 & 70.4 & 69.5\\
\ \ macro average only & 0.545 & $+$16.7 & 71.1 & 68.8 & \multicolumn{5}{@{}l}{\textit{Readouts and neurons (ours: fitted, least active)}}\\
\multicolumn{5}{@{}l}{\textit{Blocks written (ours: all 64)}} & \ \ mean gap only & 0.075 & $+$14.1 & 68.5 & 69.1\\
\ \ last block & 0.005 & $+$14.4 & 68.5 & 67.6 & \ \ regression-matrix directions & 0.576 & $+$16.7 & 70.5 & 68.0\\
\ \ last 16 & 0.304 & $+$14.1 & 68.9 & 67.4 & \ \ random directions & 0.250 & $+$16.0 & 69.1 & 69.3\\
\ \ first 32 & 0.055 & $+$16.7 & 69.5 & 68.2 & \ \ random neurons & 0.474 & $+$17.3 & 71.0 & 67.9\\
\ \ 48 linear-attention & 0.093 & $+$14.4 & 69.4 & 68.8 & \ \ most active neurons & $-$2.119 & $+$13.7 & 65.1 & 65.3\\
\smash{\rlap{\hspace*{159.5pt}\color{pinkbg}\rule[-2.9pt]{183.4pt}{9.4pt}}}\ \ 16 full-attention & 0.580 & $+$14.7 & 70.0 & 67.1 & \textbf{\name{}} & \textbf{0.593} & \textbf{$+$19.0} & \textbf{72.1} & \textbf{70.7}\\
\bottomrule
\end{tabular*}
\par\vspace{2pt}
{\small (a) Fine-grained ablations of the procedure edit}
\end{minipage}\hfill
\begin{minipage}[t]{0.292\textwidth}\vspace{0pt}
\centering
\includegraphics[width=\linewidth,trim=0 4.7pt 0 0,clip]{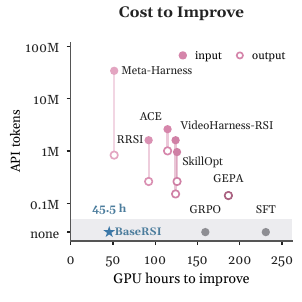}\\[-1pt]
{\small\hspace*{0.205\linewidth}(b) Cost of improving}
\end{minipage}
\par\vspace{3pt}
\centering
\includegraphics[width=0.245\textwidth]{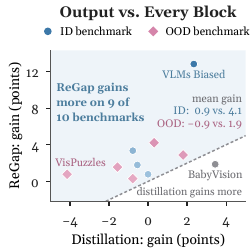}\hfill\includegraphics[width=0.245\textwidth]{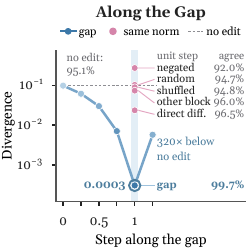}\hfill\includegraphics[width=0.245\textwidth]{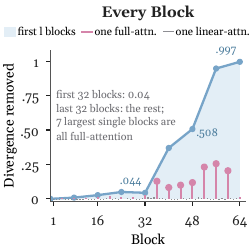}\hfill\includegraphics[width=0.245\textwidth]{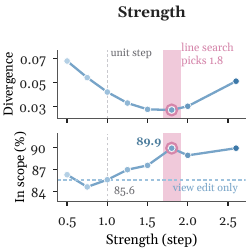}\\[-1pt]
{\small \makebox[0.245\textwidth]{\hspace*{0.0429\textwidth}(c) Inside vs. output}\hfill\makebox[0.245\textwidth]{\hspace*{0.0483\textwidth}(d) Exactness of the gap}\hfill\makebox[0.245\textwidth]{\hspace*{0.0429\textwidth}(e) Each block}\hfill\makebox[0.245\textwidth]{\hspace*{0.0497\textwidth}(f) Measured strength}}
\caption{\textbf{Fine-grained ablations, cost, and mechanisms of ReGap and ReKey (RQ3--RQ5).} (a) Each row changes one choice of \name{} (setting in parentheses); Restor.: restoration; In sc.: gain (points) on the in-scope validation questions. (b) GPU hours and API tokens (filled: input, open: output). (c) Gain over the view edit per benchmark, distillation ($x$) against ReGap ($y$). (d) Divergence left after a step along the gap and after a unit step along other directions of equal norm; percentages: next-token agreement with the teacher. (e) Divergence removed by one block (stems) and by the first $l$ blocks (area). (f) Held-out divergence and in-scope accuracy per step.}
\label{fig:mech}
\end{figure*}

\runin{Setup.}\label{sec:setup}
Five in-distribution (ID) benchmarks, BabyVision \citep{chen2026babyvision}, BLINK \citep{fu2024blink}, CharXiv \citep{wang2024charxiv}, VisuLogic \citep{xu2026visulogic}, and VLMs Are Biased \citep{vo2025vision}, supply the development questions; five out-of-distribution (OOD) benchmarks, ChartQAPro \citep{masry2025chartqapro}, HallusionBench \citep{guan2024hallusionbench}, PerceptionBench \citep{lin2026perceptionbench}, SpatialViz \citep{wang2026spatialviz}, and VisualPuzzles \citep{song2025visualpuzzles}, are held out from every loop, and each test set has 128 questions on unseen images.
We compare with prompting and sampling, visual zoom-in, harness evolution, and fine-tuning on the same base model, Qwen3.8-27B, and list two closed-source models for reference \citep{claudehaiku55,gpt6luna} (Table~\ref{tab:main}).
Over eight generations the base model proposes 269 candidates from its own failures, and the accepted entries, a finer view and a counting procedure, are internalized as the view edit and the procedure edit.
We report accuracy over three runs and the restoration of an edit, the fraction of the divergence from the teacher that it removes on held-out responses (Appendix~\ref{app:exp}).

\runin{Main results (RQ1, RQ2).}\label{sec:main}
\name{} has the highest mean accuracy on all ten benchmarks and raises the ID and OOD averages of the base model by 11.7 and 6.1 points (Table~\ref{tab:main}).
Fine-tuning on the same development questions (SFT, GRPO) barely moves these averages, while the harness that the outer loop verifies lifts them, so the gain lies in what the harness learns rather than in fitting these questions.
Prompting, zoom-in, and harness evolution act through the request, whereas \name{} writes this gain into the weights, where it is largest on the benchmarks of the internalized entries (Table~\ref{tab:ablation}).
The acceptance test carries the gain to the held-out benchmarks: the finer view, which the base model proposes from its own failures, keeps its gain with any one benchmark left out, and testing the macro average alone lowers the OOD average (Figure~\ref{fig:mech}a).
The same loop, run from the original weights of Gemma-4-31B \citep{team2026gemma}, Qwen3.6-27B, and Qwen3.5-9B, internalizes entries that differ in part from those of Qwen3.8-27B and raises every benchmark score of each model (Figure~\ref{fig:analysis}e, Appendix~\ref{app:data}).

\runin{Recursive Harness Internalization (RQ3).}\label{sec:analysis}
The edited base model on original requests exceeds the same model with its harness in the request on all ten benchmarks and answers in one call (Figures~\ref{fig:analysis}a and~\ref{fig:analysis}h).
Inside the entry's scope the two are level (Appendix~\ref{app:mech}), and the surplus comes from the questions that the router leaves out: there the teacher's request has no procedure, while the student carries it in its weights behind a key that reads its own state, the advantage of a student trained with privileged information \citep{vapnik2009new,lopez2015unifying}.
Its edits compose without refitting because each is fitted on the base weights at its own site, and together they exceed the sum of their gains (Figure~\ref{fig:analysis}b).
At the same view, the procedure edit gains on every benchmark (Table~\ref{tab:ablation}), while the up-scaled image alone in the request leaves the base model near the view edit (Figure~\ref{fig:analysis}h).
Rounds on the student's own responses make the edit hold on its own outputs, and three suffice (Figure~\ref{fig:analysis}c): a one-shot edit fitted on the teacher's responses loses part of its effect on its own responses (Figure~\ref{fig:analysis}d), and the rounds add to it, while as many refits on the teacher's add nothing (Table~\ref{tab:ablation}), so the gain comes from whose responses are replayed.

\runin{ReGap (RQ4).}
Supervising every block with the gap carries the entry into the weights: in its scope the edited model matches the entry in the request (Figure~\ref{fig:mech}a, Appendix~\ref{app:mech}), and it follows the steps of the entry about as often as the teacher (Figure~\ref{fig:analysis}g).
A unit step along the gap removes the divergence from the teacher, while gaps that are negated, shuffled, or taken from another block leave it (Figure~\ref{fig:mech}d), and the direct difference of the two runs' states as the signal scores lower (Table~\ref{tab:ablation}).
Supervision at the output alone, distilling the same teacher, gains less than half of what ReGap gains, also when it trains only the neurons that \name{} writes (Figure~\ref{fig:mech}c, Appendix~\ref{app:exp}), since its signal travels back through every block from a few hundred responses.
On these neurons, gradient training toward the per-block gaps comes close to the closed-form solve, while the output loss restores nothing (Appendix~\ref{app:mech}), so the gain comes from the signal.
Null edits with the same neurons, key, and norms leave the ID average unchanged (Table~\ref{tab:ablation}), and the divergence falls only as all blocks' gaps are added (Figure~\ref{fig:mech}e), so the fitted readouts of every block carry the gain.

\runin{ReKey (RQ5).}
Each measured quantity decides part of the edit, and every preset site, layer, or strength stays below the measured one (Figure~\ref{fig:analysis}f).
The measured site sends the view entry to the patch embedding; in MLP neurons it loses most of its gain (Table~\ref{tab:ablation}).
The regression makes the edit a function of what each block reads: the mean gap alone adds little, and readouts along the principal directions of the fit supply the gain (Table~\ref{tab:ablation}, Figure~\ref{fig:mech}a).
The measured strength offsets the shrinkage of the fit, and the line search picks the step with the least held-out divergence (Figure~\ref{fig:mech}f).
The fitted key decides where the edit acts: the edited model cites the procedure on 32\% of in-scope responses and 0.5\% of the others (Appendix~\ref{app:mech}), and a constant key lowers both averages (Table~\ref{tab:ablation}).
Internalization runs forward passes and one linear solve per block, with no backward pass or API call, in 45.5 GPU hours, the least of all methods (Figure~\ref{fig:mech}b); the procedure edit rewrites 0.23\% of the weights and leaves the remaining weights bit-identical.

\section{Conclusion}\label{sec:conclusion}

\name{} unifies harness evolution, weight editing, and internalization in one recursive loop with ReGap and ReKey, and the base model alone improves on every benchmark.

\section*{LLM Usage Statement}
Claude (Anthropic) was used to polish and edit the writing and for LaTeX typesetting.
Qwen3.8-27B is part of the method and is described in Sections~\ref{sec:method} and~\ref{sec:exp}.
The authors reviewed all content and take full responsibility for it.

{
    \small
    \setlength{\bibsep}{5.7pt plus 0.3pt}%
    \let\textcircaccent\^\renewcommand{\^}[1]{\ifmmode^\else\textcircaccent{#1}\fi}%
    \bibliographystyle{ieeenat_fullname}
    \bibliography{refs}

@inproceedings{agrawal2026gepa,
  title={Gepa: Reflective prompt evolution can outperform reinforcement learning},
  author={Agrawal, Lakshya A and Tan, Shangyin and Soylu, Dilara and Ziems, Noah and Khare, Rishi and Opsahl-Ong, Krista and Singhvi, Arnav and Shandilya, Herumb and Ryan, Michael J and Jiang, Meng and others},
  booktitle={International Conference on Learning Representations},
  volume={2026},
  pages={8479--8565},
  year={2026}
}

@article{anderson1972simple,
  title={A simple neural network generating an interactive memory},
  author={Anderson, James A},
  journal={Mathematical biosciences},
  volume={14},
  number={3-4},
  pages={197--220},
  year={1972},
  publisher={Elsevier}
}

@article{anderson1951estimating,
  title={Estimating linear restrictions on regression coefficients for multivariate normal distributions},
  author={Anderson, Theodore Wilbur},
  journal={The Annals of Mathematical Statistics},
  pages={327--351},
  year={1951},
  publisher={JSTOR}
}

@article{arditi2024refusal,
  title={Refusal in language models is mediated by a single direction},
  author={Arditi, Andy and Obeso, Oscar and Syed, Aaquib and Paleka, Daniel and Panickssery, Nina and Gurnee, Wes and Nanda, Neel},
  journal={Advances in Neural Information Processing Systems},
  volume={37},
  pages={136037--136083},
  year={2024}
}

@article{baghel2025resolving,
  title={Resolving UnderEdit \& OverEdit with Iterative \& Neighbor-Assisted Model Editing},
  author={Baghel, Bhiman Kumar and Jordan, Emma and Shi, Zheyuan Ryan and Li, Xiang Lorraine},
  journal={arXiv preprint arXiv:2503.11895},
  year={2025}
}

@inproceedings{bau2020rewriting,
  title={Rewriting a deep generative model},
  author={Bau, David and Liu, Steven and Wang, Tongzhou and Zhu, Jun-Yan and Torralba, Antonio},
  booktitle={European conference on computer vision},
  pages={351--369},
  year={2020},
  organization={Springer}
}

@article{behrouz2026language,
  title={Language models need sleep: Learning to self-modify and consolidate memories},
  author={Behrouz, Ali and Hashemi, Farnoosh and Javanmard, Adel and Mirrokni, Vahab},
  journal={arXiv preprint arXiv:2606.03979},
  year={2026}
}

@inproceedings{beyer2023flexivit,
  title={Flexivit: One model for all patch sizes},
  author={Beyer, Lucas and Izmailov, Pavel and Kolesnikov, Alexander and Caron, Mathilde and Kornblith, Simon and Zhai, Xiaohua and Minderer, Matthias and Tschannen, Michael and Alabdulmohsin, Ibrahim and Pavetic, Filip},
  booktitle={2023 IEEE/CVF Conference on Computer Vision and Pattern Recognition (CVPR)},
  pages={14496--14506},
  year={2023},
  organization={IEEE}
}

@article{chen2024exact,
  title={Exact conversion of in-context learning to model weights in linearized-attention transformers},
  author={Chen, Brian K and Hu, Tianyang and Jin, Hui and Lee, Hwee Kuan and Kawaguchi, Kenji},
  journal={arXiv preprint arXiv:2406.02847},
  year={2024}
}

@article{chen2026babyvision,
  title={Babyvision: Visual reasoning beyond language},
  author={Chen, Liang and Xie, Weichu and Liang, Yiyan and He, Hongfeng and Zhao, Hans and Yang, Zhibo and Huang, Zhiqi and Wu, Haoning and Lu, Haoyu and Bao, Yiping and others},
  journal={arXiv preprint arXiv:2601.06521},
  year={2026}
}

@article{cohen2024evaluating,
  title={Evaluating the ripple effects of knowledge editing in language models},
  author={Cohen, Roi and Biran, Eden and Yoran, Ori and Globerson, Amir and Geva, Mor},
  journal={Transactions of the Association for Computational Linguistics},
  volume={12},
  pages={283--298},
  year={2024},
  publisher={MIT Press, One Broadway, 12th Floor, Cambridge, Massachusetts 02142, USA~…}
}

@article{dherin2025learning,
  title={Learning without training: The implicit dynamics of in-context learning},
  author={Dherin, Benoit and Munn, Michael and Mazzawi, Hanna and Wunder, Michael and Gonzalvo, Javier},
  journal={arXiv preprint arXiv:2507.16003},
  year={2025}
}

@article{ebrahimpour2026resisting,
  title={Resisting Harmful Instructions through Training-Free Weight Editing},
  author={Ebrahimpour-Boroojeny, Ali and Yang, Xiaocong and Schapiro, Samuel and Borhani, Zahra and Ortega, Francisco R and Zheng, Yujia and Zhai, ChengXiang},
  journal={arXiv preprint arXiv:2610.21189},
  year={2026}
}

@article{eckart1936approximation,
  title={The approximation of one matrix by another of lower rank},
  author={Eckart, Carl and Young, Gale},
  journal={Psychometrika},
  volume={1},
  number={3},
  pages={211--218},
  year={1936},
  publisher={Cambridge University Press}
}

@article{friedman2001greedy,
  title={Greedy function approximation: a gradient boosting machine},
  author={Friedman, Jerome H},
  journal={Annals of statistics},
  pages={1189--1232},
  year={2001},
  publisher={JSTOR}
}

@inproceedings{fu2024blink,
  title={Blink: Multimodal large language models can see but not perceive},
  author={Fu, Xingyu and Hu, Yushi and Li, Bangzheng and Feng, Yu and Wang, Haoyu and Lin, Xudong and Roth, Dan and Smith, Noah A and Ma, Wei-Chiu and Krishna, Ranjay},
  booktitle={European Conference on Computer Vision},
  pages={148--166},
  year={2024},
  organization={Springer}
}

@article{gao2026experience,
  title={Experience Funnel: A State-Policy Alternating Loop for Self-Evolving Agents},
  author={Gao, Wenbo and Song, Zhaomou and Ji, Zhiyuan and Liu, Renxi and Li, Xing and Yu, Xianzhi and Li, Xiaoguang and Cheung, James Chung-wai and Lin, Weizhe and Wang, Yaoyuan},
  journal={arXiv preprint arXiv:2609.08919},
  year={2026}
}

@article{team2026gemma,
  title={Gemma 4 technical report},
  author={Team, Gemma and Abd, Sherif El and Aggarwal, Vaibhav and Algayres, Robin and Andreev, Alek and Bachem, Olivier and Ballantyne, Ian and Brick, Cormac and C{\u{a}}rbune, Victor and Casbon, Michelle and others},
  journal={arXiv preprint arXiv:2607.02770},
  year={2026}
}

@inproceedings{geva2021transformer,
  title={Transformer feed-forward layers are key-value memories},
  author={Geva, Mor and Schuster, Roei and Berant, Jonathan and Levy, Omer},
  booktitle={Proceedings of the 2021 conference on empirical methods in natural language processing},
  pages={5484--5495},
  year={2021}
}

@article{goldwaser2025equivalence,
  title={Equivalence of Context and Parameter Updates in Modern Transformer Blocks},
  author={Goldwaser, Adrian and Munn, Michael and Gonzalvo, Javier and Dherin, Benoit},
  journal={arXiv preprint arXiv:2511.17864},
  year={2025}
}

@inproceedings{gu2024model,
  title={Model editing harms general abilities of large language models: Regularization to the rescue},
  author={Gu, Jia-Chen and Xu, Hao-Xiang and Ma, Jun-Yu and Lu, Pan and Ling, Zhen-Hua and Chang, Kai-Wei and Peng, Nanyun},
  booktitle={Proceedings of the 2024 Conference on Empirical Methods in Natural Language Processing},
  pages={16801--16819},
  year={2024}
}

@inproceedings{guan2024hallusionbench,
  title={Hallusionbench: an advanced diagnostic suite for entangled language hallucination and visual illusion in large vision-language models},
  author={Guan, Tianrui and Liu, Fuxiao and Wu, Xiyang and Xian, Ruiqi and Li, Zongxia and Liu, Xiaoyu and Wang, Xijun and Chen, Lichang and Huang, Furong and Yacoob, Yaser and others},
  booktitle={2024 IEEE/CVF Conference on Computer Vision and Pattern Recognition (CVPR)},
  pages={14375--14385},
  year={2024},
  organization={IEEE}
}

@article{hase2023does,
  title={Does localization inform editing? surprising differences in causality-based localization vs. knowledge editing in language models},
  author={Hase, Peter and Bansal, Mohit and Kim, Been and Ghandeharioun, Asma},
  journal={Advances in Neural Information Processing Systems},
  volume={36},
  pages={17643--17668},
  year={2023}
}

@article{hebbar2026sia,
  title={SIA: Self Improving AI with Harness \& Weight Updates},
  author={Hebbar, Prannay and Manawat, Yogendra and Verboomen, Samuel and Ivanova, Alesia and Palanimalai, Selvam and Bhatia, Kunal and Baskaran, Vignesh},
  journal={arXiv preprint arXiv:2605.27276},
  year={2026}
}

@article{holm1979simple,
  title={A simple sequentially rejective multiple test procedure},
  author={Holm, Sture},
  journal={Scandinavian journal of statistics},
  pages={65--70},
  year={1979},
  publisher={JSTOR}
}

@article{huang2023transformer,
  title={Transformer-patcher: One mistake worth one neuron},
  author={Huang, Zeyu and Shen, Yikang and Zhang, Xiaofeng and Zhou, Jie and Rong, Wenge and Xiong, Zhang},
  journal={arXiv preprint arXiv:2301.09785},
  year={2023}
}

@article{izenman1975reduced,
  title={Reduced-rank regression for the multivariate linear model},
  author={Izenman, Alan Julian},
  journal={Journal of multivariate analysis},
  volume={5},
  number={2},
  pages={248--264},
  year={1975},
  publisher={Elsevier}
}

@article{jiang2026distill,
  title={Distill skills into weights, not prompts: Abstract skills as privileged signals for on-policy self-distillation},
  author={Jiang, Yubo and Xie, Fengying and Jiang, Zhiguo and Zhang, Haopeng},
  journal={arXiv preprint arXiv:2608.09826},
  year={2026}
}

@article{kohonen1972correlation,
  title={Correlation matrix memories},
  author={Kohonen, Teuvo},
  journal={IEEE transactions on computers},
  volume={100},
  number={4},
  pages={353--359},
  year={1972},
  publisher={IEEE}
}

@article{kumaran2016learning,
  title={What learning systems do intelligent agents need? Complementary learning systems theory updated},
  author={Kumaran, Dharshan and Hassabis, Demis and McClelland, James L},
  journal={Trends in cognitive sciences},
  volume={20},
  number={7},
  pages={512--534},
  year={2016},
  publisher={Elsevier}
}

@article{lee2026meta,
  title={Meta-harness: End-to-end optimization of model harnesses},
  author={Lee, Yoonho and Nair, Roshen and Zhang, Qizheng and Lee, Kangwook and Khattab, Omar and Finn, Chelsea},
  journal={arXiv preprint arXiv:2603.28052},
  year={2026}
}

@article{li2026revisiting,
  title={Revisiting DAgger in the Era of LLM-Agents},
  author={Li, Changhao and Qiang, Rushi and Huang, Jiawei and Gao, Chenxiao and Zhang, Chao and He, Niao and Dai, Bo},
  journal={arXiv preprint arXiv:2605.12913},
  year={2026}
}

@article{lin2026perceptionbench,
  title={Perceptionbench: Evaluating atomic visual perception in multimodal large language models},
  author={Lin, Zichao and Xie, Yifeng and Qu, Bowen and Wang, Haiming and Li, Jia and Wu, Haoning and Dong, Yuhao and Yang, Zuhao and Zhu, Jinguo and Lu, Haoyu and others},
  journal={arXiv preprint arXiv:2607.24957},
  year={2026}
}

@article{lopez2015unifying,
  title={Unifying distillation and privileged information},
  author={Lopez-Paz, David and Bottou, L{\'e}on and Sch{\"o}lkopf, Bernhard and Vapnik, Vladimir},
  journal={arXiv preprint arXiv:1511.03643},
  year={2015}
}

@article{luo2026video,
  title={Video-RSI: Recursive Self-Improvement of Video Understanding Agents via Harness Evolution},
  author={Luo, Bingjun and Guo, Jialin and Li, Siqi},
  journal={arXiv preprint arXiv:2609.37950},
  year={2026}
}

@article{luo2026harness,
  title={Harness-Aware Self-Evolving: Co-Evolving Model Weights, Harness, and Task Solutions},
  author={Luo, Haochen and Huang, Yi and Luo, Sichun and Liu, Fengyuan and Li, Lei and Hu, Zefa and Feng, Junlan and Liu, Qi},
  journal={arXiv preprint arXiv:2607.03935},
  year={2026}
}

@article{madaan2023self,
  title={Self-refine: Iterative refinement with self-feedback},
  author={Madaan, Aman and Tandon, Niket and Gupta, Prakhar and Hallinan, Skyler and Gao, Luyu and Wiegreffe, Sarah and Alon, Uri and Dziri, Nouha and Prabhumoye, Shrimai and Yang, Yiming and others},
  journal={Advances in neural information processing systems},
  volume={36},
  pages={46534--46594},
  year={2023}
}

@article{mason1999boosting,
  title={Boosting algorithms as gradient descent},
  author={Mason, Llew and Baxter, Jonathan and Bartlett, Peter and Frean, Marcus},
  journal={Advances in neural information processing systems},
  volume={12},
  year={1999}
}

@inproceedings{masry2025chartqapro,
  title={Chartqapro: A more diverse and challenging benchmark for chart question answering},
  author={Masry, Ahmed and Islam, Mohammed Saidul and Ahmed, Mahir and Bajaj, Aayush and Kabir, Firoz and Kartha, Aaryaman and Laskar, Md Tahmid Rahman and Rahman, Mizanur and Rahman, Shadikur and Shahmohammadi, Mehrad and others},
  booktitle={Findings of the Association for Computational Linguistics: ACL 2025},
  pages={19123--19151},
  year={2025}
}

@article{mazzawi2025transmuting,
  title={Transmuting prompts into weights},
  author={Mazzawi, Hanna and Dherin, Benoit and Munn, Michael and Goldwaser, Adrian and Wunder, Michael and Gonzalvo, Javier},
  journal={arXiv preprint arXiv:2510.08734},
  year={2025}
}

@article{mcclelland1995there,
  title={Why there are complementary learning systems in the hippocampus and neocortex: insights from the successes and failures of connectionist models of learning and memory.},
  author={McClelland, James L and McNaughton, Bruce L and O'Reilly, Randall C},
  journal={Psychological review},
  volume={102},
  number={3},
  pages={419},
  year={1995},
  publisher={American Psychological Association}
}

@article{meng2022locating,
  title={Locating and editing factual associations in gpt},
  author={Meng, Kevin and Bau, David and Andonian, Alex and Belinkov, Yonatan},
  journal={Advances in neural information processing systems},
  volume={35},
  pages={17359--17372},
  year={2022}
}

@article{meng2022mass,
  title={Mass-editing memory in a transformer},
  author={Meng, Kevin and Sharma, Arnab Sen and Andonian, Alex and Belinkov, Yonatan and Bau, David},
  journal={arXiv preprint arXiv:2210.07229},
  year={2022}
}

@inproceedings{merzin2026context,
  title={In-Context Compilation: Encoding Context-Induced Behavioral Shift into Model Weights},
  author={Merzin, Michael and Elisha, Yehonatan and Barkan, Oren and Toib, Yonatan and Koenigstein, Noam},
  booktitle={Findings of the Association for Computational Linguistics: EMNLP 2026},
  pages={20777--20796},
  year={2026}
}

@inproceedings{pan2025precise,
  title={Precise localization of memories: A fine-grained neuron-level knowledge editing technique for llms},
  author={Pan, Haowen and Wang, Xiaozhi and Cao, Yixin and Shi, Zenglin and Yang, Xun and Li, Juanzi and Wang, Meng},
  booktitle={International Conference on Learning Representations},
  volume={2025},
  pages={482--502},
  year={2025}
}

@incollection{pearl2022direct,
  title={Direct and indirect effects},
  author={Pearl, Judea},
  booktitle={Probabilistic and causal inference: the works of Judea Pearl},
  pages={373--392},
  year={2022}
}

@article{ren2026scaling,
  title={Scaling Self-Evolving Agents via Parametric Memory},
  author={Ren, Tao and Luo, Weiyao and Yang, Hui and Zhu, Rongzhi and Huang, Xiang and Wu, Yuchuan and Chou, Bingxue and Ye, Jieping and Liang, Jiafeng and Li, Yongbin and others},
  journal={arXiv preprint arXiv:2606.04536},
  year={2026}
}

@inproceedings{ross2010efficient,
  title={Efficient reductions for imitation learning},
  author={Ross, St{\'e}phane and Bagnell, Drew},
  booktitle={Proceedings of the thirteenth international conference on artificial intelligence and statistics},
  pages={661--668},
  year={2010},
  organization={JMLR Workshop and Conference Proceedings}
}

@inproceedings{ross2011reduction,
  title={A reduction of imitation learning and structured prediction to no-regret online learning},
  author={Ross, St{\'e}phane and Gordon, Geoffrey and Bagnell, Drew},
  booktitle={Proceedings of the fourteenth international conference on artificial intelligence and statistics},
  pages={627--635},
  year={2011},
  organization={JMLR Workshop and Conference Proceedings}
}

@article{shao2024deepseekmath,
  title={Deepseekmath: Pushing the limits of mathematical reasoning in open language models},
  author={Shao, Zhihong and Wang, Peiyi and Zhu, Qihao and Xu, Runxin and Song, Junxiao and Bi, Xiao and Zhang, Haowei and Zhang, Mingchuan and Li, YK and Wu, Yang and others},
  journal={arXiv preprint arXiv:2402.03300},
  year={2024}
}

@article{shazeer2020glu,
  title={Glu variants improve transformer},
  author={Shazeer, Noam},
  journal={arXiv preprint arXiv:2002.05202},
  year={2020}
}

@inproceedings{shen2025zoomeye,
  title={Zoomeye: Enhancing multimodal llms with human-like zooming capabilities through tree-based image exploration},
  author={Shen, Haozhan and Zhao, Kangjia and Zhao, Tiancheng and Xu, Ruochen and Zhang, Zilun and Zhu, Mingwei and Yin, Jianwei},
  booktitle={Proceedings of the 2025 Conference on Empirical Methods in Natural Language Processing},
  pages={6613--6629},
  year={2025}
}

@article{song2025visualpuzzles,
  title={Visualpuzzles: Decoupling multimodal reasoning evaluation from domain knowledge},
  author={Song, Yueqi and Ou, Tianyue and Kong, Yibo and Li, Zecheng and Neubig, Graham and Yue, Xiang},
  journal={arXiv preprint arXiv:2504.10342},
  year={2025}
}

@article{su2026coevo,
  title={CoEvo-RSI: When Harnesses Evolve, Which Models Should Follow?},
  author={Su, Jiaqi and Zhang, Guibin and Shao, Jie-Jing and Pang, Cong and Yao, Tiankuo and Tsang, Ivor and Yin, Haiyan and Yan, Shuicheng},
  journal={arXiv preprint arXiv:2610.25078},
  year={2026}
}

@article{takehi2026sparks,
  title={Sparks of Weight-Space Programming: LLMs can natively read \& write neural network weights},
  author={Takehi, Rikiya and Teoh, Jayden and Zhang, Alex and Khattab, Omar},
  journal={MIT CSAIL research blog. Published November},
  volume={2},
  pages={2026},
  year={2026}
}

@article{tan2026metarsi,
  title={MetaRSI/RSI2: A meta-recursive self-improving system for recursive self-improving systems themselves},
  author={Tan, Zihan and Sun, Leixin and Shi, Zitong and Liu, Yitao and Wu, Jiajun and Brooks, Nathaniel and Qian, Jiaru and Shang, Xiaoran and Huang, Suyuan and Ding, Yi and others},
  journal={arXiv preprint arXiv:2609.06396},
  year={2026}
}

@article{vapnik2009new,
  title={A new learning paradigm: Learning using privileged information},
  author={Vapnik, Vladimir and Vashist, Akshay},
  journal={Neural networks},
  volume={22},
  number={5-6},
  pages={544--557},
  year={2009},
  publisher={Elsevier}
}

@article{vo2025vision,
  title={Vision language models are biased},
  author={Vo, An and Nguyen, Khai-Nguyen and Taesiri, Mohammad Reza and Dang, Vy Tuong and Nguyen, Anh Totti and Kim, Daeyoung},
  journal={arXiv preprint arXiv:2505.23941},
  year={2025}
}

@inproceedings{wang2026spatialviz,
  title={Spatialviz-bench: A cognitively-grounded benchmark for diagnosing spatial visualization in mllms},
  author={Wang, Siting and Pei, Minnan and Sun, Luoyang and Deng, Cheng and Li, Yuchen and Shao, Kun and Tian, Zheng and Zhang, Haifeng and Wang, Jun},
  booktitle={International Conference on Learning Representations},
  volume={2026},
  pages={47210--47279},
  year={2026}
}

@article{wang2022self,
  title={Self-consistency improves chain of thought reasoning in language models},
  author={Wang, Xuezhi and Wei, Jason and Schuurmans, Dale and Le, Quoc and Chi, Ed and Narang, Sharan and Chowdhery, Aakanksha and Zhou, Denny},
  journal={arXiv preprint arXiv:2203.11171},
  year={2022}
}

@article{wang2024charxiv,
  title={Charxiv: Charting gaps in realistic chart understanding in multimodal llms},
  author={Wang, Zirui and Xia, Mengzhou and He, Luxi and Chen, Howard and Liu, Yitao and Zhu, Richard and Liang, Kaiqu and Wu, Xindi and Liu, Haotian and Malladi, Sadhika and others},
  journal={Advances in Neural Information Processing Systems},
  volume={37},
  pages={113569--113697},
  year={2024}
}

@article{wei2022chain,
  title={Chain-of-thought prompting elicits reasoning in large language models},
  author={Wei, Jason and Wang, Xuezhi and Schuurmans, Dale and Bosma, Maarten and Xia, Fei and Chi, Ed and Le, Quoc V and Zhou, Denny and others},
  journal={Advances in neural information processing systems},
  volume={35},
  pages={24824--24837},
  year={2022}
}

@article{xia2026rrsi,
  title={RRSI: Regularized recursive self-improvement of agent harnesses},
  author={Xia, Peng and Han, Rujun and Wang, Zifeng and Chen, Yanfei and Zhuang, Yufan and Lee, Yoonho and Huang, Chengsong and Yu, Han and CuiZhu, Zhongying and Ming, Yifei and others},
  journal={arXiv preprint arXiv:2609.24972},
  year={2026}
}

@article{xu2026evo,
  title={Evo $\^{} 2$: Evolving the Evolver through Weight-Harness Co-Evolution},
  author={Xu, Chejian and Zhang, Shujian and Zhou, Wenxuan and Lambert, John and Chen, Beck and Li, Bo and Hard, Andrew and Mathews, Rajiv},
  journal={arXiv preprint arXiv:2610.13025},
  year={2026}
}

@article{xu2026videoharness,
  title={Videoharness-rsi: Recursive harness self-improvement for long-video understanding with frozen vision-language models},
  author={Xu, Guoyang and Chen, Hao},
  journal={arXiv preprint arXiv:2608.24302},
  year={2026}
}

@inproceedings{xu2026visulogic,
  title={Visulogic: A benchmark for evaluating visual reasoning in multi-modal large language models},
  author={Xu, Weiye and Wang, Jiahao and Wang, Weiyun and Chen, Zhe and Zhou, Wengang and Yang, Aijun and Lu, Lewei and Li, Houqiang and Wang, Xiaohua and Zhu, Xizhou and others},
  booktitle={International Conference on Learning Representations},
  volume={2026},
  pages={25966--26003},
  year={2026}
}

@article{yang2026training,
  title={Training-Free On-Policy Distillation},
  author={Yang, Bangji and Fan, Jiajun and Zhu, Zhaoyu and Ma, Hongbo and Liu, Ge},
  journal={arXiv preprint arXiv:2610.22236},
  year={2026}
}

@article{yang2026skillopt,
  title={Skillopt: Executive strategy for self-evolving agent skills},
  author={Yang, Yifan and Gong, Ziyang and Huang, Weiquan and Yang, Qihao and Zhou, Ziwei and Huang, Zisu and Li, Yan and Gao, Xuemei and Dai, Qi and Liu, Bei and others},
  journal={arXiv preprint arXiv:2605.23904},
  year={2026}
}

@article{yao2026learning,
  title={Learning to Look Beyond Screens: Transferable Visual Observation from GUI Grounding},
  author={Yao, Ruilin and Qiu, Xinyu and Zou, Tianyu and Rong, Yi and Xiong, Shengwu and Shen, Shuheng and Meng, Changhua},
  journal={arXiv preprint arXiv:2610.37093},
  year={2026}
}

@article{ye2026harness,
  title={Harness-Zero: Harness Distillation via Agent-as-Harness},
  author={Ye, Haoran and Lu, Yuxing and Dong, Haonan and Su, Zhaochen and Song, Guojie},
  journal={arXiv preprint arXiv:2609.24974},
  year={2026}
}

@article{yu2026co,
  title={Co-Evolving Harnesses and Models: On-Policy Correction Helps Weaker Models Catch Up Where Imitation Fails},
  author={Yu, Zhou and Bi, Bin and Pentyala, Shiva Kumar and Mehrotra, Shubham and Chaudhuri, Sougata and Bhagavath, Shilpa and Chen, Zeyuan and Xu, Ran and Mui, Phil and Zhu, James and others},
  journal={arXiv preprint arXiv:2609.09134},
  year={2026}
}

@article{zeng2026lazare,
  title={Lazare, veni foras: Reprogramming Dormant Neurons to Backdoor AI Agents},
  author={Zeng, Yuchen and Severi, Giorgio and Bullwinkel, Blake and Horvitz, Eric and Papailiopoulos, Dimitris},
  journal={arXiv preprint arXiv:2611.15118},
  year={2026}
}

@inproceedings{zhang2025uncovering,
  title={Uncovering overfitting in large language model editing},
  author={Zhang, Mengqi and Ye, Xiaotian and Liu, Qiang and Ren, Pengjie and Chen, Zhumin and others},
  booktitle={International Conference on Learning Representations},
  volume={2025},
  pages={79550--79568},
  year={2025}
}

@inproceedings{zhang2026agentic,
  title={Agentic context engineering: Evolving contexts for self-improving language models},
  author={Zhang, Qizheng and Hu, Changran and Upasani, Shubhangi and Ma, Boyuan and Hong, Fenglu and Kamanuru, Vamsidhar and Rainton, Jay and Wu, Chen and Ji, Mengmeng and Li, Hanchen and others},
  booktitle={International Conference on Learning Representations},
  volume={2026},
  pages={86069--86100},
  year={2026}
}

@article{zhao2026rpmem,
  title={RPMem: Learning Long-Term Recurrent Parametric Memory Across Sessions for LLM Agents},
  author={Zhao, Fanyu and Cao, Ruike and Dong, Liang and Yao, Fugen and Xu, Jian and Jiang, Guanjun and Zhang, Han and Zhao, Yifei and Li, Yinsheng},
  journal={arXiv preprint arXiv:2609.23466},
  year={2026}
}

@article{zhao2026training,
  title={Training with Harnesses: On-Policy Harness Self-Distillation for Complex Reasoning},
  author={Zhao, Zhengyang and Ma, Lu and Zhang, Wentao},
  journal={arXiv preprint arXiv:2605.08741},
  year={2026}
}

@article{zhou2026hierarchical,
  title={Hierarchical self-improvement: A framework for task-specific evolvable agent harnesses},
  author={Zhou, Tailin},
  journal={arXiv preprint arXiv:2608.08466},
  year={2026}
}

@article{zweiger2026self,
  title={Self-adapting language models},
  author={Zweiger, Adam and Pari, Jyo and Guo, Han and Kim, Yoon and Agrawal, Pulkit},
  journal={Advances in Neural Information Processing Systems},
  volume={38},
  pages={74084--74115},
  year={2026}
}

@misc{claudehaiku55,
  author={{Anthropic}},
  title={Introducing {Claude Haiku 5.5}},
  howpublished={\url{https://www.anthropic.com/claude-haiku-5-5}},
  year={2026}
}

@article{rhi,
  author={Lee, Hyunin and Xu, Jinglue and Seely, Jeffrey and Lee, Donghyun and Zaharia, Matei and Tang, Yujin},
  title={Recursive Harness Self-Improvement},
  journal={arXiv preprint arXiv:2607.15524},
  year={2026}
}

@inproceedings{harnessgoes,
  author={Liu, Kaiyuan and Huang, Zezhou and Singh, Janvijay and Celikyilmaz, Asli},
  title={What Happens When the Harness Comes and Goes? A Controlled Study of On-Policy Harness Self-Distillation},
  booktitle={NeurIPS 2026 Workshop on Transitioning from Pre-training to Post-training},
  year={2026}
}

@misc{gpt6luna,
  author={{OpenAI}},
  title={Introducing {GPT-6 Sol} and {Luna}},
  howpublished={\url{https://openai.com/index/introducing-gpt-6-sol-and-luna/}},
  year={2026}
}

@misc{pixelproof2026,
  key={PixelProof},
  title={{PixelProof}: Visual Question Generation through Forward--Inverse Agreement},
  year={2026},
  howpublished={\url{https://github.com/ProjectPixelProof/PixelProof}}
}
}
\clearpage
\maketitlesupplementary
\appendix
\renewcommand{\thesection}{\Alph{section}}
\captionsetup{singlelinecheck=false}%
\renewcommand{\topfraction}{0.7}\renewcommand{\bottomfraction}{0.5}\renewcommand{\textfraction}{0.15}\renewcommand{\floatpagefraction}{0.8}
\renewcommand{\dbltopfraction}{0.9}\renewcommand{\dblfloatpagefraction}{0.7}\setcounter{topnumber}{2}\setcounter{bottomnumber}{1}\setcounter{totalnumber}{4}\setcounter{dbltopnumber}{4}
\makeatletter\setlength{\@fptop}{0pt}\setlength{\@dblfptop}{0pt}\setlength{\@fpsep}{8pt plus 1fil}\setlength{\@dblfpsep}{10pt plus 1fil}\setlength{\@fpbot}{0pt}\setlength{\@dblfpbot}{0pt plus 1fil}\makeatother
\raggedbottom
\providecommand{\FAS}{1.18}%
\theoremstyle{plain}
\newtheorem{lemma}{Lemma}[section]
\newtheorem{proposition}[lemma]{Proposition}
\theoremstyle{definition}
\newtheorem{property}[lemma]{Property}
\newtheorem{remark}[lemma]{Remark}

\noindent This appendix treats the three parts of \name{} in turn: the teacher--student gap of ReGap (Appendix~\ref{app:gap}), the key--value edit of ReKey with the patch-embedding edit (Appendices~\ref{app:rekey} and~\ref{app:interface}), and the guarantee of the rounds of Recursive Harness Internalization (Appendix~\ref{app:dagger}), followed by the outer loop and its acceptance test (Appendix~\ref{app:judge}), the algorithms (Appendix~\ref{app:algo}), experimental details (Appendix~\ref{app:exp}), mechanism checks (Appendix~\ref{app:mech}), the complete data (Appendix~\ref{app:data}), and a case study (Appendix~\ref{app:case}).
Algebraic identities hold on every replayed response, and every bound is stated with its explicit constants.

\section{Theory of \name{}}\label{app:theory}

\subsection{Setting and the Exact Gap}\label{app:gap}

Fix weights $\theta$, an entry $c$, a request $q$ inside its scope, and a response $y_{1:T}$.
A replay places the tokens of the request first and the tokens of $y$ after them; the former are the prompt positions, whose number may differ between $c(q)$ and $q$, and the latter are the response positions $\cS=\{1,\dots,T\}$, with embeddings $\vst_0$ of the tokens of $y$.

\begin{property}[Replay]\label{ass:replay}
(a) Blocks are causal: the state of a position after block $l$ depends only on the states of that position and of earlier positions after block $l-1$.
(b) The embedding of a response token is the same in both replays; positions enter through relative encodings within attention.
\end{property}

By (a), the prompt states evolve on their own, so $\vz_l$ and $\vz^{c}_l$ do not depend on $y$.
The block map $F_l(\vst;\vz)$ returns the output of block $l$ at the response positions when its input is $\vz$ at the prompt positions and $\vst$ at the response positions, with the positional layout of the replay that $\vz$ comes from.
The teacher and the student are the recursions $\vst^{c}_l=F_l(\vst^{c}_{l-1};\vz^{c}_{l-1})$ and $\vst_l=F_l(\vst_{l-1};\vz_{l-1})$, both started at $\vst_0$ by (b), and the gap of Eq.~\eqref{eq:gap} is $\vg_l=F_l(\vst^{c}_{l-1};\vz^{c}_{l-1})-F_l(\vst^{c}_{l-1};\vz_{l-1})$.

\begin{lemma}[Hybrid state]\label{lem:hybrid}
Let additions $\vect{a}_1,\dots,\vect{a}_L$ enter the response states of the student, $\tilde\vst_l=F_l(\tilde\vst_{l-1};\vz_{l-1})+\vect{a}_l$ with $\tilde\vst_0=\vst_0$.
Then the prompt states stay $\vz_l$, and whenever $\tilde\vst_{l-1}=\vst^{c}_{l-1}$, we have $\tilde\vst_l=\vst^{c}_l$ if and only if $\vect{a}_l=\vg_l$.
\end{lemma}
\begin{proof}
Prompt positions precede the response, so by Property~\ref{ass:replay}(a) additions at response positions do not reach them.
If $\tilde\vst_{l-1}=\vst^{c}_{l-1}$, then $\tilde\vst_l=F_l(\vst^{c}_{l-1};\vz_{l-1})+\vect{a}_l$, while $\vst^{c}_l=F_l(\vst^{c}_{l-1};\vz^{c}_{l-1})=F_l(\vst^{c}_{l-1};\vz_{l-1})+\vg_l$ by the definition of the gap in Eq.~\eqref{eq:gap}.
\end{proof}

\begin{proposition}[Proposition 2, formal]\label{prop:exact}
By Property~\ref{ass:replay}, the student with additions $\vect{a}_l=\vg_l$ for $l=1,\dots,L$ has $\tilde\vst_l=\vst^{c}_l$ for every $l$, so its next-token distribution equals that of the teacher at every response position; they are the only additions that keep $\tilde\vst_l=\vst^{c}_l$ at every block.
\end{proposition}
\begin{proof}
Induction on $l$. The case $l=0$ is Property~\ref{ass:replay}(b), and the step is Lemma~\ref{lem:hybrid}. The next-token distribution at a response position is a function of its final state.
\end{proof}

The two terms of $\vg_l$ run block $l$ on the same response states and differ only in the prompt states that its attention layer reads.
What the entry does to later blocks through the response states is carried by these states once the earlier gaps are added, so no part is counted twice.
For an addition $\vect{a}$ at block $l$, let $D_l(\vect{a})=\tfrac12\|F_l(\vst^{c}_{l-1};\vz_{l-1})+\vect{a}-\vst^{c}_l\|^2$ be the squared distance of the block to its teacher. Then $-\nabla_{\vect{a}}D_l(0)=\vg_l$ and $D_l(\vg_l)=0$: the gap is the exact descent direction of $D_l$ in activation space, a unit step along it closes the distance, and it comes from forward passes.

An edit replaces the additions by a function of the state.
Let the edit add $E_l(\vx_l)$ to the output of block $l$ at every position, where $\vx_l$ is the input of its MLP and $E_l$ is taken net of the former output of the rewritten neurons; let $\vh_l$ collect the states at all positions, $\hat\vh_l=(\vz_l,\vst^{c}_l)$ be the hybrid state, $\tilde F_l$ the edited block on all positions, $\tilde\vh_l$ the states of the edited student, and
\begin{equation}\label{eq:residual}
\vr_l=\tilde F_l(\hat\vh_{l-1})-\hat\vh_l .
\end{equation}
The MLP input of the edited block on $\hat\vh_{l-1}$ is the input $\vx_l$ of Eq.~\eqref{eq:pairs}, so $\vr_l$ equals $E_l(\vx_l)-\vg_l$ at the response positions and $E_l(\vx_l)$ at the prompt positions.

\begin{lemma}[Error propagation]\label{lem:prop}
Let $\|\cdot\|$ be a norm on states and $\Lambda_l$ a Lipschitz constant of $\tilde F_l$ on a set that contains $\tilde\vh_{l-1}$ and $\hat\vh_{l-1}$. Then
\begin{equation}\label{eq:prop}
\big\|\tilde\vh_L-\hat\vh_L\big\|\le\sum_{l=1}^{L}\Big(\prod_{m=l+1}^{L}\Lambda_m\Big)\|\vr_l\| .
\end{equation}
With $\Lambda_{\mathrm{out}}$ a Lipschitz constant of the map from the final state to the logits, from $\|\cdot\|$ to the largest absolute logit difference at any position, the divergence from the teacher to the edited student at every response position is at most $\tfrac12\Lambda_{\mathrm{out}}^2$ times the square of the right side of Eq.~\eqref{eq:prop}.
\end{lemma}
\begin{proof}
Since $\tilde\vh_l-\hat\vh_l=\tilde F_l(\tilde\vh_{l-1})-\tilde F_l(\hat\vh_{l-1})+\vr_l$, we get $\|\tilde\vh_l-\hat\vh_l\|\le\Lambda_l\|\tilde\vh_{l-1}-\hat\vh_{l-1}\|+\|\vr_l\|$, and $\tilde\vh_0=\hat\vh_0$; unrolling gives Eq.~\eqref{eq:prop}.
For logits $\vect{a},\vect{a}'$ with $\vect{d}=\vect{a}'-\vect{a}$ and $p$ the distribution of $\vect{a}$, the divergence from $p$ to the distribution of $\vect{a}'$ is $\log\E_p e^{d}-\E_p d$, which Hoeffding's lemma bounds by $(\max d-\min d)^2/8\le\|\vect{d}\|_\infty^2/2$.
\end{proof}

Lemma~\ref{lem:prop} is the reason for the two regressions of ReKey: Eq.~\eqref{eq:ridge} minimizes $\vr_l$ at the response positions in mean square, and the third group of Eq.~\eqref{eq:key} keeps the edit, hence $\vr_l$, small at the prompt positions.
With $\vr_l=0$ at every block the lemma returns Proposition~\ref{prop:exact}.

\subsection{ReKey}\label{app:rekey}

Fix a block and drop its index. Let $\bar\vx\neq0$ be the mean input on the fitted pairs, $\hat\vx=\bar\vx/\|\bar\vx\|$, $P=I-\hat\vx\hat\vx^{\T}$, and $\vk^{0}=\bar\vx/\|\bar\vx\|^2$ the mean direction.

\begin{lemma}[Centred inputs are linear readouts]\label{lem:linear}
$P(\vx-\bar\vx)=P\vx$ for every $\vx$, so a readout $\vu^{\T}P(\vx-\bar\vx)$ of the centred input equals $(P\vu)^{\T}\vx$, a readout of the input with no offset term, and $\vk^{0\T}\bar\vx=1$.
\end{lemma}
\begin{proof}
$P\bar\vx=\bar\vx-\hat\vx(\hat\vx^{\T}\bar\vx)=0$.
\end{proof}

Let $\tilde X\in\R^{n\times H}$ stack the rows $(P\vx)^{\T}$ of the fitted pairs and $\tilde G$ the rows $(\vg-\vb)^{\T}$.
The ridge solution of Eq.~\eqref{eq:ridge} and the readouts that ReKey writes are
\begin{equation}\label{eq:rrr}
\begin{gathered}
\hat W=(\Sigma+\lambda I)^{-1}C,\qquad
\tilde X\hat W\approx ASV_r^{\T},\\
\vu_i=P\hat W\vv_i,\quad i=1,\dots,r,
\end{gathered}
\end{equation}
where $\Sigma=\tilde X^{\T}\tilde X/n$, $C=\tilde X^{\T}\tilde G/n$, the middle expression is the rank-$r$ truncated singular value decomposition of the fitted values, and $\vv_i$ is column $i$ of $V_r$.
The $r$ directions are shared by all positions of all responses, and the fitted gaps hold no answer of a development question.
For any $\hat W\in\R^{H\times H}$, in particular the ridge solution of Eq.~\eqref{eq:rrr}, let $\hat Y=\tilde X\hat W$ have singular values $s_1\ge s_2\ge\cdots$ and right singular vectors $\vv_1,\vv_2,\dots$, and let $V_r=[\vv_1,\dots,\vv_r]$ and $\hat W_r=\hat WV_rV_r^{\T}$, of rank at most $r$.

\begin{lemma}[Best reproduction through $r$ dimensions]\label{lem:ey}
For every matrix $Z\in\R^{n\times H}$ of rank at most $r$,
\begin{equation}\label{eq:ey}
\big\|\hat Y-\tilde X\hat W_r\big\|_F^2=\sum_{i>r}s_i^2\le\big\|\hat Y-Z\big\|_F^2 .
\end{equation}
If $\lambda=0$ and $\hat W$ is a least-squares solution, $\hat W_r$ also minimizes $\|\tilde G-\tilde XW\|_F$ over all $W$ of rank at most $r$.
\end{lemma}
\begin{proof}
$\tilde X\hat W_r=\hat YV_rV_r^{\T}$ is the rank-$r$ truncation of the singular value decomposition of $\hat Y$, so Eq.~\eqref{eq:ey} is the Eckart--Young theorem.
If $\lambda=0$, the normal equations make $\tilde G-\hat Y$ orthogonal to the column space of $\tilde X$, which contains $\hat Y-\tilde XW$; hence $\|\tilde G-\tilde XW\|_F^2=\|\tilde G-\hat Y\|_F^2+\|\hat Y-\tilde XW\|_F^2$, and by Eq.~\eqref{eq:ey} the second term is minimized by $\hat W_r$.
\end{proof}

The map $\vx\mapsto\hat W_r^{\T}P\vx$ equals $\sum_{i=1}^{r}(\vu_i^{\T}\vx)\vv_i$ with $\vu_i=P\hat W\vv_i$, the readouts of Eq.~\eqref{eq:rrr}.
For $\lambda>0$, Eq.~\eqref{eq:ey} is the optimality of $\hat W_r$: among readouts through $r$ dimensions it reproduces the predictions of the ridge fit most closely.

\begin{lemma}[One pair is one neuron]\label{lem:neuron}
A neuron of a SwiGLU MLP with rows $\vw_{\mathrm g}=\beta\vk$, $\vw_{\mathrm u}=\vu$, and $\vw_{\mathrm d}=\gamma\vv/\beta$ outputs $\gamma\,\omega_\beta(\vk^{\T}\vx)(\vu^{\T}\vx)\vv$ with $\omega_\beta(z)=z\sigma(\beta z)$, and
\begin{equation}\label{eq:ramp}
\sup_{z\in\R}\big|\omega_\beta(z)-\max(z,0)\big|=\frac{w_0-1}{\beta}<\frac{0.2785}{\beta},
\end{equation}
where $w_0\approx1.2785$ solves $w=1+e^{-w}$.
\end{lemma}
\begin{proof}
The neuron computes $\silu(\beta\vk^{\T}\vx)(\vu^{\T}\vx)\gamma\vv/\beta$ and $\silu(\beta z)=\beta z\sigma(\beta z)$.
For $z\ge0$, $z-\omega_\beta(z)=z\sigma(-\beta z)$, and for $z<0$, $|\omega_\beta(z)|=|z|\sigma(-\beta|z|)$; both equal $w\sigma(-w)/\beta$ with $w=\beta|z|$.
The function $w\sigma(-w)$ on $w\ge0$ is maximal where $w\sigma(w)=1$, that is, $w=1+e^{-w}$, and there $w\sigma(-w)=w-w\sigma(w)=w-1$.
\end{proof}
An edit with $r$ readouts thus rewrites $3(r{+}1)H$ weights of a block, the gate, up, and down rows of its $r{+}1$ neurons.

\begin{proposition}[Proposition 3, formal]\label{prop:rekey}
On the fitted pairs, among all linear readouts of the input through at most $r$ dimensions, the readouts and values $(\vu_i,\vv_i)_{i\le r}$ of Eq.~\eqref{eq:rrr} give the predictions closest in Frobenius norm to those of the ridge fit, and each pair is one neuron of the original MLP, scaled by $\omega_\beta$ of the key.
\end{proposition}
\begin{proof}
On the fitted pairs, the predictions of a linear readout through at most $r$ dimensions form a matrix of rank at most $r$, so the first part is Lemma~\ref{lem:ey}; by Lemma~\ref{lem:linear} the pairs read the input itself, and the second part is Lemma~\ref{lem:neuron}.
\end{proof}

The truncation is computed by a randomized range finder with oversampling and power iterations.
Pair $0$ has readout $\vk^{0}$ and value $\vb$, so by Lemma~\ref{lem:linear} it adds the mean gap on average, scaled by the same key.

\subsection{Patch-Embedding Edit}\label{app:interface}

\begin{lemma}[Resampled patch kernel]\label{lem:adjoint}
Let the patch embedding map a patch $\vect{v}\in\R^{C\times p\times p}$ to $\langle K_o,\vect{v}\rangle$ for each output channel $o$, let $B$ be a linear up-sampling from $p'\times p'$ to $p\times p$ pixels applied to every input channel, and let $K'_o=B^{\T}K_o$. Then $\langle K'_o,\vect{v}'\rangle=\langle K_o,B\vect{v}'\rangle$ for every patch $\vect{v}'\in\R^{C\times p'\times p'}$ of the original image.
\end{lemma}
\begin{proof}
$B^{\T}$ is the adjoint of $B$ in each channel.
\end{proof}

When the view entry up-samples the image by an integer factor with nearest neighbours, each $p\times p$ patch of the up-sampled image is $B$ applied to one $p'\times p'$ patch of the original image.
By Lemma~\ref{lem:adjoint}, the vision tower with kernel $K'$ and patch size $p'$ on the original image then produces, on the same grid, the patch embeddings of the teacher; the bias and all other weights of the tower are unchanged.

\subsection{Rounds on the Model's Own Responses}\label{app:dagger}

Lemma~\ref{lem:prop} holds on one replayed response.
Which responses are replayed is the subject of this subsection: the student generates its own responses, so the distribution of the regression pairs moves with the edit.

Let $W=(W_1,\dots,W_L)$ stack one linear readout per block, and let the teacher-forced loss of $W$ on a replayed response $\xi=(q,y)$ be
\begin{equation}\label{eq:tfloss}
\ell(W;\xi)=\sum_{l=1}^{L}\frac{1}{|\cS(y)|}\sum_{t\in\cS(y)}\big\|\vg_l(t)-\vb_l-W_l^{\T}P_l\vx_l(t)\big\|^2 .
\end{equation}
Let $d_j$ be the distribution of the responses replayed at round $j$, $L_j(W)=\E_{\xi\sim d_j}\ell(W;\xi)$, and $L^{\lambda}_j(W)=L_j(W)+\lambda\|W\|_F^2$, with $\vb_l$, $P_l$, and $\lambda>0$ held fixed across rounds.
The refit of Eq.~\eqref{eq:aggregate} solves Eq.~\eqref{eq:ridge} on the pairs of all rounds so far, $W^{(j)}=\argmin_W\sum_{i\le j}L^{\lambda}_i(W)$ with $W^{(0)}=0$, which is the follow-the-leader update.

Let $M_x$ and $M_g$ bound the pairs: $\|P_l\vx_l(t)\|\le M_x$ and $\sum_l\|\vg_l(t)-\vb_l\|^2\le M_g^2$ for every pair.

\begin{lemma}[Regret of the aggregated refit]\label{lem:ftl}
Let $\rho=M_g/\sqrt{\lambda}$ and $G=2M_x(M_x\rho+M_g)+2\lambda\rho$. For every $N\ge1$ and any sequence $d_1,\dots,d_N$,
\begin{equation}\label{eq:regret}
\frac1N\sum_{j=1}^{N}L^{\lambda}_j\big(W^{(j-1)}\big)-\min_{W}\frac1N\sum_{j=1}^{N}L^{\lambda}_j(W)\le\frac{G^2(1+\ln N)}{\lambda N}.
\end{equation}
\end{lemma}
\begin{proof}
Each $L_j$ is a convex quadratic, so $L^{\lambda}_j$ is $2\lambda$-strongly convex and $\Phi_j=\sum_{i\le j}L^{\lambda}_i$ is $2\lambda j$-strongly convex with minimizer $W^{(j)}$.
Since $\lambda\|W^{(j)}\|_F^2\le\Phi_j(W^{(j)})/j\le\Phi_j(0)/j\le M_g^2$, every iterate lies in the ball of radius $\rho$, on which the gradient of each $L^{\lambda}_j$ has norm at most $G$.
By induction on $N$, $\sum_{j\le N}L^{\lambda}_j(W^{(j)})\le\sum_{j\le N}L^{\lambda}_j(W)$ for every $W$, so the left side of Eq.~\eqref{eq:regret} is at most $\frac1N\sum_j[L^{\lambda}_j(W^{(j-1)})-L^{\lambda}_j(W^{(j)})]$.
For each $j$, strong convexity gives $\lambda j\|W^{(j-1)}-W^{(j)}\|_F^2\le\Phi_j(W^{(j-1)})-\Phi_j(W^{(j)})$, and the right side equals $\Phi_{j-1}(W^{(j-1)})-\Phi_{j-1}(W^{(j)})+L^{\lambda}_j(W^{(j-1)})-L^{\lambda}_j(W^{(j)})\le G\|W^{(j-1)}-W^{(j)}\|_F$, because $W^{(j-1)}$ minimizes $\Phi_{j-1}$.
Hence $\|W^{(j-1)}-W^{(j)}\|_F\le G/(\lambda j)$ and $L^{\lambda}_j(W^{(j-1)})-L^{\lambda}_j(W^{(j)})\le G^2/(\lambda j)$; summing $1/j$ gives the claim.
\end{proof}

Lemma~\ref{lem:ftl} holds for any sequence of response distributions, so it holds when the responses of each round are generated by the student that the previous refit produces; Lemma~\ref{lem:own} states this case, with the first $j_0$ rounds left to responses that the student does not generate.

\begin{lemma}[Own responses]\label{lem:own}
Suppose that for every round $j>j_0$ the responses replayed at round $j$ are generated on original requests by the student that carries the edit written from $W^{(j-1)}$. Then
\begin{equation}\label{eq:own}
\begin{aligned}
\min_{j_0<j\le N}L_j\big(W^{(j-1)}\big)\le\frac{N}{N-j_0}\Big(&\min_{W}\frac1N\sum_{j=1}^{N}L^{\lambda}_j(W)\\
&+\frac{G^2(1+\ln N)}{\lambda N}\Big),
\end{aligned}
\end{equation}
and the left side is the smallest teacher-forced gap that the fit of a round leaves on the responses of its own student.
\end{lemma}
\begin{proof}
The minimum is at most the mean over $j_0<j\le N$; all terms are non-negative and $L_j\le L^{\lambda}_j$, so this mean is at most $\frac{1}{N-j_0}\sum_{j\le N}L^{\lambda}_j(W^{(j-1)})$, and Lemma~\ref{lem:ftl} bounds the sum. For $j>j_0$, $d_j$ is the distribution of the responses of the student of round $j-1$, as the left side reads.
\end{proof}

\begin{proposition}[Proposition 1, formal]\label{prop:dagger}
In the setting of Lemma~\ref{lem:own}, for every $N>j_0$ the smallest teacher-forced gap that a round's fit leaves on the responses of its own student is at most $N/(N-j_0)$ times the smallest regularized gap $\min_W\frac1N\sum_{j\le N}L^{\lambda}_j(W)$ that a single linear readout leaves on the responses of all rounds together, plus a term of order $\ln N/N$; with $j_0$ fixed, the factor tends to one.
\end{proposition}
\begin{proof}
Lemma~\ref{lem:own}; the added term is $\frac{N}{N-j_0}\cdot\frac{G^2(1+\ln N)}{\lambda N}$.
\end{proof}

\begin{remark}[Rounds as run]\label{rem:scope}
(i) The warm-up rounds of Algorithm~\ref{alg:sd}, which replay responses generated under the entry and by the unedited base model, are the $j_0$ rounds of Lemma~\ref{lem:own}.
(ii) The lemma asks only that the responses of round $j$ come from the student of round $j-1$; it holds for the student as deployed, with its readouts reduced to rank $r$, keyed, and scaled by the step.
(iii) With $m$ replayed responses per round, Lemma~\ref{lem:ftl} holds for the empirical losses $\hat L_j$ as stated. The differences between $L_j(W^{(j-1)})$ and the loss of each response replayed at round $j$ are martingale differences bounded by $\ell_{\max}$, a bound on the loss of Eq.~\eqref{eq:tfloss} on the ball of radius $\rho$, so by the Azuma--Hoeffding inequality the mean of $L_j(W^{(j-1)})-\hat L_j(W^{(j-1)})$ over $j_0<j\le N$ is at most $\ell_{\max}\sqrt{2\ln(1/\delta)/(m(N-j_0))}$ with probability $1-\delta$; Eq.~\eqref{eq:own} then holds for the population gap with the empirical losses on its right side and this term added.
(iv) The step of Eq.~\eqref{eq:stop} acts on the measured divergence: $\Gamma$ contains the unit step, so the chosen step never leaves a larger held-out divergence than the unit step. 
\end{remark}

\Needspace{4.5\baselineskip}%
\section{Outer Loop and Acceptance Test}\label{app:judge}

\paragraph{Entries, proposer, and router.}
An entry holds a scope in words, or the scope of every request, and an action: a procedure text added to the prompt, request settings, or an operation on the views of the image.
The base model with its original weights is the proposer: it reads its own responses on the fitting questions and writes candidate entries (Table~\ref{tab:routing}). The reference answers of the development questions mark which of these responses fail and score the acceptance test.
The base model is also the router: asked, without thinking and on the original view of the image, whether a request falls inside the scope of an entry, it answers yes or no, which gives $\kappa_c(q)$.
The system kept between generations holds a general layer of entries and a layer of entries accepted for single ID benchmarks.
In the reported run the loop runs eight generations and its proposer writes 269 candidates; the finer view, accepted in generation 1, and the counting procedure, accepted in generation 2, are internalized in this order, the procedure edit on top of the view edit.

\begin{table}[htb]
\centering
\footnotesize
\setlength{\tabcolsep}{3pt}
\begin{tabular}{@{}p{0.37\columnwidth}p{0.60\columnwidth}@{}}
\toprule
Questions & Read by \\
\midrule
Fitting (development) & proposer, router examples, paired replay, fits of readouts and keys, site and strength \\
Validation (development) & acceptance of entries and edits \\
ID test & reported scores \\
OOD tests & reported scores \\
\bottomrule
\end{tabular}
\caption{Data routing; development questions are split by image.}
\label{tab:routing}
\end{table}

\paragraph{Acceptance test.}
A candidate is scored on the validation questions only, against the current system at an equal token budget.
For each scored benchmark $b$, $d_b(c)$ is the mean of the paired differences of its validation questions, with the questions of one image forming one cluster; for a general entry, the benchmark that it was written from is left unscored.
The $p$-value $p(\cdot)$ of Eq.~\eqref{eq:judge} is the one-sided sign-flip $p$-value of the macro mean of the listed benchmark effects, obtained by flipping the signs of the image clusters inside each benchmark.
A candidate is rejected when some scored benchmark is significantly negative at a level divided by the number of scored benchmarks.
An entry for a single ID benchmark is accepted when its gain on the validation questions of that benchmark is significant, with the same rejection on the other benchmarks.
An edit is tested in the same way (Algorithm~\ref{alg:gen}), and in addition against a null edit that writes random readouts into the same neurons.

\begin{lemma}[Family-wise error of the acceptance test]\label{lem:fwer}
For a candidate $c$, let $\mathcal{N}_c$ be the hypothesis that its macro gain is at most zero with some scored benchmark left out. If every leave-one-out $p$-value is valid under its own null, then $p_{\mathrm{gen}}(c)$ is valid under $\mathcal{N}_c$, and the probability that the test accepts at least one candidate of a generation for which $\mathcal{N}_c$ holds is at most the test level $\alpha$.
\end{lemma}
\begin{proof}
Under $\mathcal{N}_c$ some benchmark $b^\star$ has a non-positive macro gain of the others, and $p_{\mathrm{gen}}(c)$ is at least the $p$-value with $b^\star$ left out, which is valid; this is the intersection--union principle. Holm's procedure over the candidates controls the family-wise error at $\alpha$ under any dependence.
\end{proof}

\Needspace{4.5\baselineskip}
\section{Algorithms}\label{app:algo}

\begin{algorithm}[!htb]
\caption{\name{}, generation $n$}\label{alg:gen}
\small
\begin{algorithmic}[1]
\Require weights $\theta_n$, harness $\cH_n$, fitting and validation questions
\State The base model $\theta_0$ proposes candidate entries from its responses on the fitting questions
\State Score them on the validation questions; accept $\cA_n$ by Eq.~\eqref{eq:judge} or for one benchmark
\State $\theta\gets\theta_n$; $\cH\gets\cH_n\cup\cA_n$
\For{each entry $c\in\cA_n$ chosen for internalization}
  \State Compute the fraction $\eta(c)$ by Eq.~\eqref{eq:share}
  \If{$c$ is a view entry with a low fraction}
    \State $e\gets$ the patch kernel resampled to $K'=B^{\T}K$
  \Else
    \State $e\gets$ Rounds$(\theta_0,c)$ (Algorithm~\ref{alg:sd})
  \EndIf
  \If{$e$ passes its acceptance test (Appendix~\ref{app:judge})}
    \State $\theta\gets\theta\oplus e$; $\cH\gets\cH\setminus\{c\}$
  \EndIf
\EndFor
\State $\theta_{n+1}\gets\theta$; $\cH_{n+1}\gets\cH$
\end{algorithmic}
\end{algorithm}

\begin{algorithm}[!htb]
\caption{Rounds of the inner loop}\label{alg:sd}
\small
\begin{algorithmic}[1]
\Require weights $\theta$, entry $c$, questions inside and outside the scope, number of rounds $J$
\State $\cD^{(0)}\gets\emptyset$; $e^{(0)}\gets$ empty edit
\For{$j=1,\dots,J$}
  \State Collect responses: under the entry ($j{=}1$), of the base model ($j{=}2$), of the student exported with $\gamma_{j-1}e^{(j-1)}$ ($j{\ge}3$)
  \State Replay each response with $c(q)$, with $q$, and block by block on the mixed input; add the pairs of Eq.~\eqref{eq:pairs} to $\cD^{(j)}$
  \State $e^{(j)}\gets$ ReKey$(\cD^{(j)})$ (Algorithm~\ref{alg:rk}), one solve per block
  \State Choose $\gamma_j$ and measure $\delta_j(\gamma_j)$ on held-out responses of the current student by Eq.~\eqref{eq:stop}
  \EndFor
\State \textbf{return} $\gamma_Je^{(J)}$, the edit with its values scaled by $\gamma_J$
\end{algorithmic}
\end{algorithm}

\begin{algorithm}[!htb]
\caption{ReKey, one block}\label{alg:rk}
\small
\begin{algorithmic}[1]
\Require pairs $(\vx,\vg)$ of the block, rank $r$, key positions
\State $\bar\vx\gets$ mean input; $\vb\gets$ mean gap; $P\gets I-\hat\vx\hat\vx^{\T}$
\State $\hat W\gets(\Sigma+\lambda I)^{-1}C$, with $\lambda$ chosen on a held-out fold
\State $V_r\gets$ top $r$ right singular vectors of $\tilde X\hat W$; $\vu_i\gets P\hat W\vv_i$
\State Rescale each readout to unit variance and its value by the inverse factor
\State Solve the key of Eq.~\eqref{eq:key}; shift it by the threshold
\State Write rows $(\beta\vk,\vu_i,\vv_i/\beta)$ and $(\beta\vk,\vk^{0},\vb/\beta)$ into the $r+1$ neurons that contribute least
\end{algorithmic}
\end{algorithm}

\paragraph{Paired replay.}
A response is replayed three times by teacher forcing: with the entry, without it, and block by block on the mixed input of Eq.~\eqref{eq:gap}.
The response positions of a replay are the generated tokens and the tail of the chat template before them; the tail is the same with and without the entry, so it is treated as the first tokens of $y$.
For block $l$, the mixed input takes the states of the replay without the entry at the prompt positions and the states of the replay with the entry at the response positions; block $l$ is run once on it with original weights, which rebuilds its attention and recurrent states from the mixed sequence, so each block is asked what it has to add once every block before it is closed.
Pairs are stored at the template tail and at a random sample of reasoning positions of each response.
On the first response of every run the script adds the gaps to the replay without the entry and prints the largest absolute difference to the replay with the entry (Appendix~\ref{app:mech} reports the check in full precision).
For the procedure entry, both replays carry the view entry, so the two differ only in the procedure; the student responses come from the base weights with the edit of the previous round, under the same view entry.

\paragraph{Key.}
The key of a block is the ridge solution over three groups of positions on original requests:
\begin{equation}\label{eq:key}
\vk_l\propto\argmin_{\vk}\sum_{g}\frac{1}{|\cP_g|}\sum_{t\in\cP_g}\big(\vk^{\T}\vx_l(t)-y_g\big)^2+\lambda\|\vk\|^2,
\end{equation}
where the groups are the response positions of requests inside the scope of the entry ($y_g=1$), on which the gap is present, those of requests outside it, and the prompt positions (both $y_g=0$).
With $\vk^{\mathrm{r}}$ the solution of Eq.~\eqref{eq:key}, the written key is $\vk\propto\vk^{\mathrm{r}}-\vartheta\vk^{0}$, so that the factor $\omega_\beta$ is a ramp that starts at the threshold $\vartheta$ of the key value.
The intercept of the key is carried by $\vk^{0}$, since the MLP has no bias terms, and an entry for every request keeps the constant key.

\paragraph{Neurons and controls.}
The neurons of a block are ranked by the mean absolute activation times the norm of the output column on responses inside and outside the scope; the pairs take the lowest-ranked neurons, and neurons of earlier edits are excluded.
Two null edits use the same neurons, key, and norms: \emph{random readouts} rotates the readouts by a random orthogonal map, and \emph{random values} the values.

\paragraph{Restoration.}
The restoration of an edit is $1-\KLb(\pic_\theta\|\pi_e)/\KLb(\pic_\theta\|\pi_\theta)$, the fraction of the divergence from the teacher that the edit removes on one response, averaged over the responses of a set; the divergences we report are means of $\KLb$ over the same responses, so the two are averaged separately.

\Needspace{4.5\baselineskip}
\section{Experimental Details}\label{app:exp}

\paragraph{Benchmarks and splits.}
The five ID benchmarks supply 640 development questions, split by image into 270 fitting and 370 validation questions (54 and 74 per benchmark), and an ID test of 640 questions on disjoint images.
The procedure edit is fitted on the 57 fitting questions that the router sends to the counting entry, and its key is fitted against 75 that it does not.
The five OOD benchmarks supply 640 test questions and are held out from the proposer, the acceptance test, and every fit.
Each test set has 128 questions, and a test score is the accuracy on them, averaged over three runs.
The regression uses the fitting questions alone. A held-out fold of them, drawn by a hash of the question identifier, selects the ridge weight of each block, and the step and the restoration are measured on responses to further fitting questions held out from the regression. The validation questions serve the acceptance test alone. Every setting in Table~\ref{tab:config} is chosen on the development questions. The view edit sets the patch size to 8 and, in the preprocessing configuration stored with the weights, 524,288 to 1,048,576 pixels per image, that is 2,048 to 4,096 image tokens of 2$\times$2 patches; the request carries no image setting. The development splits are released together with the code.

\begin{table}[htb]
\centering
\footnotesize
\setlength{\tabcolsep}{3pt}
\begin{tabular}{@{}>{\raggedright\arraybackslash}p{0.42\columnwidth}>{\raggedright\arraybackslash}p{0.55\columnwidth}@{}}
\toprule
Setting & Value \\
\midrule
Base model, proposer, router & Qwen3.8-27B ($L{=}64$, $H{=}5120$), original weights \\
Reasoning effort & medium \\
Test level $\alpha$ & 0.10 \\
Rank $r$, neurons per block & 64, 65 \\
Edited blocks & all 64 \\
Ridge $\lambda$ & $\{0.03,0.3,3\}\times$ mean diag.\ of $\Sigma_l$ \\
Key slope $\beta$, threshold & 8, 0.25 \\
Steps $\Gamma$ & $\{0.5,0.75,1,1.25,1.5,1.8,2,2.6\}$; selected $\gamma=1.8$ \\
Rounds $J$ & 3 \\
Positions per response & 5 template-tail positions and up to 512 (round 1) or 256 (rounds 2--3) reasoning positions \\
Patch size & 16 to 8 \\
Image tokens after the edit & 2{,}048 to 4{,}096 \\
\bottomrule
\end{tabular}
\caption{Configuration of the released code.}
\label{tab:config}
\end{table}

\paragraph{Metrics.}
Accuracy is scored by one rule-based grader shared by all methods: it reads the final answer of a response, matches option letters, yes or no, and normalized short text, and matches numbers exactly. The standard deviation in Table~\ref{tab:main} is the sample standard deviation of the accuracy over the three runs of a method. Restoration is the fraction of the divergence from the teacher that an edit removes on a held-out response, averaged over the responses of a set (defined in Appendix~\ref{app:algo}).

\paragraph{Significance.}
Table~\ref{tab:sig} tests the gain of \name{} over the base model on the two averages of Table~\ref{tab:main}. Each run of \name{} is paired with the run of the base model on the same sampling seed. Every paired run gains 10.6 to 12.8 points in distribution and 5.3 to 6.6 points held out, so both gains are significant under the two-sided paired $t$-test.

\begin{table}[htb]
\centering
\footnotesize
\setlength{\tabcolsep}{0pt}
\begin{tabular*}{\linewidth}{@{}l@{\extracolsep{\fill}}ccccc@{}}
\toprule
 & Base & \name{} & Gain & 95\% CI & $p$ \\
\midrule
ID average & 60.4 & 72.1 & $+$11.7 & [9.0, 14.4] & 0.003\\
OOD average & 64.5 & 70.7 & $+$6.1 & [4.4, 7.9] & 0.005\\
\bottomrule
\end{tabular*}
\caption{Gain of \name{} over the base model (Qwen3.8-27B): accuracy (\%), paired over the three runs.}
\label{tab:sig}
\end{table}

\paragraph{Baselines and fairness.}
All methods but the closed-source models use the same base model, reasoning effort, token budget, evaluator, and test questions.
Self-consistency takes the majority of five responses of the base model, and the 4-shot baseline adds four examples to the request.
SFT and GRPO train on the development questions until convergence, which gives each its full training budget: SFT with a learning rate of $10^{-6}$, cosine decay, 10\% warm-up, and 32 samples per update; GRPO with a constant learning rate of $2\times10^{-6}$ after 20 warm-up updates and, in each round, 8 sampled responses to each of 64 questions and 8 updates of 64 responses. LBS-27B is a released checkpoint fine-tuned from the same base model and answers with its own crop loop. PixelProof is the LoRA adapter released by its authors (rank 16, $\alpha=32$), merged into the same base model with no further training.
The closed-source models are Claude Haiku 5.5 and GPT-6 Luna, each at its default reasoning effort.
\emph{Harness kept in the request} (Table~\ref{tab:ablation}, Figure~\ref{fig:analysis}a) is the base model with the finer view and the counting procedure kept in the request and no weight edit applied.
\emph{Up-scaled image in request} (Figure~\ref{fig:analysis}h) is the base model with the image enlarged in the request, which gives it the finer view while all of its weights stay unchanged; on the OOD benchmarks it reaches 67.2, against 68.8 for the view edit alone and 70.7 for \name{}.

\paragraph{Ablation variants.}
Each variant of Table~\ref{tab:ablation} and Figure~\ref{fig:mech}a changes one choice of \name{}; Table~\ref{tab:ablation_full} gives their scores on every benchmark.
\emph{View edit only} and \emph{procedure edit only}: one of the edits alone.
\emph{One round}: round 1 only, fitted on responses generated under the entry; \emph{two rounds}: rounds 1 and 2.
\emph{Teacher's responses only}: the same number of refits, with every round replaying responses generated under the entry; \emph{base model's responses only}: every round replays responses of the unedited base model.
\emph{Macro average only}: the acceptance test scores the macro average with no benchmark left out.
\emph{Output only}: the same teacher distilled at the output into every weight; with training restricted to the 4,160 neurons that \name{} writes, and the learning rate and the number of passes chosen on the development questions ($3\times10^{-4}$ and 16, from $10^{-3}$ or $3\times10^{-4}$ and 16 or 32), it reaches 69.8 ID and 69.1 OOD.
\emph{Last block only}, \emph{last 16}, \emph{first 32}, \emph{48 linear-attention}, and \emph{16 full-attention}: the pairs are written into these blocks alone.
\emph{Direct difference as signal}: the difference of the states of the two replays at each block replaces the gap of Eq.~\eqref{eq:gap}; \emph{last-block state}: the difference of their last-block states is the target.
\emph{Tail + 64 positions}: pairs at the template tail and at 64 reasoning positions of each response; \emph{16} and \emph{64 responses}: that many replayed responses per round (\name{}: 344 over three rounds).
\emph{Mean gap only}: pair $0$ only, one constant direction per block.
\emph{Regression-matrix directions} and \emph{random directions}: the readouts at rank 64 along these directions; \emph{random neurons} and \emph{most active neurons}: the pairs written into 65 such neurons per block.
\emph{Site preset}: the view entry is written into MLP neurons in place of the patch embedding.
\emph{Strength preset}: the unit step, $\gamma=1$.
\emph{Constant key}: the pairs with the mean direction as their key.
\emph{Random readouts} and \emph{random values}: null edits with the neurons, key, and norms of the procedure edit, whose readouts or whose values are rotated by a random orthogonal matrix.

\paragraph{Compute.}
Table~\ref{tab:cost} gives the cost of improving the base model for every method, in GPU hours measured on the H800 GPUs with 80\,GB that every method ran on and in the input and output tokens of the language-model API calls made while improving.
\name{} takes 45.5 GPU hours, the least of all methods, against 51.4 to 186.6 for the harness methods and 159.3 and 230.9 for GRPO and SFT; LBS-27B and PixelProof are released weights and have no cost of improving here.
Like fine-tuning it spends no API tokens, while the harness methods spend 0.14\,M to 34.0\,M input tokens.
Fine-tuning pays backward passes once, the harness kept in the request pays its entries again in every request, and \name{} pays forward passes and one linear solve per block once and then answers the original request.

\begin{table*}[tp]
\centering
\footnotesize
\setlength{\tabcolsep}{0pt}
\begin{tabular*}{\textwidth}{@{}l@{\extracolsep{\fill}}ccccc@{}}
\toprule
Improving the base model & GPU hours & API input tokens & API output tokens & Backward passes & Weights changed\\
\midrule
\textbf{\name{}} & \textbf{45.5} & \textbf{0} & \textbf{0} & \textbf{no} & \textbf{0.23\% $+$ patch embedding}\\
Meta-Harness & 51.4 & 34.0\,M & 0.83\,M & no & none\\
RRSI & 92.2 & 1.6\,M & 0.26\,M & no & none\\
ACE & 114.6 & 2.6\,M & 1.0\,M & no & none\\
VideoHarness-RSI & 124.1 & 1.6\,M & 0.15\,M & no & none\\
SkillOpt & 125.8 & 0.95\,M & 0.26\,M & no & none\\
GRPO & 159.3 & 0 & 0 & yes & all\\
GEPA & 186.6 & 0.14\,M & 0.14\,M & no & none\\
SFT & 230.9 & 0 & 0 & yes & all\\
\bottomrule
\end{tabular*}
\caption{Cost of improving the base model: measured H800 GPU hours, and input and output tokens of the API calls made for it.}
\label{tab:cost}
\end{table*}

\paragraph{Repeated runs.}
Each method improves the base model once, and the resulting model or harness answers every test question once in each of three runs, with sampling seeds 0, 1, and 2 shared by all methods.

\Needspace{4.5\baselineskip}
\section{Mechanism Checks}\label{app:mech}

The divergence checks replay 28 held-out responses of the edited student by teacher forcing; the divergence is the mean KL from the teacher, 0.0977 for the unedited student.

\subsection{Recursive Harness Internalization (RQ3)}

\paragraph{The procedure is learned.}
On the 102 in-scope validation questions the edited model alone answers 89.9\% right, against 90.2\% for the teacher and 70.9\% for the base model, and it gives the teacher's final answer on 89.9\% of them (Table~\ref{tab:m3}). On the 266 test questions that the router sends to the entry, it answers 81.2\% right, against 80.2\% for the teacher and 65.5\% for the base model. A judge, the base model reading the steps of the entry and one response, decides whether the response carries the steps out; its wording was fixed on a calibration set of 40 responses and agrees with human labels on 18 of 20 unseen ones. On these 266 questions it finds the steps followed in 65.7\% of the responses of the edited model, against 70.1\% for the teacher, 59.4\% for the view edit alone, and 54.5\% for the base model.

\begin{table}[tb]
\centering
\footnotesize
\setlength{\tabcolsep}{0pt}
\begin{tabular*}{\linewidth}{@{}l@{\extracolsep{\fill}}cccc@{}}
\toprule
Deployment & In scope & Agree & Routed & Steps\\
\midrule
base model & 70.9 & 76.5 & 65.5 & 0.545\\
teacher (entry in request) & 90.2 & 100 & 80.2 & 0.701\\
\textbf{edited model (ours)} & \textbf{89.9} & \textbf{89.9} & \textbf{81.2} & \textbf{0.657}\\
\bottomrule
\end{tabular*}
\caption{The procedure in the weights. In scope: accuracy (\%) on the 102 in-scope validation questions. Agree: share (\%) of them answered as the teacher answers. Routed: accuracy on the 266 test questions the router sends to the entry. Steps: share of their responses that follow the steps of the entry, by the judge.}
\label{tab:m3}
\end{table}

\paragraph{Rounds.}
An edit removes more of the divergence on the responses it is fitted on than on the responses it generates: the round-1 edit, fitted on responses under the entry, removes 0.652 on teacher responses and 0.504 on its own (Table~\ref{tab:m6a}). Each later round raises the restoration on every group of responses, and on the student's own held-out responses it rises from 0.583 to 0.593 through round 3. The rounds on the student's own responses add their gain on held-out benchmarks, 70.7 against 68.3 for the same number of refits on teacher responses (Table~\ref{tab:ablation}).

\begin{table}[tb]
\centering
\footnotesize
\setlength{\tabcolsep}{0pt}
\begin{tabular*}{\linewidth}{@{}l@{\extracolsep{\fill}}ccc@{}}
\toprule
Responses (count) & Round 1 & Round 2 & Round 3\\
\midrule
teacher (14) & 0.652 & 0.654 & 0.655\\
base model (28) & 0.586 & 0.604 & 0.607\\
student of round 1 (20) & 0.504 & 0.523 & 0.527\\
student of round 2 (28) & 0.595 & 0.608 & 0.610\\
student of round 3, held out (28) & 0.583 & 0.590 & 0.593\\
\bottomrule
\end{tabular*}
\caption{Restoration of each round's edit by response group.}
\label{tab:m6a}
\end{table}

\subsection{ReGap (RQ4)}

\paragraph{The gap closes the divergence.}
Table~\ref{tab:m1} adds a direction to the activations of every block. Along the gap the divergence falls with the step and reaches 0.0003 at the unit step, where 99.7\% of the next tokens agree with the teacher, and it rises again past it. In full precision the unit step leaves a divergence below $10^{-10}$, with outputs within a relative error of $1.9\times10^{-4}$ of the teacher, as Proposition~2 states; the remainder in bf16 is rounding. Every other direction of the same norm leaves far more: the direct difference 0.024, the gap of another block 0.073, the gap shuffled across positions 0.089, and a random direction 0.102, while the negated gap raises the divergence to 0.271.

\begin{table}[tb]
\centering
\footnotesize
\setlength{\tabcolsep}{0pt}
\begin{tabular*}{\linewidth}{@{}l@{\extracolsep{\fill}}ccc@{}}
\toprule
Direction added & Step & Divergence & Agreement\\
\midrule
none (student) & 0 & 0.0977 & 0.9507\\
gap & 0.25 & 0.0613 & 0.9577\\
gap & 0.5 & 0.0299 & 0.9723\\
gap & 0.75 & 0.0071 & 0.9858\\
\textbf{gap} & \textbf{1} & \textbf{0.0003} & \textbf{0.9965}\\
gap & 1.25 & 0.0057 & 0.9854\\
\midrule
random direction & 1 & 0.1022 & 0.9466\\
gap shuffled across positions & 1 & 0.0890 & 0.9484\\
gap of another block & 1 & 0.0732 & 0.9601\\
negated gap & 1 & 0.2714 & 0.9202\\
direct difference & 1 & 0.0235 & 0.9651\\
\bottomrule
\end{tabular*}
\caption{Divergence from the teacher and next-token agreement on 28 held-out responses of the student (bf16), after adding a direction of the same norm to the activations of every block.}
\label{tab:m1}
\end{table}

\paragraph{The gaps of all blocks.}
Adding the gaps of the first $l$ blocks removes the whole divergence only when all 64 are added (0.997), and most of it comes between blocks 40 and 56 (Table~\ref{tab:m4a}). Added one at a time, the gaps of the full-attention blocks of the second half remove the most, 0.08 to 0.26 each for blocks 36, 40, 44, 48, 52, 56, and 60; the single-block removals sum to 1.36, more than their joint 0.997, since the divergence is not additive over blocks.

\begin{table}[tb]
\centering
\footnotesize
\setlength{\tabcolsep}{0pt}
\begin{tabular*}{\linewidth}{@{}l@{\extracolsep{\fill}}cc@{}}
\toprule
$l$ & Divergence & Removed\\
\midrule
0 & 0.0977 & 0\\
8 & 0.0968 & 0.009\\
16 & 0.0951 & 0.027\\
24 & 0.0928 & 0.050\\
32 & 0.0934 & 0.044\\
40 & 0.0615 & 0.370\\
48 & 0.0481 & 0.508\\
56 & 0.0050 & 0.948\\
64 & 0.0003 & 0.997\\
\bottomrule
\end{tabular*}
\caption{Divergence after adding the gaps of blocks 1 to $l$.}
\label{tab:m4a}
\end{table}

\paragraph{Controls at equal capacity.}
Three controls hold the edit at 4,160 neurons and change one choice each. Distilled from the same teacher at the output, the neurons that \name{} writes reach 69.8 ID and 69.1 OOD, against 72.1 and 70.7 for \name{} (Appendix~\ref{app:exp}).
Trained by gradient, they restore 0.560 $\pm$ 0.037 with the per-block gaps as targets and $-$0.052 $\pm$ 0.159 with the output loss alone; the closed-form solve restores 0.593 on the same 28 held-out responses. Written as 260 neurons in each of the 16 full-attention blocks, the edit answers 89.3 ID and 87.9 OOD test questions of 128 right, against 92.3 and 90.5 over all 64 blocks. On 306 further responses that entered no choice of a setting, the exported edit restores 0.562 $\pm$ 0.026.

\subsection{ReKey (RQ5)}

\paragraph{Site and step.}
Outside the entry's scope the edited model is 7.0 points more accurate than the base model on validation questions and answers 2.4\% shorter (Table~\ref{tab:m7}). Writing the view entry into MLP neurons loses the out-of-scope gain at every step, and writing the procedure into the last block alone loses part of the in-scope gain at every step, so the measured site and step give the edit its gain both inside and outside the scope of the entry.

\begin{table}[tb]
\centering
\footnotesize
\setlength{\tabcolsep}{0pt}
\begin{tabular*}{\linewidth}{@{}l@{\extracolsep{\fill}}ccc@{}}
\toprule
Site and step & In scope & Out of scope & Length\\
\midrule
procedure in MLP, $\gamma=0.5$ & $+$15.4 & $+$5.6 & $-$4.3\%\\
procedure in MLP, $\gamma=1.0$ & $+$14.7 & $+$5.5 & $-$12.6\%\\
procedure in MLP, $\gamma=2.0$ & $+$18.0 & $+$7.2 & $-$2.4\%\\
view entry in MLP, $\gamma=0.5$ & $+$11.8 & $+$0.4 & $+$4.4\%\\
view entry in MLP, $\gamma=1.0$ & $+$11.8 & $-$0.6 & $+$0.4\%\\
view entry in MLP, $\gamma=2.0$ & $+$9.8 & $-$0.1 & $+$4.2\%\\
last block only, $\gamma=0.5$ & $+$15.4 & $+$4.0 & $-$8.0\%\\
last block only, $\gamma=1.0$ & $+$13.4 & $+$6.2 & $-$4.3\%\\
last block only, $\gamma=2.0$ & $+$15.0 & $+$3.6 & $-$6.3\%\\
\textbf{measured site and step (ours)} & \textbf{$+$19.0} & \textbf{$+$7.0} & \textbf{$-$2.4\%}\\
\bottomrule
\end{tabular*}
\caption{Site and step: accuracy change (points) against the base model on in-scope and out-of-scope validation questions, and the change in length of out-of-scope responses.}
\label{tab:m7}
\end{table}

\paragraph{Rank.}
The restoration rises with the rank up to 64, from 0.075 at rank 0 to 0.593, and rises no further at 96 and 128 (0.576 and 0.566, refitted; Table~\ref{tab:m2b}), so the edit uses $r=64$. The fitted values spread over more directions than that: the first 64 carry 81--85\% of their variance in a block, and 90\% takes 113--153 directions on average. The directions past the first 64 thus carry variance of the fit that the restoration on held-out responses does not use.

\begin{table}[tb]
\centering
\footnotesize
\setlength{\tabcolsep}{0pt}
\begin{tabular*}{\linewidth}{@{}l@{\extracolsep{\fill}}cccccccccc@{}}
\toprule
Rank $r$ & 0 & 1 & 2 & 4 & 8 & 16 & 32 & \textbf{64} & 96 & 128\\
Restor. & 0.075 & 0.296 & 0.379 & 0.467 & 0.522 & 0.552 & 0.575 & \textbf{0.593} & 0.576 & 0.566\\
\bottomrule
\end{tabular*}

\vspace{0.6em}
\begin{tabular*}{\linewidth}{@{}l@{\extracolsep{\fill}}ccccc@{}}
\toprule
Blocks & $k{=}1$ & $k{=}4$ & $k{=}16$ & $k{=}64$ & For 90\%\\
\midrule
1--16 & 0.280 & 0.490 & 0.684 & 0.850 & 114.4\\
17--32 & 0.271 & 0.455 & 0.659 & 0.840 & 122.1\\
33--48 & 0.223 & 0.412 & 0.616 & 0.810 & 153.1\\
49--64 & 0.236 & 0.424 & 0.643 & 0.841 & 112.7\\
\bottomrule
\end{tabular*}
\caption{Rank. Top: restoration on held-out responses against the rank of the fit (96 and 128 refitted). Bottom: share of the variance of the fitted values carried by the first $k$ singular directions of a block, and the number for 90\%, averaged over 16 blocks.}
\label{tab:m2b}
\end{table}

\paragraph{Reduced rank.}
The readouts of Eq.~\eqref{eq:rrr} reach the Eckart--Young bound to four decimals in every group of blocks, as Proposition~3 states (Table~\ref{tab:m5a}). The directions of the regression matrix leave 21--24\% more error, and the best of 300 random sets of readouts 2.0 to 2.2 times as much; the same order holds on the validation fold.

\begin{table}[tb]
\centering
\footnotesize
\setlength{\tabcolsep}{0pt}
\begin{tabular*}{\linewidth}{@{}l@{\extracolsep{\fill}}cccc@{}}
\toprule
Readouts ($r=64$) & 1--16 & 17--32 & 33--48 & 49--64\\
\midrule
\textbf{principal directions of the fit (ours)} & \textbf{0.3825} & \textbf{0.3967} & \textbf{0.4296} & \textbf{0.3879}\\
Eckart--Young bound & 0.3825 & 0.3967 & 0.4296 & 0.3879\\
directions of the regression matrix & 0.4655 & 0.4909 & 0.5314 & 0.4677\\
best of 300 random readouts & 0.8413 & 0.8444 & 0.8635 & 0.8631\\
\bottomrule
\end{tabular*}
\caption{Error relative to the ridge predictions of a rank-64 readout, averaged over 16 blocks, on the fitting responses.}
\label{tab:m5a}
\end{table}

\begin{figure*}[t]
\centering
\includegraphics[width=0.245\textwidth]{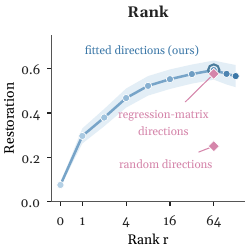}\hspace{0.03\textwidth}\includegraphics[width=0.245\textwidth]{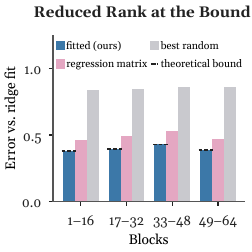}\hspace{0.03\textwidth}\includegraphics[width=0.245\textwidth]{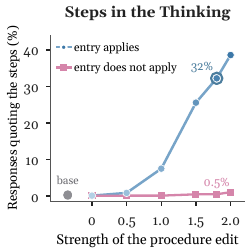}\\[-1pt]
{\small\makebox[0.245\textwidth]{(a)}\hspace{0.03\textwidth}\makebox[0.245\textwidth]{(b)}\hspace{0.03\textwidth}\makebox[0.245\textwidth]{(c)}}
\caption{Rank, the bound of Proposition~3, and the steps in the thinking. (a) Restoration against the rank $r$ of the readout; at $r=64$, the directions of the regression matrix and random directions. (b) Error of the $r$ neurons against the ridge fit for each group of blocks, with the bound. (c) Share of responses whose thinking cites the procedure, by the step of the procedure edit, inside and outside its scope.}
\label{fig:mech_more}
\end{figure*}

\paragraph{Strength.}
The line search of Eq.~\eqref{eq:stop} picks the step 1.8, the one with the lowest held-out divergence (Table~\ref{tab:m5c}). The in-scope gain is largest there as well (+19.0), tied only at 2.6, where the divergence is 1.9 times higher and the out-of-scope change falls to +3.9 points.

\begin{table}[tb]
\centering
\footnotesize
\setlength{\tabcolsep}{0pt}
\begin{tabular*}{\linewidth}{@{}l@{\extracolsep{\fill}}cccc@{}}
\toprule
Step $\gamma$ & Divergence & Restor. & In scope & Out of scope\\
\midrule
0.5 & 0.0681 & 0.238 & $+$15.4 & $+$5.6\\
0.75 & 0.0542 & 0.354 & $+$13.7 & $+$5.5\\
1.0 & 0.0421 & 0.452 & $+$14.7 & $+$5.5\\
1.25 & 0.0329 & 0.527 & $+$16.0 & $+$5.8\\
1.5 & 0.0276 & 0.579 & $+$16.7 & $+$4.5\\
\textbf{1.8 (line search)} & \textbf{0.0271} & \textbf{0.593} & \textbf{$+$19.0} & \textbf{$+$7.0}\\
2.0 & 0.0302 & 0.574 & $+$18.0 & $+$7.2\\
2.6 & 0.0512 & 0.395 & $+$19.0 & $+$3.9\\
\bottomrule
\end{tabular*}
\caption{Step of the procedure edit: mean held-out divergence and restoration, and accuracy change (points) against the base model on in-scope and out-of-scope validation questions.}
\label{tab:m5c}
\end{table}

\begin{table*}[t]
\centering
\renewcommand{\arraystretch}{\FAS}
\scriptsize
\setlength{\tabcolsep}{0pt}
\begin{tabular}{@{}p{0.135\textwidth}*{5}{>{\centering\arraybackslash}p{0.0685\textwidth}}>{\centering\arraybackslash}p{0.09\textwidth}*{5}{>{\centering\arraybackslash}p{0.0685\textwidth}}>{\centering\arraybackslash}p{0.09\textwidth}@{}}
\toprule
 & \shortstack{Baby\\Vision} & BLINK & CharXiv & \shortstack{Visu\\Logic} & \shortstack{VLMs Are\\Biased} & ID & \shortstack{ChartQA\\Pro} & \shortstack{Hallusion\\Bench} & \shortstack{Perception\\Bench} & \shortstack{Spatial\\Viz} & \shortstack{Visual\\Puzzles} & OOD\\
\midrule
run 1 & 48 & 93 & 99 & 46 & 98 & 384 & 86 & 96 & 64 & 79 & 85 & 410\\
run 2 & 50 & 95 & 101 & 40 & 94 & 380 & 87 & 95 & 68 & 81 & 85 & 416\\
run 3 & 52 & 96 & 107 & 46 & 95 & 396 & 88 & 94 & 61 & 81 & 89 & 413\\
mean correct (of 128) & 50.0 & 94.7 & 102.3 & 44.0 & 95.7 & 77.3 & 87.0 & 95.0 & 64.3 & 80.3 & 86.3 & 82.6\\
mean (\%) & 39.1 & 74.0 & 79.9 & 34.4 & 74.7 & 60.4 & 68.0 & 74.2 & 50.3 & 62.8 & 67.4 & 64.5\\
std (points) & 1.6 & 1.2 & 3.3 & 2.7 & 1.6 & 1.3 & 0.8 & 0.8 & 2.7 & 0.9 & 1.8 & 0.5\\
\bottomrule
\end{tabular}
\caption{Base model (Qwen3.8-27B), three runs (sampling seeds 0, 1, 2). ID and OOD per run: correct answers out of 640.}
\label{tab:full_runs_base}
\end{table*}

\begin{table*}[t]
\centering
\renewcommand{\arraystretch}{\FAS}
\scriptsize
\setlength{\tabcolsep}{0pt}
\begin{tabular}{@{}p{0.135\textwidth}*{5}{>{\centering\arraybackslash}p{0.0685\textwidth}}>{\centering\arraybackslash}p{0.09\textwidth}*{5}{>{\centering\arraybackslash}p{0.0685\textwidth}}>{\centering\arraybackslash}p{0.09\textwidth}@{}}
\toprule
 & \shortstack{Baby\\Vision} & BLINK & CharXiv & \shortstack{Visu\\Logic} & \shortstack{VLMs Are\\Biased} & ID & \shortstack{ChartQA\\Pro} & \shortstack{Hallusion\\Bench} & \shortstack{Perception\\Bench} & \shortstack{Spatial\\Viz} & \shortstack{Visual\\Puzzles} & OOD\\
\midrule
run 1 & 75 & 99 & 113 & 55 & 117 & 459 (71.7\%) & 93 & 107 & 72 & 86 & 94 & 452 (70.6\%)\\
run 2 & 81 & 96 & 110 & 60 & 115 & 462 (72.2\%) & 97 & 102 & 79 & 81 & 91 & 450 (70.3\%)\\
run 3 & 77 & 99 & 113 & 57 & 118 & 464 (72.5\%) & 94 & 103 & 76 & 87 & 95 & 455 (71.1\%)\\
sum & 233 & 294 & 336 & 172 & 350 & 1,385 & 284 & 312 & 227 & 254 & 280 & 1,357\\
mean (\%) & 60.7 & 76.6 & 87.5 & 44.8 & 91.1 & 72.1 & 74.0 & 81.3 & 59.1 & 66.1 & 72.9 & 70.7\\
std (points) & 2.4 & 1.4 & 1.4 & 2.0 & 1.2 & 0.4 & 1.6 & 2.1 & 2.7 & 2.5 & 1.6 & 0.4\\
\bottomrule
\end{tabular}
\caption{BaseRSI, correct answers out of 128 in each of three runs (sampling seeds 0, 1, 2), with mean and standard deviation.}
\label{tab:full_runs_ours}
\end{table*}

\begin{table*}[t]
\centering
\renewcommand{\arraystretch}{\FAS}
\scriptsize
\setlength{\tabcolsep}{0pt}
\begin{tabular}{@{}p{0.085\textwidth}p{0.05\textwidth}*{5}{>{\centering\arraybackslash}p{0.0685\textwidth}}>{\centering\arraybackslash}p{0.09\textwidth}*{5}{>{\centering\arraybackslash}p{0.0685\textwidth}}>{\centering\arraybackslash}p{0.09\textwidth}@{}}
\toprule
 &  & \shortstack{Baby\\Vision} & BLINK & CharXiv & \shortstack{Visu\\Logic} & \shortstack{VLMs Are\\Biased} & ID & \shortstack{ChartQA\\Pro} & \shortstack{Hallusion\\Bench} & \shortstack{Perception\\Bench} & \shortstack{Spatial\\Viz} & \shortstack{Visual\\Puzzles} & OOD\\
\midrule
4-shot & run 1 & 44 & 85 & 97 & 40 & 86 & 352 & 82 & 100 & 65 & 75 & 89 & 411\\
 & run 2 & 51 & 90 & 99 & 40 & 89 & 369 & 84 & 102 & 66 & 81 & 91 & 424\\
 & run 3 & 53 & 92 & 96 & 42 & 86 & 369 & 86 & 102 & 69 & 83 & 88 & 428\\
\addlinespace[2pt]
ZoomEye & run 1 & 41 & 91 & 101 & 48 & 80 & 361 & 85 & 93 & 64 & 82 & 85 & 409\\
 & run 2 & 37 & 94 & 94 & 47 & 75 & 347 & 83 & 97 & 64 & 77 & 91 & 412\\
 & run 3 & 42 & 100 & 100 & 49 & 77 & 368 & 85 & 95 & 59 & 81 & 87 & 407\\
\addlinespace[2pt]
PixelProof & run 1 & 55 & 94 & 107 & 48 & 92 & 396 & 90 & 95 & 69 & 77 & 90 & 421\\
 & run 2 & 47 & 100 & 102 & 50 & 96 & 395 & 84 & 99 & 73 & 79 & 86 & 421\\
 & run 3 & 50 & 94 & 105 & 53 & 91 & 393 & 88 & 95 & 68 & 78 & 83 & 412\\
\addlinespace[2pt]
SFT & run 1 & 52 & 92 & 102 & 47 & 95 & 388 & 92 & 98 & 74 & 80 & 78 & 422\\
 & run 2 & 51 & 95 & 99 & 46 & 96 & 387 & 86 & 93 & 68 & 82 & 81 & 410\\
 & run 3 & 48 & 91 & 105 & 49 & 102 & 395 & 86 & 98 & 77 & 81 & 76 & 418\\
\addlinespace[2pt]
GRPO & run 1 & 63 & 87 & 100 & 47 & 99 & 396 & 88 & 97 & 67 & 69 & 83 & 404\\
 & run 2 & 54 & 87 & 101 & 48 & 96 & 386 & 87 & 93 & 71 & 66 & 92 & 409\\
 & run 3 & 58 & 90 & 103 & 42 & 100 & 393 & 85 & 100 & 64 & 73 & 91 & 413\\
\bottomrule
\end{tabular}
\caption{Five baselines of Table~\ref{tab:main}, correct answers out of 128 in three runs (seeds 0, 1, 2); ID and OOD: correct out of 640 per run.}
\label{tab:full_runs_five}
\end{table*}

\begin{table*}[t]
\centering
\renewcommand{\arraystretch}{\FAS}
\footnotesize
\setlength{\tabcolsep}{0pt}
\begin{tabular*}{\textwidth}{@{}l@{\extracolsep{\fill}}ccccccccccccc@{}}
\toprule
Model & \shortstack{Baby\\Vision} & BLINK & CharXiv & \shortstack{Visu\\Logic} & \shortstack{VLMs Are\\Biased} & Avg. & \shortstack{ChartQA\\Pro} & \shortstack{Hallusion\\Bench} & \shortstack{Perception\\Bench} & \shortstack{Spatial\\Viz} & \shortstack{Visual\\Puzzles} & Avg. & $\Delta$ ID / OOD\\
\midrule
Gemma-4-31B & 12.8 & 69.0 & 69.0 & 26.6 & 45.8 & 44.6 & 66.4 & 76.8 & 45.3 & 47.7 & 63.5 & 59.9 & --\\
\ \ + \name{} & 19.3 & 76.8 & 83.1 & 34.6 & 66.4 & 56.0 & 69.0 & 77.9 & 54.2 & 54.9 & 70.1 & 65.2 & $+$11.4 / $+$5.3\\
\midrule
Qwen3.6-27B & 29.2 & 67.7 & 74.2 & 28.6 & 41.9 & 48.3 & 69.0 & 72.1 & 44.0 & 56.3 & 60.9 & 60.5 & --\\
\ \ + \name{} & 45.3 & 74.0 & 81.8 & 39.6 & 50.3 & 58.2 & 71.9 & 75.0 & 49.7 & 58.1 & 67.4 & 64.4 & $+$9.8 / $+$4.0\\
\midrule
Qwen3.5-9B & 22.9 & 65.4 & 69.8 & 38.8 & 36.5 & 46.7 & 64.8 & 75.3 & 40.6 & 44.3 & 58.3 & 56.7 & --\\
\ \ + \name{} & 33.9 & 67.2 & 75.5 & 41.9 & 45.1 & 52.7 & 67.4 & 76.0 & 44.0 & 44.8 & 63.3 & 59.1 & $+$6.0 / $+$2.4\\
\bottomrule
\end{tabular*}
\caption{\name{} on other base models: accuracy (\%), three-run mean. $\Delta$: gain over the unedited model, from unrounded means.}
\label{tab:transfer}
\end{table*}

\begin{table*}[t]
\centering
\renewcommand{\arraystretch}{\FAS}
\scriptsize
\setlength{\tabcolsep}{0pt}
\setlength{\aboverulesep}{0.25ex}
\setlength{\belowrulesep}{0.4ex}
\providecommand{\hd}[2]{\shortstack{#1\\#2}}
\begin{tabular*}{\textwidth}{@{}l@{\extracolsep{\fill}}cccccccccccc@{}}
\toprule
 & \multicolumn{6}{c}{In-distribution (ID)} & \multicolumn{6}{c}{Held-out (OOD)} \\
\cmidrule{2-7}\cmidrule{8-13}
Variant & \hd{Baby}{Vision} & BLINK & CharXiv & \hd{Visu}{Logic} & \hd{VLMs Are}{Biased} & Avg. & \hd{ChartQA}{Pro} & \hd{Hallusion}{Bench} & \hd{Perception}{Bench} & \hd{Spatial}{Viz} & \hd{Visual}{Puzzles} & Avg. \\
\midrule
Qwen3.8-27B (base) & 39.1 & 74.0 & 79.9 & 34.4 & 74.7 & 60.4 & 68.0 & 74.2 & 50.3 & 62.8 & 67.4 & 64.5\\
\name{} (full) & 60.7 & 76.6 & 87.5 & 44.8 & 91.1 & 72.1 & 74.0 & 81.3 & 59.1 & 66.1 & 72.9 & 70.7\\
\midrule
\multicolumn{13}{@{}l}{\textit{Recursive Harness Internalization (Section~\ref{sec:rhc})}}\\
\ \ harness kept in the request & 58.6 & 76.3 & 85.4 & 40.6 & 87.0 & 69.6 & 73.2 & 77.1 & 57.8 & 61.7 & 72.1 & 68.4\\
\ \ procedure edit only & 39.1 & 72.1 & 82.0 & 39.6 & 81.0 & 62.8 & 67.7 & 75.8 & 52.9 & 63.8 & 66.1 & 65.3\\
\ \ view edit only & 58.9 & 75.8 & 84.1 & 43.0 & 78.4 & 68.0 & 73.7 & 79.7 & 54.9 & 63.3 & 72.1 & 68.8\\
\ \ one round & 59.6 & 75.5 & 85.9 & 44.5 & 88.0 & 70.7 & 73.4 & 78.6 & 58.9 & 62.2 & 71.4 & 68.9\\
\ \ two rounds & 61.5 & 75.8 & 85.2 & 45.3 & 88.3 & 71.2 & 73.7 & 77.9 & 58.1 & 63.8 & 72.1 & 69.1\\
\ \ teacher's responses only & 60.2 & 74.7 & 85.9 & 44.0 & 89.3 & 70.8 & 72.7 & 79.9 & 58.1 & 63.0 & 68.0 & 68.3\\
\ \ base model's responses only & 58.1 & 77.1 & 86.5 & 40.9 & 90.1 & 70.5 & 72.7 & 81.3 & 56.8 & 59.9 & 68.8 & 67.9\\
\ \ macro average only & 59.6 & 75.3 & 85.9 & 44.3 & 90.6 & 71.1 & 72.9 & 79.2 & 59.4 & 61.7 & 70.6 & 68.8\\
\midrule
\multicolumn{13}{@{}l}{\textit{ReGap (Section~\ref{sec:regap})}}\\
\ \ output only (distillation) & 62.2 & 75.8 & 83.3 & 42.4 & 80.7 & 68.9 & 72.9 & 78.1 & 55.2 & 65.1 & 68.0 & 67.9\\
\ \ last block only & 59.4 & 75.8 & 84.6 & 43.8 & 78.9 & 68.5 & 72.9 & 77.6 & 55.5 & 61.2 & 70.6 & 67.6\\
\ \ last 16 blocks & 58.6 & 77.3 & 85.9 & 42.4 & 80.2 & 68.9 & 72.4 & 75.5 & 56.3 & 62.2 & 70.6 & 67.4\\
\ \ first 32 blocks & 58.9 & 79.2 & 84.9 & 42.2 & 82.6 & 69.5 & 72.9 & 77.3 & 57.3 & 62.8 & 70.6 & 68.2\\
\ \ 48 linear-attention blocks & 60.9 & 75.5 & 84.6 & 41.9 & 84.1 & 69.4 & 73.2 & 77.9 & 58.9 & 63.8 & 70.3 & 68.8\\
\ \ 16 full-attention blocks & 60.4 & 75.5 & 84.6 & 43.2 & 86.2 & 70.0 & 72.7 & 77.1 & 54.9 & 62.5 & 68.5 & 67.1\\
\ \ direct difference as signal & 59.6 & 74.7 & 82.3 & 40.9 & 89.3 & 69.4 & 73.4 & 80.5 & 54.7 & 64.3 & 71.6 & 68.9\\
\ \ last-block state & 56.8 & 76.6 & 86.2 & 43.5 & 79.9 & 68.6 & 73.4 & 76.0 & 54.4 & 64.8 & 69.8 & 67.7\\
\ \ tail + 64 positions & 60.7 & 75.0 & 85.2 & 42.7 & 89.8 & 70.7 & 73.2 & 77.1 & 56.5 & 63.0 & 72.1 & 68.4\\
\ \ 16 responses & 60.9 & 76.6 & 84.6 & 45.6 & 86.5 & 70.8 & 72.9 & 75.8 & 57.6 & 63.8 & 69.8 & 68.0\\
\ \ 64 responses & 57.0 & 76.0 & 85.9 & 43.2 & 89.8 & 70.4 & 73.2 & 81.8 & 57.6 & 63.8 & 71.1 & 69.5\\
\ \ random readouts & 60.7 & 75.8 & 86.7 & 41.4 & 75.8 & 68.1 & 72.7 & 76.6 & 56.8 & 61.7 & 72.9 & 68.1\\
\ \ random values & 60.2 & 75.8 & 84.1 & 41.4 & 77.9 & 67.9 & 73.4 & 77.3 & 58.3 & 63.5 & 72.9 & 69.1\\
\midrule
\multicolumn{13}{@{}l}{\textit{ReKey (Section~\ref{sec:rekey})}}\\
\ \ mean gap only (rank 0) & 60.2 & 76.0 & 84.4 & 42.4 & 79.7 & 68.5 & 74.0 & 78.4 & 57.0 & 63.5 & 72.7 & 69.1\\
\ \ rank 1 & 57.6 & 77.1 & 83.6 & 44.5 & 82.6 & 69.1 & 72.1 & 78.1 & 57.3 & 62.0 & 71.9 & 68.3\\
\ \ rank 4 & 60.4 & 77.9 & 83.6 & 45.8 & 85.2 & 70.6 & 74.5 & 76.6 & 56.5 & 60.9 & 70.6 & 67.8\\
\ \ rank 8 & 59.4 & 75.8 & 84.4 & 42.7 & 84.9 & 69.4 & 73.7 & 77.3 & 56.0 & 62.0 & 71.4 & 68.1\\
\ \ regression-matrix directions & 60.9 & 76.3 & 83.3 & 42.7 & 89.1 & 70.5 & 73.4 & 77.9 & 56.8 & 63.5 & 68.2 & 68.0\\
\ \ random directions & 58.3 & 76.8 & 83.6 & 42.7 & 83.9 & 69.1 & 72.9 & 79.2 & 59.6 & 63.3 & 71.4 & 69.3\\
\ \ random neurons & 60.4 & 75.0 & 85.4 & 44.3 & 89.8 & 71.0 & 71.9 & 78.1 & 55.7 & 61.7 & 71.9 & 67.9\\
\ \ most active neurons & 49.7 & 71.4 & 82.3 & 38.5 & 83.6 & 65.1 & 71.9 & 75.5 & 55.2 & 52.6 & 71.4 & 65.3\\
\ \ constant key & 60.4 & 76.3 & 85.4 & 43.2 & 89.8 & 71.0 & 71.9 & 80.7 & 56.5 & 63.3 & 72.7 & 69.0\\
\ \ site preset (view in MLP) & 38.3 & 73.4 & 81.3 & 38.0 & 81.8 & 62.6 & 67.7 & 76.8 & 52.6 & 61.2 & 67.7 & 65.2\\
\ \ step 0.5 & 59.4 & 75.3 & 85.7 & 44.8 & 80.2 & 69.1 & 72.9 & 80.7 & 58.1 & 62.8 & 73.4 & 69.6\\
\ \ strength preset (step 1.0) & 59.4 & 74.0 & 84.1 & 42.4 & 82.3 & 68.4 & 72.9 & 78.6 & 56.3 & 63.8 & 71.4 & 68.6\\
\ \ step 1.5 & 59.4 & 76.0 & 85.9 & 42.2 & 86.7 & 70.1 & 74.0 & 78.6 & 59.1 & 59.6 & 70.1 & 68.3\\
\ \ step 2.0 & 59.6 & 76.6 & 84.9 & 45.3 & 92.2 & 71.7 & 73.4 & 80.2 & 57.0 & 63.8 & 71.1 & 69.1\\
\ \ step 2.6 & 60.7 & 77.3 & 85.2 & 43.8 & 91.4 & 71.7 & 73.4 & 79.4 & 55.7 & 62.0 & 70.6 & 68.2\\
\bottomrule
\end{tabular*}
\caption{Ablation variants, accuracy (\%) on each benchmark, mean over three runs.}
\label{tab:ablation_full}
\end{table*}

\begin{table*}[t]
\centering
\renewcommand{\arraystretch}{\FAS}
\scriptsize
\setlength{\tabcolsep}{0pt}
\begin{tabular*}{\textwidth}{@{}l@{\extracolsep{\fill}}cccccccc@{}}
\toprule
Setting & Restoration & In scope $\Delta$ & Out of scope $\Delta$ & In scope (of 102) & Out of scope (of 268) & ID & OOD & Runs\\
\midrule
base model & -- & -- & -- & 72.3 & 147.3 & 77.3 & 82.6 & 3 / 3\\
one round & 0.583 $\pm$ 0.044 & $+$16.3 & $+$4.4 & 89.0 & 159.0 & 90.5 & 88.2 & 3 / 3\\
two rounds & 0.590 $\pm$ 0.043 & $+$17.6 & $+$6.0 & 90.3 & 163.3 & 91.1 & 88.5 & 3 / 3\\
\name{} (three rounds) & 0.593 $\pm$ 0.043 & $+$19.0 & $+$7.0 & 91.7 & 166.0 & 92.3 & 90.5 & 3 / 3\\
base model's responses only & 0.515 $\pm$ 0.038 & $+$17.3 & $+$5.5 & 90.0 & 162.0 & 90.3 & 86.9 & 3 / 3\\
macro average only (+ one entry) & 0.545 $\pm$ 0.049 & $+$16.7 & $+$6.6 & 89.3 & 165.0 & 91.1 & 88.0 & 3 / 3\\
\bottomrule
\end{tabular*}
\caption{Ablations of Recursive Harness Internalization in full. Restoration $\pm$ standard error on 28 held-out responses; $\Delta$ in points against the base model on 102 / 268 validation questions; ID and OOD: mean correct answers out of 128; runs: test runs / validation runs.}
\label{tab:full_a2}
\end{table*}

\begin{table*}[t]
\centering
\renewcommand{\arraystretch}{\FAS}
\scriptsize
\setlength{\tabcolsep}{0pt}
\begin{tabular*}{\textwidth}{@{}l@{\extracolsep{\fill}}ccccccc@{}}
\toprule
Rounds & 1 & 2 & \textbf{3 (\name{})} & 4 & 5 & 6 & 7\\
\midrule
Restoration & 0.583 & 0.590 & \textbf{0.593} & 0.595 & 0.596 & 0.595 & 0.594\\
Change & -- & $+$0.007 $\pm$ 0.005 & \textbf{$+$0.003 $\pm$ 0.003} & $+$0.002 $\pm$ 0.002 & $+$0.001 $\pm$ 0.002 & $-$0.001 $\pm$ 0.003 & $-$0.001 $\pm$ 0.002\\
ID & 90.5 & 91.1 & \textbf{92.3} & 92.5 & 92.7 & 92.3 & 93.0\\
OOD & 88.2 & 88.5 & \textbf{90.5} & 89.2 & 88.4 & 90.1 & 89.2\\
\bottomrule
\end{tabular*}
\caption{Rounds of the inner loop, each built as a full model. Restoration on the 28 held-out responses; change: paired difference from the previous round $\pm$ standard error; ID and OOD: mean correct answers out of 128 over three runs.}
\label{tab:full_rounds}
\end{table*}

\begin{table*}[t]
\centering
\renewcommand{\arraystretch}{\FAS}
\scriptsize
\setlength{\tabcolsep}{0pt}
\begin{tabular*}{\textwidth}{@{}l@{\extracolsep{\fill}}cccccccc@{}}
\toprule
Setting & Restoration & In scope $\Delta$ & Out of scope $\Delta$ & In scope (of 102) & Out of scope (of 268) & ID & OOD & Runs\\
\midrule
base model & -- & -- & -- & 72.3 & 147.3 & 77.3 & 82.6 & 3 / 3\\
\name{} (three rounds) & 0.593 $\pm$ 0.043 & $+$19.0 & $+$7.0 & 91.7 & 166.0 & 92.3 & 90.5 & 3 / 3\\
blocks: last block only (65 neurons) & 0.005 $\pm$ 0.003 & $+$14.4 & $+$4.1 & 87.0 & 158.3 & 87.7 & 86.5 & 3 / 3\\
blocks: last 16 (1,040 neurons) & 0.304 $\pm$ 0.032 & $+$14.1 & $+$5.5 & 86.7 & 162.0 & 88.2 & 86.3 & 3 / 3\\
blocks: first 32 (2,080 neurons) & 0.055 $\pm$ 0.022 & $+$16.7 & $+$5.1 & 89.3 & 161.0 & 89.0 & 87.3 & 3 / 3\\
blocks: 48 linear-attention (3,120 neurons) & 0.093 $\pm$ 0.017 & $+$14.4 & $+$6.1 & 87.0 & 163.7 & 88.9 & 88.1 & 3 / 3\\
blocks: 16 full-attention (1,040 neurons) & 0.580 $\pm$ 0.046 & $+$14.7 & $+$5.2 & 87.3 & 161.3 & 89.6 & 85.9 & 3 / 3\\
signal: last-block state (step 1.0) & 0.271 $\pm$ 0.032 & $+$12.1 & $+$5.3 & 84.7 & 161.7 & 87.8 & 86.7 & 3 / 3\\
positions: tail + 64 & 0.575 $\pm$ 0.043 & $+$16.3 & $+$4.6 & 89.0 & 159.7 & 90.5 & 87.5 & 3 / 3\\
responses: 16 & 0.541 $\pm$ 0.043 & $+$17.3 & $+$4.6 & 90.0 & 159.7 & 90.7 & 87.0 & 3 / 3\\
responses: 64 & 0.578 $\pm$ 0.044 & $+$19.0 & $+$6.7 & 91.7 & 165.3 & 90.1 & 88.9 & 3 / 3\\
\bottomrule
\end{tabular*}
\caption{ReGap ablations in full; columns as in Table~\ref{tab:full_a2}.}
\label{tab:full_a3}
\end{table*}

\begin{table*}[t]
\centering
\renewcommand{\arraystretch}{\FAS}
\scriptsize
\setlength{\tabcolsep}{0pt}
\begin{tabular*}{\textwidth}{@{}l@{\extracolsep{\fill}}ccccccc@{}}
\toprule
Block & Removed & Block & Removed & Block & Removed & Block & Removed\\
\midrule
1 & $+$0.0023 & 17 & $+$0.0016 & 33 & $+$0.0142 & 49 & $+$0.0015\\
2 & $+$0.0014 & 18 & $+$0.0011 & 34 & $+$0.0063 & 50 & $+$0.0141\\
3 & $-$0.0014 & 19 & $+$0.0028 & 35 & $+$0.0145 & 51 & $+$0.0026\\
4$^{*}$ & $+$0.0043 & 20$^{*}$ & $+$0.0208 & 36$^{*}$ & $+$0.1280 & 52$^{*}$ & $+$0.2316\\
5 & $+$0.0019 & 21 & $+$0.0009 & 37 & $+$0.0046 & 53 & $+$0.0031\\
6 & $+$0.0021 & 22 & $+$0.0051 & 38 & $+$0.0030 & 54 & $+$0.0008\\
7 & $+$0.0004 & 23 & $-$0.0021 & 39 & $+$0.0008 & 55 & $+$0.0010\\
8$^{*}$ & $+$0.0037 & 24$^{*}$ & $+$0.0156 & 40$^{*}$ & $+$0.0825 & 56$^{*}$ & $+$0.2557\\
9 & $+$0.0018 & 25 & $-$0.0000 & 41 & $-$0.0009 & 57 & $+$0.0006\\
10 & $+$0.0002 & 26 & $+$0.0033 & 42 & $+$0.0002 & 58 & $+$0.0029\\
11 & $-$0.0005 & 27 & $-$0.0018 & 43 & $-$0.0028 & 59 & $+$0.0022\\
12$^{*}$ & $+$0.0046 & 28$^{*}$ & $+$0.0140 & 44$^{*}$ & $+$0.0998 & 60$^{*}$ & $+$0.2045\\
13 & $+$0.0011 & 29 & $+$0.0003 & 45 & $+$0.0082 & 61 & $+$0.0050\\
14 & $-$0.0007 & 30 & $+$0.0008 & 46 & $+$0.0012 & 62 & $+$0.0016\\
15 & $+$0.0009 & 31 & $+$0.0031 & 47 & $+$0.0033 & 63 & $+$0.0035\\
16$^{*}$ & $+$0.0121 & 32$^{*}$ & $+$0.0299 & 48$^{*}$ & $+$0.1184 & 64$^{*}$ & $+$0.0166\\
\bottomrule
\end{tabular*}
\caption{Divergence removed by the gap of block $l$ alone ($*$: full attention). They sum to 1.358; together they remove 0.997.}
\label{tab:full_m4single}
\end{table*}

\begin{table*}[t]
\centering
\renewcommand{\arraystretch}{\FAS}
\scriptsize
\setlength{\tabcolsep}{0pt}
\begin{tabular*}{\textwidth}{@{}l@{\extracolsep{\fill}}cccccccc@{}}
\toprule
Setting & Restoration & In scope $\Delta$ & Out of scope $\Delta$ & In scope (of 102) & Out of scope (of 268) & ID & OOD & Runs\\
\midrule
base model & -- & -- & -- & 72.3 & 147.3 & 77.3 & 82.6 & 3 / 3\\
\name{} (three rounds) & 0.593 $\pm$ 0.043 & $+$19.0 & $+$7.0 & 91.7 & 166.0 & 92.3 & 90.5 & 3 / 3\\
mean gap only (rank 0) & 0.075 $\pm$ 0.008 & $+$14.1 & $+$5.8 & 86.7 & 163.0 & 87.7 & 88.5 & 3 / 3\\
rank 1 & 0.296 $\pm$ 0.034 & $+$16.3 & $+$6.1 & 89.0 & 163.7 & 88.4 & 87.4 & 3 / 3\\
rank 4 & 0.467 $\pm$ 0.043 & $+$17.0 & $+$3.6 & 89.7 & 157.0 & 90.3 & 86.8 & 3 / 3\\
rank 8 & 0.522 $\pm$ 0.044 & $+$16.7 & $+$6.3 & 89.3 & 164.3 & 88.9 & 87.1 & 3 / 3\\
reduction: regression-matrix directions & 0.576 $\pm$ 0.045 & $+$16.7 & $+$4.9 & 89.3 & 160.3 & 90.2 & 87.0 & 3 / 3\\
reduction: random directions & 0.250 $\pm$ 0.025 & $+$16.0 & $+$6.2 & 88.7 & 164.0 & 88.4 & 88.7 & 3 / 3\\
neurons: random 65 per block & 0.474 $\pm$ 0.062 & $+$17.3 & $+$5.2 & 90.0 & 161.3 & 90.9 & 86.9 & 3 / 3\\
neurons: 65 most active & $-$2.119 $\pm$ 0.742 & $+$13.7 & $+$1.0 & 86.3 & 150.0 & 83.3 & 83.6 & 3 / 3\\
site preset: view in MLP neurons & -- & $+$11.8 & $-$0.6 & 84.3 & 145.7 & 80.1 & 83.5 & 3 / 3\\
step 0.5 & 0.238 $\pm$ 0.026 & $+$15.4 & $+$5.6 & 88.0 & 162.3 & 88.4 & 89.1 & 3 / 3\\
strength preset: step 1.0 & 0.452 $\pm$ 0.040 & $+$14.7 & $+$5.5 & 87.3 & 162.0 & 87.6 & 87.8 & 3 / 3\\
step 1.5 & 0.579 $\pm$ 0.046 & $+$16.7 & $+$4.5 & 89.3 & 159.3 & 89.7 & 87.4 & 3 / 3\\
step 2.0 & 0.574 $\pm$ 0.039 & $+$18.0 & $+$7.2 & 90.7 & 166.7 & 91.8 & 88.5 & 3 / 3\\
step 2.6 & 0.395 $\pm$ 0.040 & $+$19.0 & $+$3.9 & 91.7 & 157.7 & 91.7 & 87.3 & 3 / 3\\
\bottomrule
\end{tabular*}
\caption{ReKey ablations in full; columns as in Table~\ref{tab:full_a2}.}
\label{tab:full_a4}
\end{table*}

\paragraph{Size of the edit.}
The procedure edit rewrites 65 neurons in each of the 64 blocks, 4,160 in all, each with three vectors of 5,120 entries: 63,897,600 weights, 0.230\% of the 27.78 billion weights of the model and 0.249\% of the 25.62 billion of its language model. Every other weight of the language model is bit-identical to the base model, so writing back the original rows of the 4,160 neurons restores it exactly; the patch-embedding edit rewrites its own tensor, which writing back the original tensor restores.

\paragraph{Rank, bound, and the steps in the thinking.}
Restoration rises with the rank of the readout and levels off near $r=64$, where the principal directions of the fit restore 0.593, the directions of the regression matrix 0.576, and random directions 0.250 (Figure~\ref{fig:mech_more}a).
In every group of blocks, the $r$ neurons of the fit reach the bound of Proposition~3, with errors of 0.38 to 0.43 against the ridge fit, while the regression-matrix directions leave 0.47 to 0.53 and the best random directions 0.84 to 0.86 (Figure~\ref{fig:mech_more}b).
The procedure also appears in the thinking of the edited model, whose request no longer carries it: at the deployed step 1.8, 32\% of the responses inside the scope of the entry cite it by a fixed pattern, against 0.5\% outside, and the share inside grows with the step while that outside stays near zero (Figure~\ref{fig:mech_more}c).

\Needspace{4.5\baselineskip}
\section{Complete Data}\label{app:data}
This appendix gives the rest of the measured data, each table with the experiment it comes from and how to read it.
Unless a table states otherwise, a test score is the mean over three runs with sampling seeds 0, 1, and 2, the validation counts are correct answers on the 102 in-scope and 268 out-of-scope validation questions, and restoration is measured on 28 held-out responses of the student.

\begin{figure*}[t]
\centering
\includegraphics[width=\textwidth]{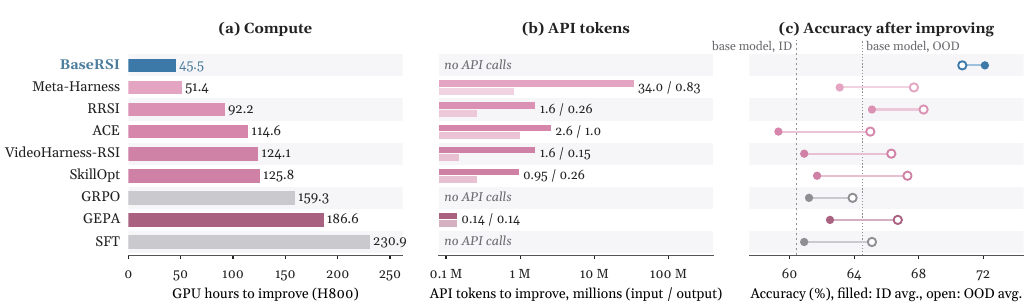}
\caption{Cost of improving the base model against accuracy. (a) H800 GPU hours. (b) API tokens spent while improving. (c) ID and OOD averages after improving (Table~\ref{tab:main}); dashed: the base model. Blue: \name{}; pink: harness methods; grey: fine-tuning.}
\label{fig:cost}
\end{figure*}

\paragraph{Runs of the base model and \name{}.}
Tables~\ref{tab:full_runs_base} and~\ref{tab:full_runs_ours} give the three test runs behind the base model and \name{} in Table~\ref{tab:main}.
Run $k$ of either model samples every test question once with seed $k$, so the two models answer the same questions under the same sampling noise, and the test of Table~\ref{tab:sig} pairs them run by run.
On the 640 ID questions the base model answers 384, 380, and 396 right and \name{} 459, 462, and 464, a gain of 68 to 82 questions in every pair; on the 640 OOD questions the gains are 42, 34, and 42 questions.
Per benchmark the standard deviation of \name{} ranges from 1.2 points on VLMs Are Biased to 2.7 on PerceptionBench, and its two averages vary by 0.4 points.

\paragraph{Runs of five baselines.}
Table~\ref{tab:full_runs_five} gives the three runs behind five baselines of Table~\ref{tab:main}, with the same seeds as the base model and \name{}.
Their standard deviations range from 0.8 to 3.9 points per benchmark and from 0.2 to 1.7 points on the ID and OOD averages, within the range of the other methods.
The best run of any of them answers 396 of the 640 ID and 428 of the 640 OOD questions right, 63 and 22 below the worst run of \name{} (459 and 450).

\paragraph{Other base models.}
Table~\ref{tab:transfer} runs the loop of \name{} on three other vision-language models and tests each of them on the same ten benchmarks over three runs. The loop starts from the original weights of each model, and the entries that it accepts and internalizes differ in part from those of Qwen3.8-27B.
It raises the ID average of every model, by 11.4, 9.8, and 6.0 points on Gemma-4-31B, Qwen3.6-27B, and Qwen3.5-9B, and the OOD average by 5.3, 4.0, and 2.4, and each of the thirty benchmark scores rises.
The gain is largest on BabyVision for the two Qwen models (16.1 and 11.0 points) and on VLMs Are Biased for Gemma-4-31B (20.6 points), the two benchmarks on which it is also the largest on Qwen3.8-27B (Table~\ref{tab:main}).

\paragraph{Ablations per benchmark.}
Table~\ref{tab:ablation_full} gives the test accuracy on each benchmark for the variants of Table~\ref{tab:ablation} and Figure~\ref{fig:mech}a, together with the sweeps over the rank and the step.
The edits act on different benchmarks: without the view edit BabyVision falls most, from 60.7 to 39.1, and without the procedure edit VLMs Are Biased falls most, from 91.1 to 78.4, while BabyVision stays at 58.9.
Most variants of the procedure edit therefore differ on VLMs Are Biased, where the counting entry applies: the last-block edit lowers it to 78.9 and the mean gap alone to 79.7, while BabyVision moves by less than 1.5 points.
Writing the most active neurons lowers the benchmarks outside the scope of the entry the most (BabyVision 49.7, SpatialViz 52.6), since it overwrites neurons that the base model uses on every question, and writing the view entry into MLP neurons loses the gain on BabyVision (38.3 against 60.7).

\paragraph{Ablations of Recursive Harness Internalization in full.}
Tables~\ref{tab:full_a2} to~\ref{tab:full_a4} extend Figure~\ref{fig:mech}a with the standard error of the restoration, the correct answers on the validation questions, and the test scores as correct answers out of 128.
Restoration measures how much of the teacher's next-token distribution an edit reproduces on held-out responses, the validation counts how many answers it changes, and the test scores whether the change carries to new questions.
In Table~\ref{tab:full_a2}, each added round raises the restoration (0.583, 0.590, 0.593) and the in-scope gain ($+$16.3, $+$17.6, $+$19.0).
Further rounds (Table~\ref{tab:full_rounds}) change the restoration by at most 0.002 each and the ID accuracy by at most 0.7 correct answers.
Fitting on base-model responses alone restores 0.515, with 90.3 and 86.9 on the tests against 92.3 and 90.5.

\paragraph{ReGap ablations in full.}
In Table~\ref{tab:full_a3}, an edit of a subset of blocks restores little unless it contains the full-attention blocks of the second half: the last block alone restores 0.005, the first 32 blocks 0.055, the 48 linear-attention blocks 0.093, and the last 16 blocks 0.304, while the 16 full-attention blocks restore 0.580.
Even this subset gains less than all 64 blocks, in scope ($+$14.7 against $+$19.0) and on the OOD tests (85.9 against 90.5 correct); Appendix~\ref{app:mech} repeats the comparison with all 4,160 neurons.
The 48 linear-attention blocks hold three times the neurons of the 16 full-attention blocks and restore 0.093 against 0.580, and ranks 96 and 128 add neurons without raising the restoration (Table~\ref{tab:m2b}), so the placement of an edit, not its size, decides what it restores.
The target matters as much as the site: the difference of the last-block states restores 0.271.
Adding 64 reasoning positions to the template tail restores 0.575, close to the 0.593 of \name{}, and 16 or 64 replayed responses per round restore 0.541 and 0.578.

\paragraph{Each block alone.}
Table~\ref{tab:full_m4single} adds the gap of one block at a time to the student and gives the fraction of the divergence that it removes on the held-out responses; Table~\ref{tab:m4a} adds the gaps cumulatively.
The full-attention blocks of the second half dominate: blocks 52, 56, and 60 remove 0.23, 0.26, and 0.20 each, blocks 36 to 48 between 0.08 and 0.13, and most linear-attention blocks less than 0.01.
The removals sum to 1.358, more than the 0.997 that all gaps remove together, since the divergence is not additive over blocks, and no subset in Table~\ref{tab:full_a3} reaches the restoration of all 64.

\paragraph{ReKey ablations in full.}
In Table~\ref{tab:full_a4}, the restoration rises with the rank from 0.075 for the mean gap alone to 0.522 at rank 8 and 0.593 at rank 64.
At rank 64, readouts along the directions of the regression matrix restore 0.576 and random directions 0.250.
Writing the pairs into 65 random neurons per block restores 0.474, and writing them into the 65 most active neurons raises the divergence ($-$2.119) and keeps only $+$1.0 out of scope, since those neurons carry what the base model computes on every request.
Writing all entries into MLP neurons instead of the patch embedding keeps $+$11.8 in scope and loses the gain outside ($-$0.6).

\paragraph{Cost against accuracy.}
Figure~\ref{fig:cost} sets the cost of improving the base model (Table~\ref{tab:cost}) beside the accuracy that each method reaches (Table~\ref{tab:main}).
\name{} is the cheapest and the most accurate: 45.5 GPU hours for 72.1 ID and 70.7 OOD, against 51.4 hours and 34.0\,M API input tokens for Meta-Harness, the next cheapest, which reaches 63.1 and 67.7.
More compute does not buy accuracy among the other methods: GEPA spends 186.6 hours for 62.5 ID and SFT 230.9 for 60.9, and SFT and GRPO stay within 0.8 points of the base model in distribution.

\FloatBarrier
\setlength{\multicolsep}{0pt}%
\Needspace{4.5\baselineskip}
\section{Case Study}\label{app:case}
\begin{table*}[t]
\centering
\small
\setlength{\tabcolsep}{0pt}
\renewcommand{\arraystretch}{1.25}
\begin{tabular*}{\textwidth}{@{}l@{\extracolsep{\fill}}ccccccccc@{}}
\toprule
 & & \multicolumn{4}{c}{VLMs Are Biased} & \multicolumn{2}{c}{BabyVision} & \shortstack{Perception\\Bench} & \shortstack{ChartQA\\Pro}\\
\cmidrule(lr){3-6}\cmidrule(lr){7-8}\cmidrule(lr){9-9}\cmidrule(lr){10-10}
Setting & Runs & Rings & Stripes & Bird & Wolf & Squares & Odd one & Lights & Chart\\
 & & Fig.~\ref{fig:case_rings} & Fig.~\ref{fig:case_stripes} & Fig.~\ref{fig:case_rhea} & Fig.~\ref{fig:case_wolf} & Fig.~\ref{fig:case_squares} & Fig.~\ref{fig:case_oddone} & Fig.~\ref{fig:case_lights} & Fig.~\ref{fig:case_chart}\\
\midrule
Base model & 10 & 0 & 2 & 1 & 0 & 0 & 0 & 2 & 0\\
View edit alone & 5 & 0 & 3 & 5 & 5 & 5 & 4 & 4 & 5\\
Procedure edit alone & 3 & 2 & 2 & 2 & 0 & 0 & 0 & 0 & 0\\
\midrule
Both edits, step 0.5 & 3 & 0 & 2 & 3 & 2 & 3 & 3 & 3 & 3\\
Both edits, step 1.0 & 3 & 0 & 2 & 3 & 3 & 3 & 3 & 2 & 2\\
Both edits, step 1.5 & 3 & 2 & 3 & 3 & 3 & 3 & 3 & 3 & 3\\
Both edits, step 2.0 & 3 & 3 & 3 & 3 & 3 & 3 & 3 & 2 & 3\\
Procedure edit of round 1 & 3 & 2 & 2 & 3 & 3 & 3 & 3 & 3 & 3\\
Procedure edit of rounds 1--2 & 3 & 3 & 2 & 3 & 3 & 3 & 3 & 3 & 2\\
\midrule
\name{} (step 1.8) & 7 & \textbf{6} & \textbf{7} & \textbf{7} & \textbf{7} & \textbf{7} & \textbf{7} & \textbf{7} & \textbf{7}\\
Up-scaled image in the request & 5 & 0 & 1 & 5 & 4 & 5 & 2 & 4 & 4\\
\midrule
Other methods, same base model & 31 & 3 & 8 & 7 & 6 & 0 & 0 & 4 & 0\\
Closed-source models & 21 & 9 & 4 & 0 & 3 & 5 & 4 & 2 & 3\\
\bottomrule
\end{tabular*}
\caption{Runs answered right on the eight case questions, by setting, out of the number of runs given. The edits alone, the steps, and \name{} use the same base model; the up-scaled image is enlarged in the request with the weights unchanged; the closed-source models run three times at each of seven effort levels. Lights and Chart come from held-out benchmarks.}
\label{tab:cases}
\end{table*}

We take the cases from the 1,280 test questions.
On 38 of them the base model is wrong in all three runs with seeds 0 to 2 and \name{} right in all three, against 6 questions the other way, a split that sampling noise alone would make even; among these we keep the ones that the other methods on the same base model mostly miss, and add questions on which the base model is right in at most two of ten runs.
Some of these answers turn on the finer view and others on the procedure; we show both kinds, eight questions from four benchmarks, two of them held out (Figures~\ref{fig:case_rings} to~\ref{fig:case_chart}; Table~\ref{tab:cases} gathers their runs by setting).
The two edits repair different failures: the view edit lets the model see what the question asks about, and the procedure edit, fitted by ReGap and written by ReKey, makes it count what it sees in place of the value it expects.
On six questions the answer turns on the finer view, with the view edit alone right in at least four of five runs; on the rings it turns on the procedure, with the procedure edit alone right in 2 of 3 runs and the view edit alone in none; on the stripes both edits contribute.
Together the edits are right more often than either alone, in 55 of 56 runs against 31 of 40 for the view edit and 6 of 24 for the procedure edit, as on the benchmarks (Figure~\ref{fig:analysis}b).
The view edit is also right more often than the up-scaled image in the request, in 31 against 25 of 40 runs: as often on four questions and more often on the other four.
The share of right runs rises with the step of the procedure edit, from 19 and 18 of 24 at steps 0.5 and 1.0 to 23 of 24 at step 1.5 and 55 of 56 at the step 1.8 that the line search of ReKey selects; the procedure edits of round 1 and of rounds 1 and 2 are right in 22 of 24 runs each.
The two held-out questions follow the same pattern, the view edit alone right in four and five of five runs and \name{} in all seven.
The base model is right in 5 of 80 runs, the other methods in 28 of 248, at most 8 of their 31 on one question, and the four closed-source models in 30 of 168 over seven effort levels, at most 9 of their 21.
Appendix~\ref{app:mech} and Table~\ref{tab:ablation} give the controls: shuffled gaps or gaps of another block leave the divergence, null edits leave the ID average unchanged, and no preset site or strength reaches the measured one.

Every figure has the same layout.
The question shows the image with the region that decides the answer marked in red.
\emph{What the model sees} draws the image tokens of that region as a grid, for the base model and after the view edit, each cell being one token that the model reads.
\emph{Becoming right step by step} gives the share of runs answered right as the edits are added, where Edit~1 is the view edit and Edit~2 the procedure edit at the step written below it.
\emph{Answers of every run} lists each run of each method on the same base model and of four closed-source models, a filled square for a right answer and a crossed square for a wrong one, with the typical answer of each method on the right.
The bottom row quotes the thinking of single runs, with [\dots] marking omitted text, and the excerpts below each case are taken from the same runs.
The eleven other methods, chain-of-thought, 4-shot examples, Self-Refine, ZoomEye, two-stage zoom-in, PixelProof, GEPA, ACE, SkillOpt, VideoHarness-RSI, and RRSI, give 31 runs in all.
The closed-source models give three runs at each of seven effort levels, Claude Haiku 5.5 at medium, Claude Sonnet 5.5 at medium and high, GPT-6 Luna at low, medium, and its highest effort, and GPT-6.1 Terra at medium, scored as in the main tables; the figures show the highest level of each, and the text counts all 21 runs.
Self-consistency appears in the figures only.

One observation recurs across the counting questions.
With the procedure edit, the thinking of the model often cites an instruction to trust direct observation rather than the typical value, although the request carries no such text: ReGap and ReKey have written the entry into the weights, and the model recalls it as if it were part of the request.
Over all replies to 21 candidate questions, a fixed pattern finds such a citation in 46 of the 147 replies of \name{}, in 0, 4, 16, and 21 of 63 replies at steps 0.5, 1.0, 1.5, and 2.0, and in none of the 42 replies of the base model, the 63 of the view edit, or the 109 of chain-of-thought, 4-shot, and ZoomEye prompting.
The citation grows with the step and is absent without the edit, so it comes from the edit, and it appears on counting questions only, never on the chart or odd-one-out questions, matching the scope that the key carries.

\begin{table*}[tp]
\centering
\vspace{-5pt}%
\includegraphics[width=\textwidth]{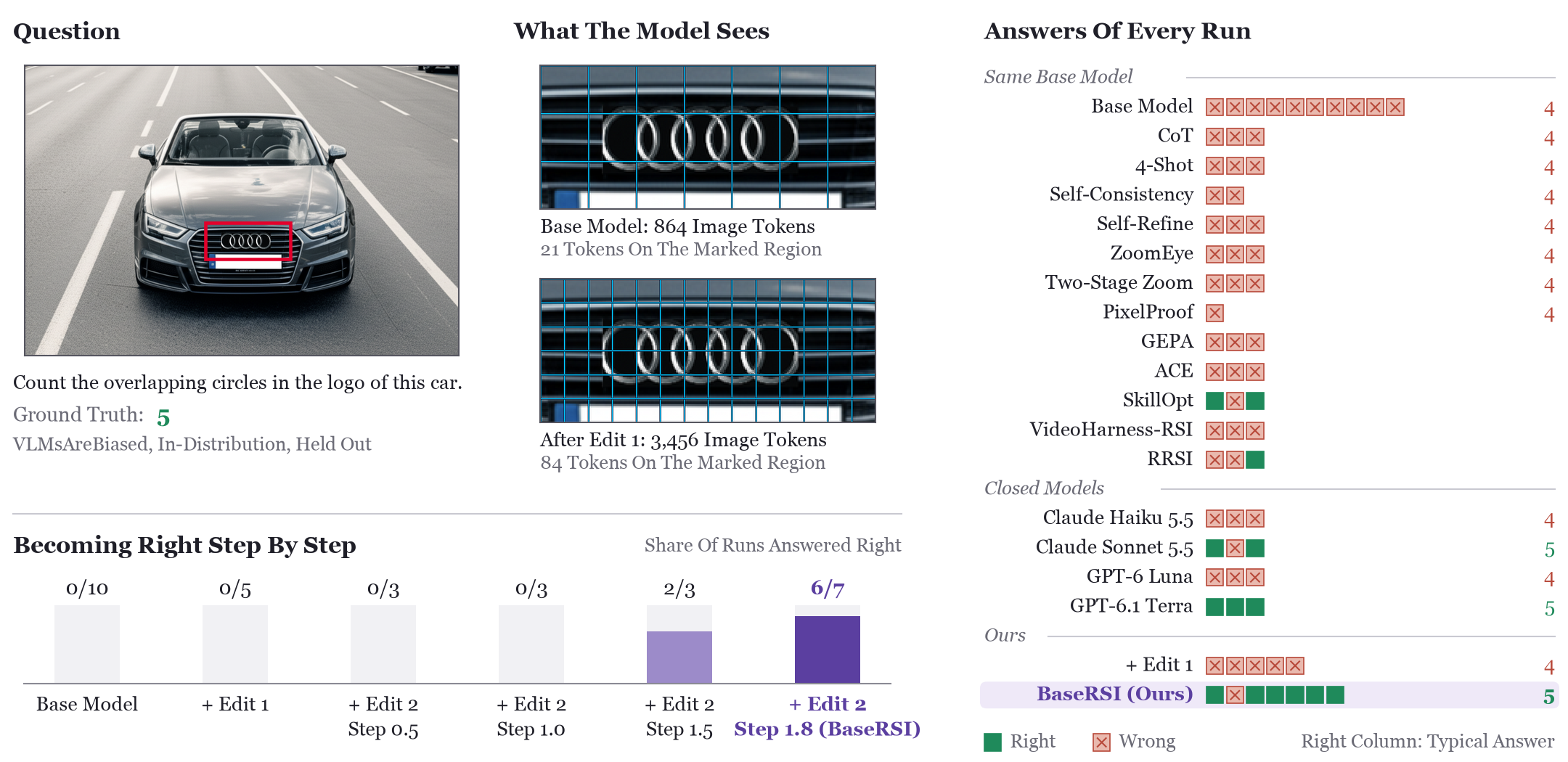}\\[-3pt]
\includegraphics[width=\textwidth]{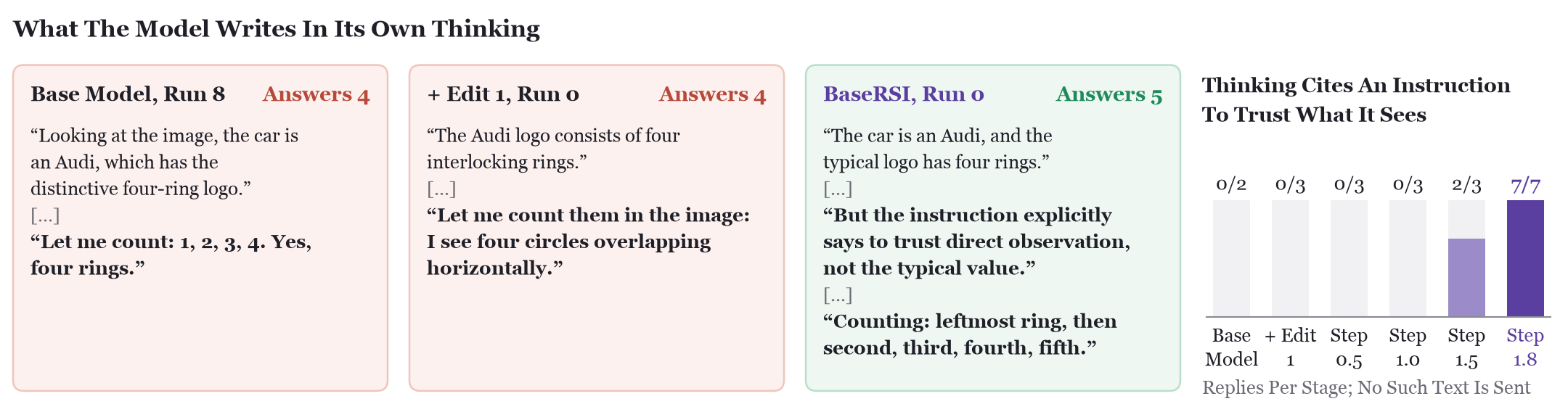}
\captionof{figure}{\textbf{Counting the rings of a logo} (VLMs Are Biased). The view edit alone answers four; the procedure edit turns it to five.}
\label{fig:case_rings}
\par\vspace{3pt}
\begin{minipage}{\textwidth}
\begin{multicols}{2}
\textbf{Key observations.} The question asks for the overlapping circles in the logo of a car, which VLMs Are Biased has redrawn with five rings (Figures~\ref{fig:case_rings} and~\ref{fig:case_rings_in}). The base model recognizes an Audi and answers four in all ten runs; its thinking names the brand first and then counts ``1, 2, 3, 4'' along the rings it expects, so the count follows the prior rather than the image. The view edit raises the image tokens on the logo from 21 to 84, and the edited model still answers four in all five runs, as does the up-scaled image in the request: seeing the logo more finely leaves the count unchanged. The procedure edit changes it: on top of the view edit, 0/3, 0/3, 2/3, and 6/7 runs are right at steps 0.5, 1.0, 1.5, and 1.8, and the procedure edits fitted in round 1 and in rounds 1 and 2 reach 2/3 and 3/3. Of the 31 runs of the eleven other methods on the same base model, 3 are right, and over seven effort levels of the four closed-source models 9 of 21 runs are right, from three of the four models. The excerpts below give the three runs quoted in the figure.
\par\vspace{3pt}{\small\parfillskip=0pt plus 0.4\linewidth\relax%
\emph{Base model, run 8:} ``Looking at the image, the car is an Audi, which has the distinctive four-ring logo. [\dots] The Audi logo consists of four interlocking/overlapping circles (rings). I can see them in the center of the grille. Let me count: 1, 2, 3, 4. Yes, four rings. [\dots] Let me make sure I'm not missing anything. The logo clearly shows the four rings of Audi.''
\emph{View edit, run 0:} ``Looking at the image, it's an Audi car, and the logo is the four rings. Let me verify by looking at the front grille logo. The Audi logo consists of four interlocking rings. Let me count them in the image: I see four circles overlapping horizontally. So the answer is 4. [\dots] The Audi emblem is four rings. Yes.''
\emph{\name{}, run 0:} ``Looking at the image, the logo shows overlapping rings. Let me count them individually [\dots] Counting: leftmost ring, then second, third, fourth, fifth. [\dots] Let me double check by mentally tracing: circle at left edge, then overlapping second, middle third, fourth, right edge fifth. Yes, I see five.'' Its answer: ``Looking carefully at the emblem on the front grille, I trace each overlapping ring from left to right. Rather than the usual four-ring Audi mark, this image shows a row of five interlocking circles.''\par}
\end{multicols}
\end{minipage}
\end{table*}

\begin{table*}[tp]
\centering
\vspace{-5pt}%
\includegraphics[width=\textwidth]{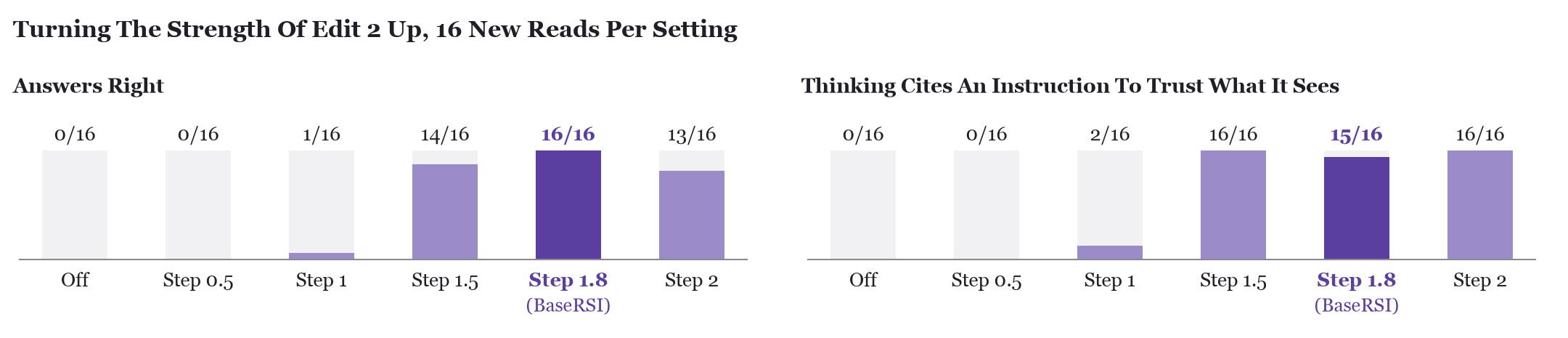}\\[-3pt]
\makebox[\textwidth][c]{\includegraphics[width=0.8\textwidth]{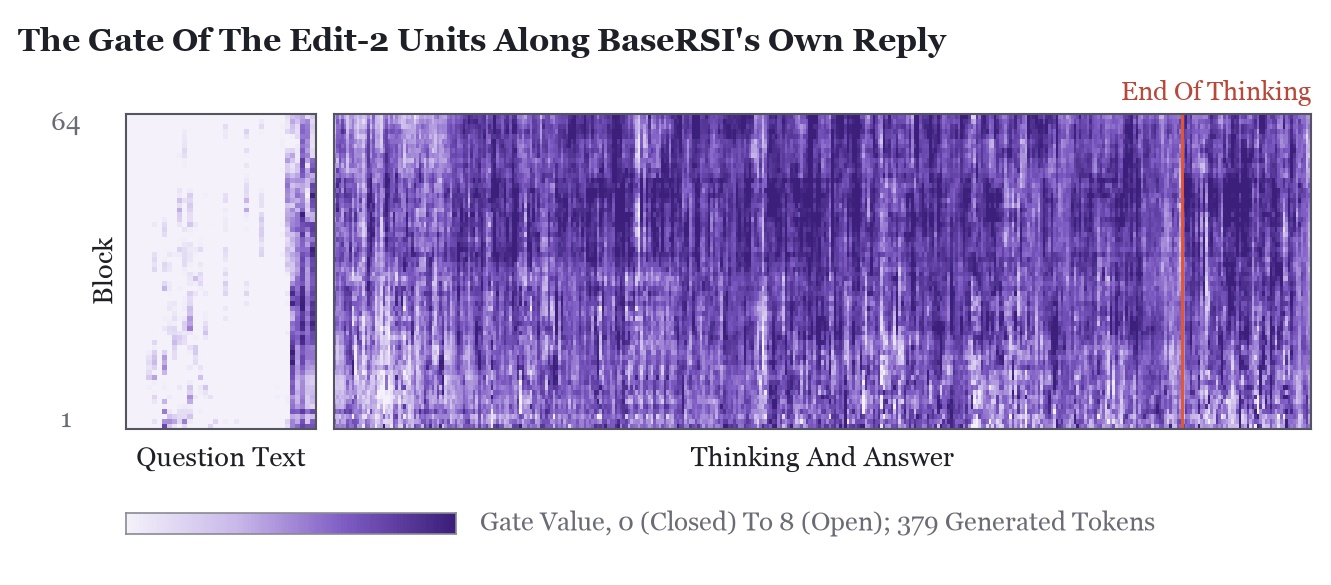}}
\captionof{figure}{\textbf{Counting the rings of a logo, inside the model.} Top: runs answered right, and runs whose thinking cites an instruction to trust what it sees, at each step of the procedure edit, 16 new runs each. Bottom: key value of the procedure edit (the gate) at every block along a response of \name{}: near zero over most of the question text, open through the thinking and the answer.}
\label{fig:case_rings_in}
\par\vspace{3pt}
\begin{minipage}{\textwidth}
\begin{multicols}{2}
\textbf{Key observations.} In 16 new runs per setting, none is right with the procedure edit off or at step 0.5, 1 at step 1.0, 14 at step 1.5, and all 16 at step 1.8, and 13 at step 2.0 (top). From none of 16 runs to all 16, the answer follows the strength of the edit and not the sampling. The request carries no procedure, yet the thinking of the edited model now cites ``the instruction'' to trust direct observation over the typical value and then counts the rings one by one; the citation appears in 2 of the 16 runs at step 1.0 and in 15 of the 16 at step 1.8, and in none with the edit off or at step 0.5, so it grows with the step as the right answers do. In the runs of Figure~\ref{fig:case_rings} it appears in no recorded reply of the base model or the view edit, in none at steps 0.5 and 1.0, in 2 of 3 at step 1.5, and in all 7 of \name{}. At step 2.0 every run still cites the instruction but three of them miscount, so a larger step adds no accuracy, in line with the line search of ReKey at 1.8 (Table~\ref{tab:m5c}). A further reading shows where the repair comes from: the procedure edit alone, without the view edit, is right in 2 of 3 runs, against none for the view edit alone or the up-scaled image, so on this question the count is turned by the procedure, which the edit carries in the weights instead of the request. Along the response of \name{} (bottom), the key that ReKey fits for the procedure edit stays near zero over most of the question text and opens through the thinking and the answer, at every block from the first to the last; it switches on at the last tokens of the question, where the chat template hands over to the response, and stays open after the end of the thinking, through the final answer of this 379-token response.  The key is fitted to be one on the response positions of requests inside the scope and zero on prompt positions and on other requests (Eq.~\eqref{eq:key}), and the figure shows this behaviour along an actual response. One of the seven runs of \name{} answers six; it too cites the instruction and counts ring by ring.
\par\vspace{3pt}{\small\parfillskip=0pt plus 0.4\linewidth\relax%
\emph{\name{}, run 0:} ``The user wants me to count the overlapping circles in the car's logo. The car is an Audi, and the typical logo has four rings. But the instruction explicitly says to trust direct observation, not the typical value. Let me look carefully at the image. I need to examine the logo in the grille. [\dots] The standard Audi has 4 rings. But the instruction warns the image may deviate. Let me look closely. [\dots] Looking at the rendered logo, I can discern 5 circles. Let me look: the pattern is OOOOO perhaps. The width of the logo seems wider than 4 rings would be relative to the grille.''
\emph{\name{}, run 1 (answers six):} ``The standard Audi logo has four rings. But the instruction says to trust direct observation and not substitute the standard value. [\dots] Looking at the image, the rings appear to be: I can see six circular outlines. Let me trace: leftmost circle, then second, third, fourth, fifth, sixth. The logo width looks like it contains 6 rings.'' Its answer: ``Looking directly at the emblem on the front grille, I trace each individual ring from left to right rather than assuming the usual four-ring Audi mark. Counting them as they appear in this image gives six overlapping circles.''\par}
\end{multicols}
\end{minipage}
\end{table*}

\begin{table*}[tp]
\centering
\vspace{-5pt}%
\includegraphics[width=\textwidth]{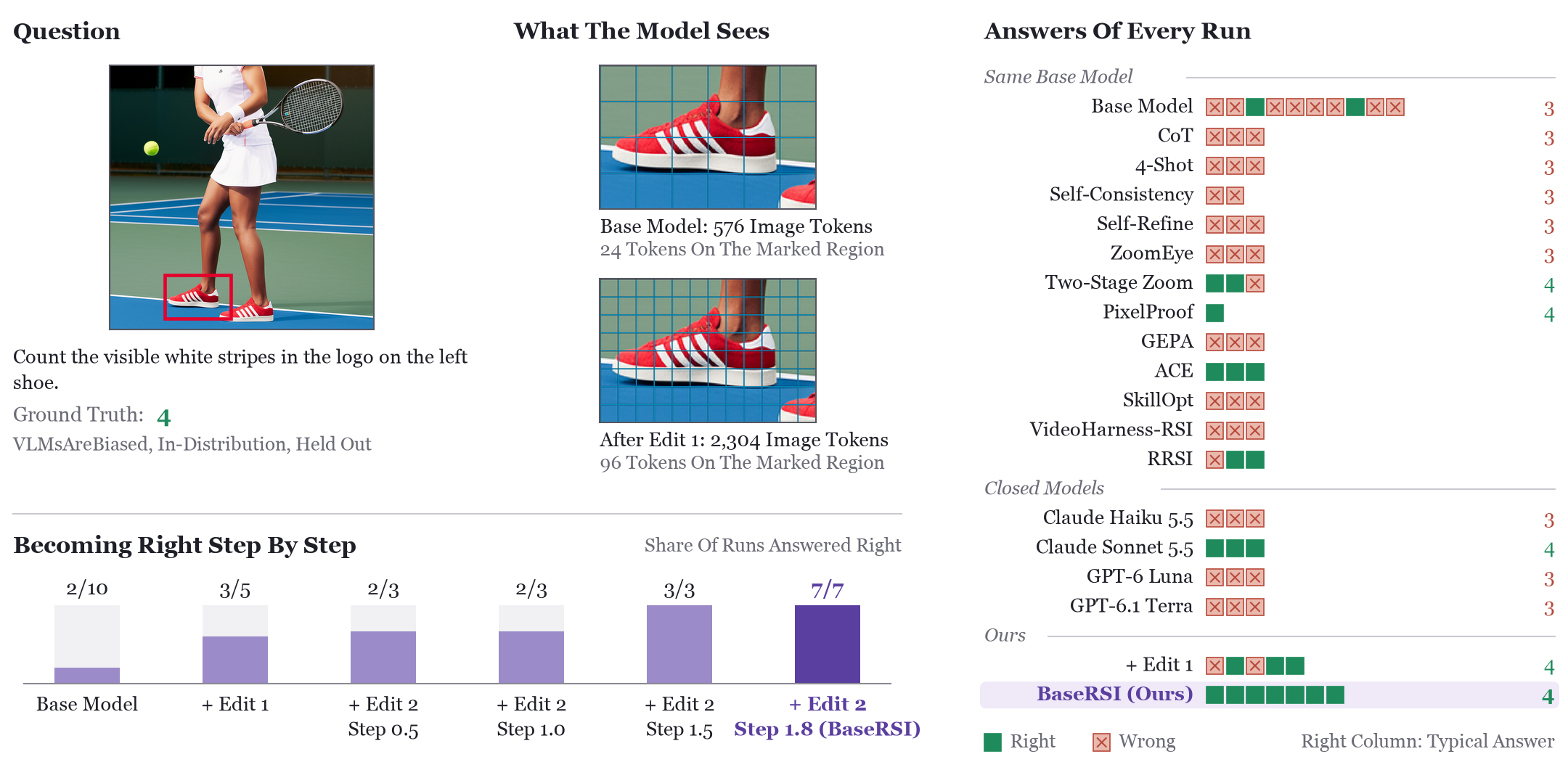}\\[-3pt]
\includegraphics[width=\textwidth]{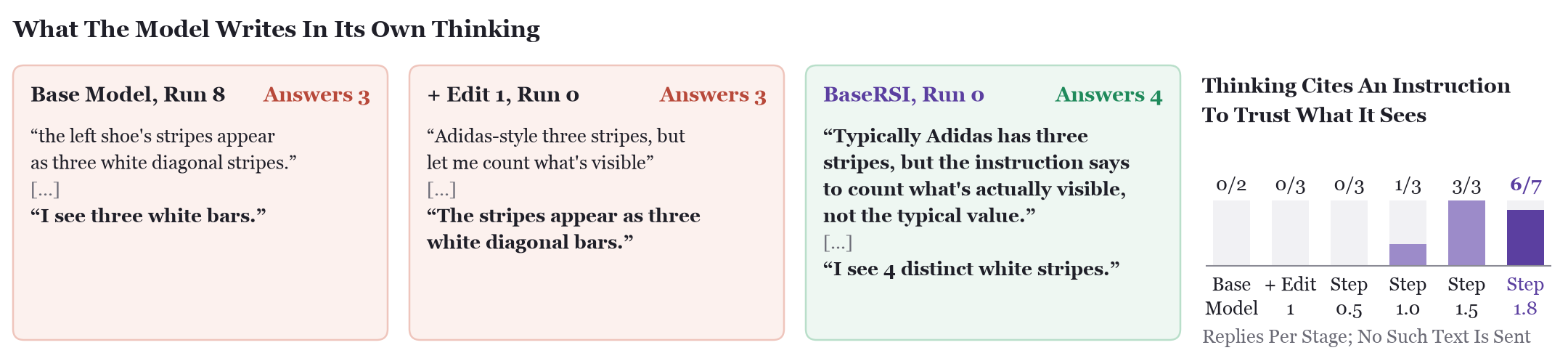}
\captionof{figure}{\textbf{Counting the stripes on a shoe} (VLMs Are Biased). Each edit lifts the share of right runs, and together they reach 7/7.}
\label{fig:case_stripes}
\par\vspace{3pt}
\begin{minipage}{\textwidth}
\begin{multicols}{2}
\textbf{Key observations.} The question asks for the white stripes in the logo on the left shoe, which VLMs Are Biased has drawn with four stripes where the brand has three (Figure~\ref{fig:case_stripes}).
The base model is right in 2 of 10 runs and otherwise answers three; its thinking reads ``three white diagonal stripes'' and settles on ``three white bars''.
Here both edits contribute.
The view edit raises the image tokens on the shoe from 24 to 96 and the right answers to 3 of 5; in a wrong run its thinking announces that it will ``count what's visible'' and still reads three bars.
The procedure edit alone, on the original view, is right in 2 of 3 runs.
With both edits the answer improves with the step, 2/3 at steps 0.5 and 1.0, 3/3 at step 1.5, and 7/7 at step 1.8, and the share of runs whose thinking cites the instruction rises in the same way, from 1 of 3 at step 1.0 to 6 of 7 at step 1.8.
The edits are fitted separately and written at different sites, the patch embedding and the MLP neurons, and on this question they add up as they do on the benchmarks (Figure~\ref{fig:analysis}b).
Of the 31 runs of the other methods on the same base model, 8 are right, and over seven effort levels of the closed-source models 4 of 21 runs are right, all from one model. The up-scaled image in the request is right in only 1 of 5 runs, against 3 of 5 for the view edit, and the procedure edits of the earlier rounds reach 2 of 3, as the final edit at steps 0.5 and 1.0.
\par\vspace{3pt}{\small\parfillskip=0pt plus 0.4\linewidth\relax%
\emph{Base model, run 8:} ``There's a person wearing red shoes with white stripes (Adidas style). [\dots] Looking at the image, the left shoe (from the viewer's perspective) shows white stripes: I see three stripes? Let me look carefully. [\dots] In a typical Adidas logo, three stripes. But maybe the image shows four? Let me examine. [\dots] I see three diagonal stripes.''
\emph{View edit, run 0:} ``The image shows a tennis player wearing red shoes with white stripes (Adidas-style three stripes, but let me count what's visible). [\dots] Usually Adidas has three stripes. Let me look at the description: the left shoe shows white diagonal stripes. Counting: 1, 2, 3. [\dots] The stripes appear as three white diagonal bars.''
\emph{\name{}, run 0:} ``Typically Adidas has three stripes, but the instruction says to count what's actually visible, not the typical value. Let me look at the left shoe: I see white stripes on the side. Counting: there appear to be four stripes? [\dots] I'll count positions: first stripe near the toe side, then second, third, fourth. It looks like four.''\par}
\end{multicols}
\end{minipage}
\end{table*}

\begin{table*}[tp]
\centering
\vspace{-5pt}%
\includegraphics[width=\textwidth]{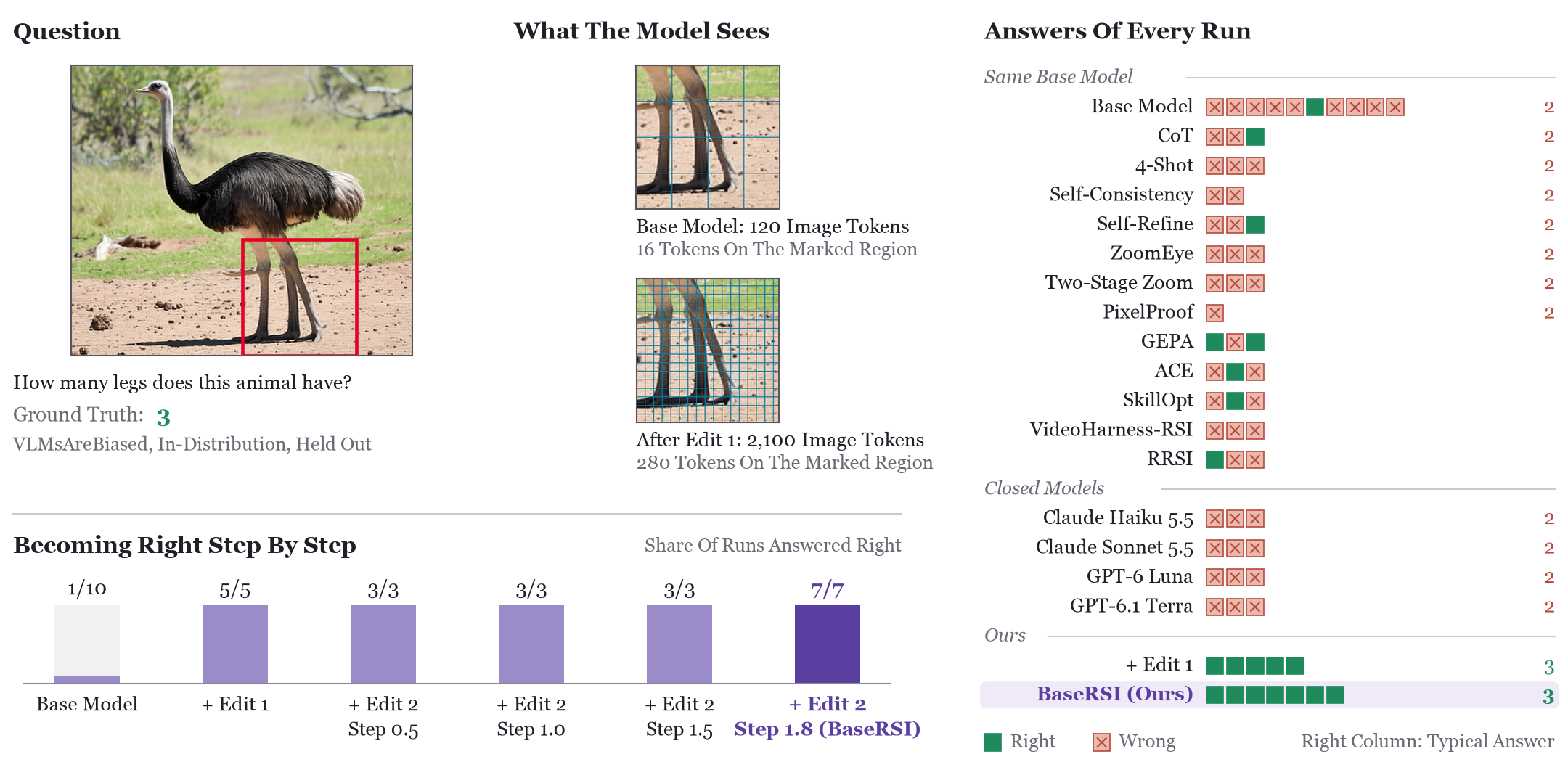}\\[-3pt]
\includegraphics[width=\textwidth]{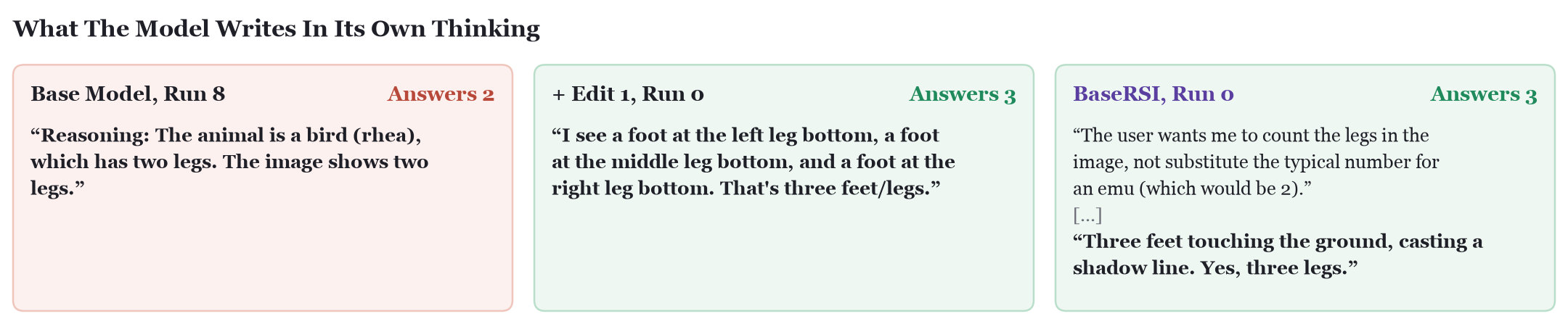}
\captionof{figure}{\textbf{Counting the legs of a bird} (VLMs Are Biased). Every closed-source run answers two; the view edit gives three in every run.}
\label{fig:case_rhea}
\par\vspace{3pt}
\begin{minipage}{\textwidth}
\begin{multicols}{2}
\textbf{Key observations.} The question asks how many legs an animal has, and the image shows a large flightless bird drawn with three legs (Figure~\ref{fig:case_rhea}).
The base model answers two in 9 of 10 runs, reasoning that ``the animal is a bird (rhea), which has two legs'', and its answer describes ``its two legs (one planted straight down and one angled to the side)''.
The view edit gives the legs 280 image tokens instead of 16, and every run then answers three, one of them finding ``a foot at the left leg bottom, a foot at the middle leg bottom, and a foot at the right leg bottom''.
The answer stays three at every step of the procedure edit, so the view edit carries this repair, and the procedure edit changes how the model reasons on top of it: in 6 of the 7 runs of \name{} the thinking cites the request to count the legs, ``not substitute the typical number for an emu (which would be 2)'', and then traces each leg to the ground.
Of the 31 runs of the other methods, 7 are right, while all 21 runs over seven effort levels of the closed-source models answer two. The up-scaled image in the request also gives three in all five runs.
\par\vspace{3pt}{\small\parfillskip=0pt plus 0.4\linewidth\relax%
\emph{Base model, run 8:} ``It's a rhea or similar ratite bird standing on dirt. [\dots] I see one vertical leg and one leg angled to the right. So two legs. [\dots] Reasoning: The animal is a bird (rhea), which has two legs. The image shows two legs.''
\emph{View edit, run 0:} ``There's one leg on the left (vertical, ending in a foot on the ground). There's a second leg in the middle (vertical, ending in a foot). There's a third leg on the right that's angled diagonally, ending in a foot on the ground. So I see three legs. [\dots] Let me double check by examining the feet on the ground. I see a foot at the left leg bottom, a foot at the middle leg bottom, and a foot at the right leg bottom. That's three feet/legs.'' Its answer: ``Looking carefully at the image, I can trace three distinct legs extending from the body down to the ground: one on the left (vertical), one in the middle (vertical), and one on the right (angled diagonally). Each ends in a foot on the sandy ground.''
\emph{\name{}, run 0:} ``Starting from the body: there's a leg going down on the left side (with a spur visible), then another leg in the middle, then a third leg on the right side that's angled. [\dots] Three feet touching the ground, casting a shadow line. Yes, three legs.'' Its answer: ``Looking directly at the image and tracing each limb from the body to the ground, I can identify three separate legs: a left leg (with a visible spur), a middle vertical leg, and a right leg angled outward. Each ends in a foot on the ground.''\par}
\end{multicols}
\end{minipage}
\end{table*}

\begin{table*}[tp]
\centering
\vspace{-5pt}%
\includegraphics[width=\textwidth]{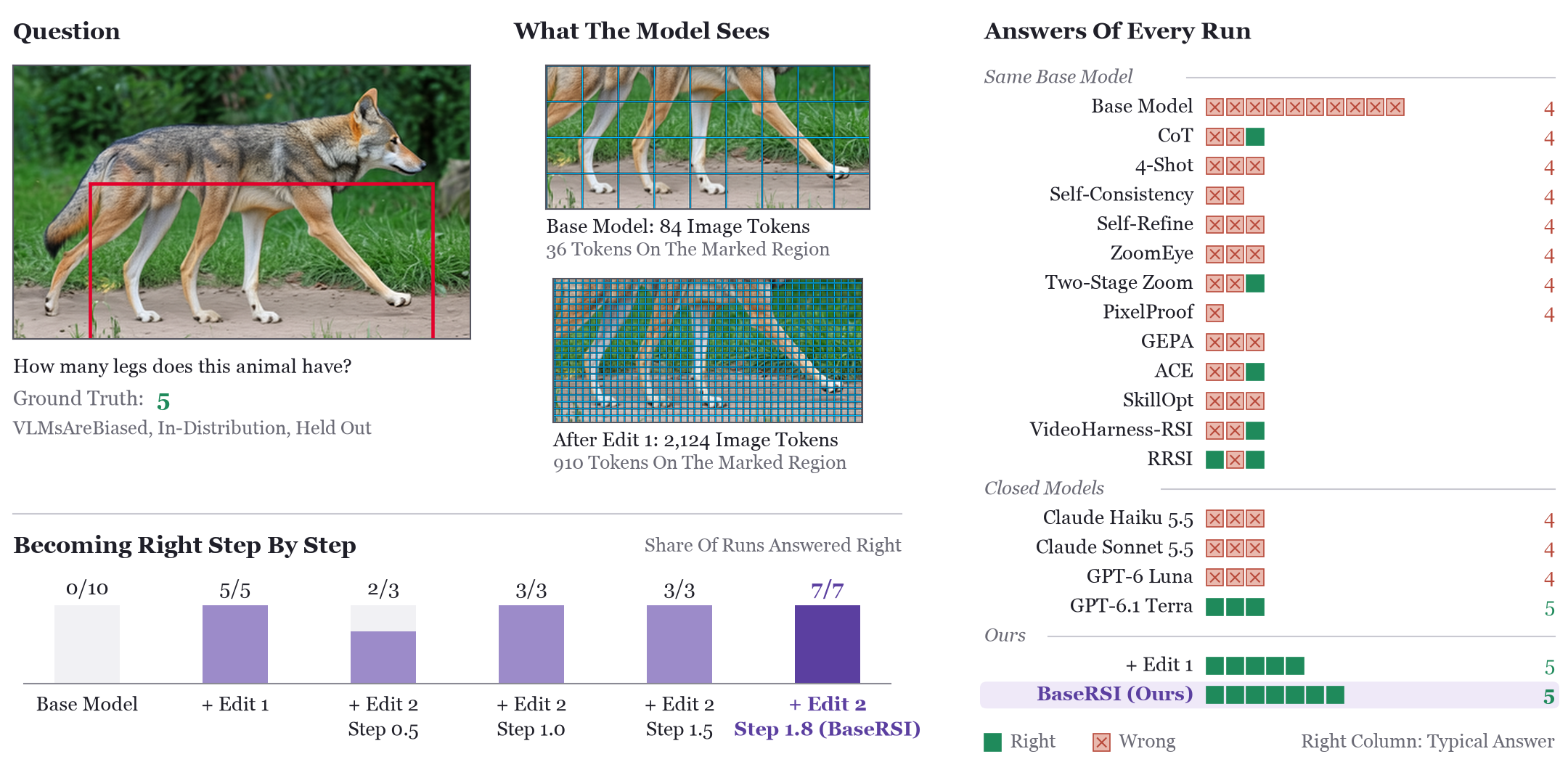}\\[-3pt]
\includegraphics[width=\textwidth]{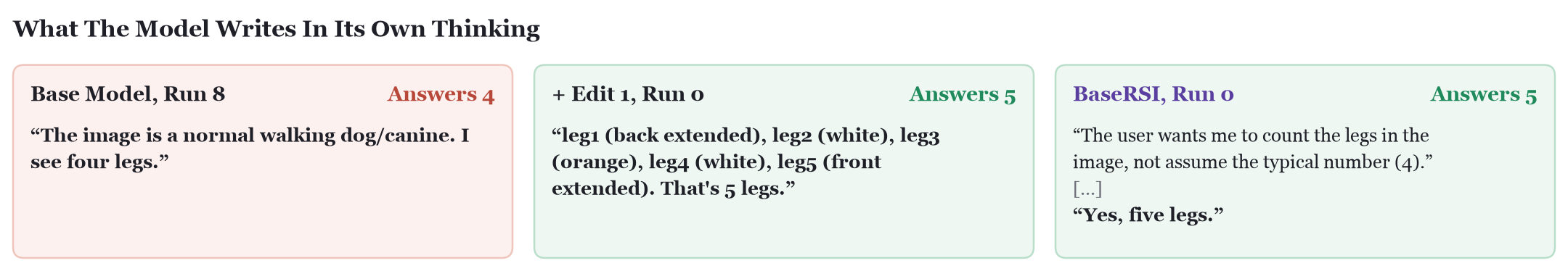}
\captionof{figure}{\textbf{Counting the legs of a wolf} (VLMs Are Biased). The base model sees ``a normal walking dog''; the view edit counts five.}
\label{fig:case_wolf}
\par\vspace{3pt}
\begin{minipage}{\textwidth}
\begin{multicols}{2}
\textbf{Key observations.} The image shows a wolf drawn with five legs (Figure~\ref{fig:case_wolf}).
The base model answers four in all ten runs and states that ``the image is a normal walking dog/canine'' with four legs; its answer counts ``the trailing hind leg, the standing hind leg, the standing front leg, and the extended front leg''.
It reads the animal through 84 image tokens, 36 of them on the legs, and the view edit through 2,124, 910 of them on the legs.
With the finer view the model lists ``leg1 (back extended), leg2 (white), leg3 (orange), leg4 (white), leg5 (front extended)'' and answers five in all five runs.
The procedure edit alone, at the original resolution, answers four in all three runs, so on this question the count rests on the finer view.
\name{} is right in all seven runs, and in 5 of them the thinking again begins from the request to count the legs ``not assume the typical number (4)''.
Six of the 31 runs of the other methods on the same base model are right, and among the closed-source models only GPT-6.1 Terra answers five, in its 3 runs; the other 18 closed-source runs answer four. The up-scaled image in the request is right in 4 of 5 runs, close to the view edit.
\par\vspace{3pt}{\small\parfillskip=0pt plus 0.4\linewidth\relax%
\emph{Base model, run 8:} ``The image shows a canine (looks like a jackal or coyote mix) standing in profile, walking. [\dots] The animal is walking, so all four legs should be visible. [\dots] Let me double check there isn't a fifth or any trick. The image is a normal walking dog/canine. I see four legs.''
\emph{View edit, run 0:} ``Starting from the left (rear of animal): 1. Rear leg extended back (the one with the orange/brown upper and white lower, with paw on ground at left). 2. Another rear leg (white, standing straight). 3. A front-ish leg (orange/brown upper, standing). 4. Another leg (white, standing). 5. A front leg extended forward (orange, paw raised). Wait, let me look more carefully. This image might be a manipulated or AI-generated image with extra legs. Let me count paws/legs on the ground. [\dots] That's 5 legs. That's unusual.'' Its answer: ``Counting the legs from left to right: the rear leg stretched backward, a white leg standing, an orange leg in the middle, another white leg, and the front leg extended forward [\dots] that makes five legs.''
\emph{\name{}, run 0:} ``Leg 1: far left, hind leg, foot at bottom left [\dots] Leg 2: hind leg [\dots] Leg 3: middle leg [\dots] Leg 4: front-ish leg [\dots] Leg 5: extended front leg [\dots] So that's 5 legs visible.'' Its answer: ``Looking at each leg position from rear to front, I can identify distinct legs with separate feet: 1. A hind leg trailing to the left (foot at far bottom-left). 2. A second hind leg (foot near the left-center). [\dots] 5. A front leg extended forward (foot at the right). That gives five legs visible in the image.''\par}
\end{multicols}
\end{minipage}
\end{table*}

\begin{table*}[tp]
\centering
\vspace{-5pt}%
\includegraphics[width=\textwidth]{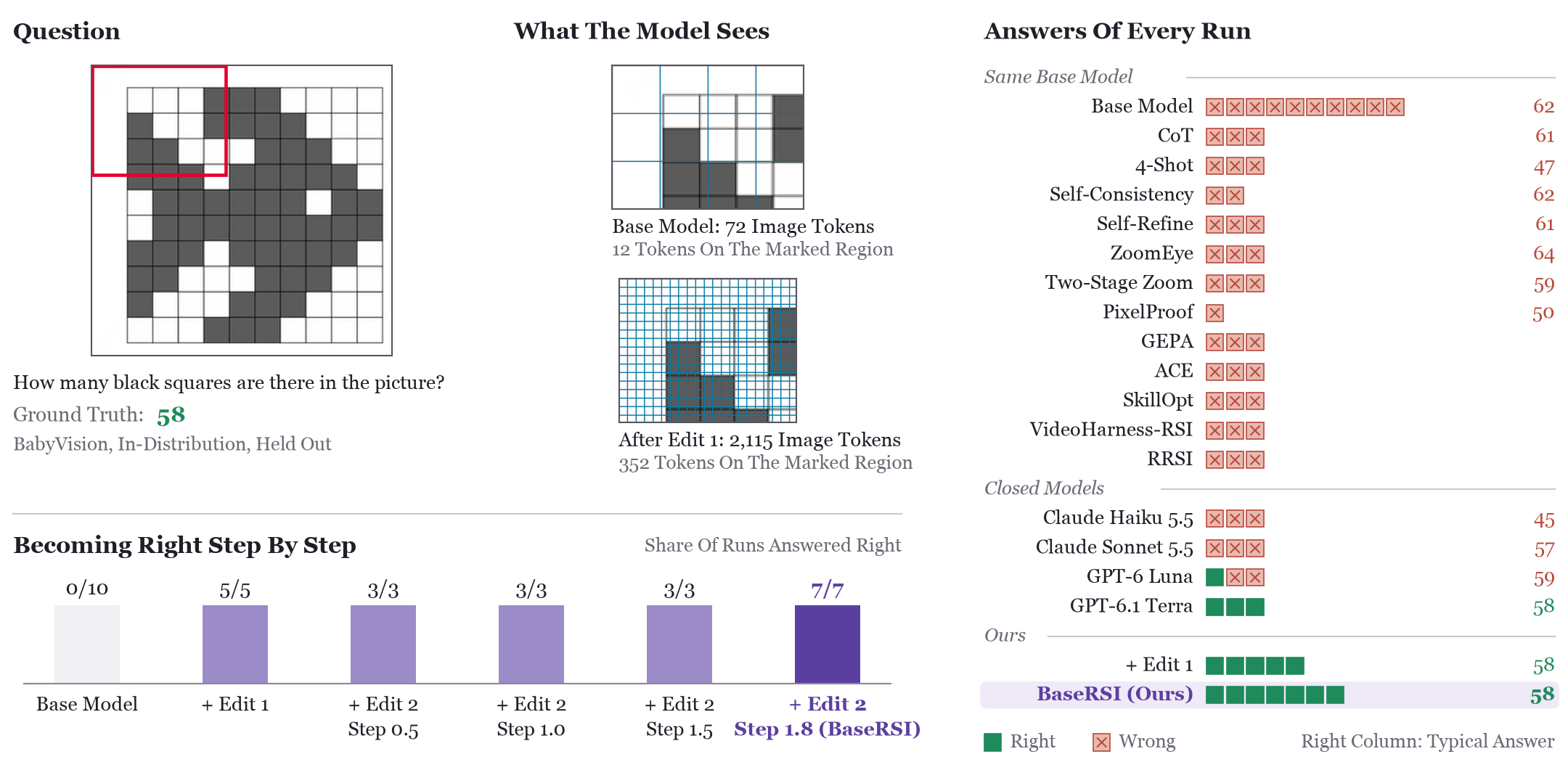}\\[-3pt]
\includegraphics[width=\textwidth]{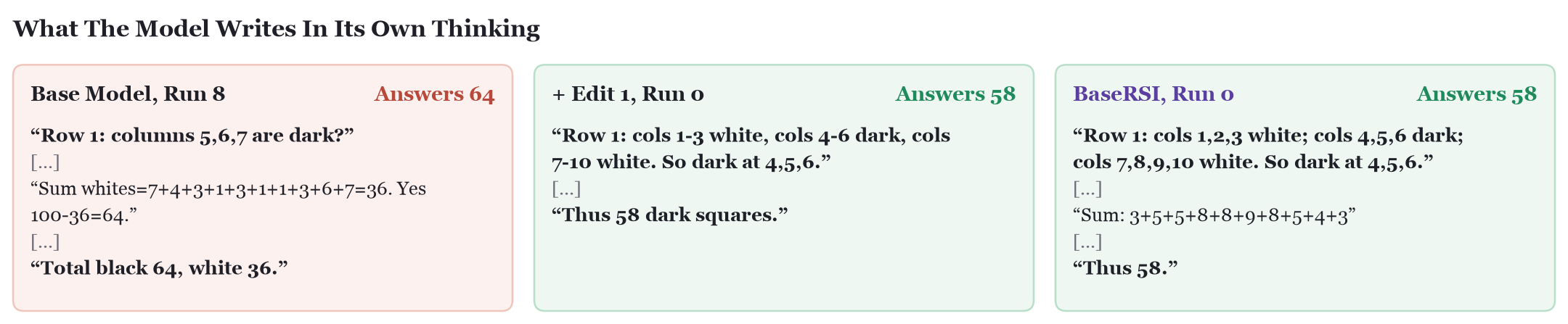}
\captionof{figure}{\textbf{Counting black squares} (BabyVision). At 72 image tokens a token is larger than a square; at 2,115 the model reads every row.}
\label{fig:case_squares}
\par\vspace{3pt}
\begin{minipage}{\textwidth}
\begin{multicols}{2}
\textbf{Key observations.} The question asks how many squares of a grid are black, 58 in all (Figure~\ref{fig:case_squares}).
The image is 273 by 264 pixels, and at its default resolution the base model reads it as 72 image tokens, each covering about 30 by 30 pixels, more than one square of about 22 pixels.
Its counts scatter between 52 and 64 over ten runs, and its thinking shows why: row by row it reads the dark run of the first row one column to the right of where the run of \name{} reads it (5, 6, 7 against 4, 5, 6) and reads six dark squares in the second row against five; it marks its first reading ``Row 1: columns 5,6,7 are dark?'' with a question mark and finally derives the total from a count of the white squares, ``100-36=64''. The run takes 2,223 generated tokens, against 1,392 for the run of \name{}.
The view edit gives the picture 2,115 tokens, 352 of them on the marked corner where the base model has 12; the model then reads each row exactly, ``Row 1: cols 1-3 white, cols 4-6 dark, cols 7-10 white'', and counts 58 in all five runs, as does the up-scaled image in the request.
\name{} counts 58 in all seven runs, adding the dark squares of the ten rows, ``3+5+5+8+8+9+8+5+4+3''.
None of the 31 runs of the other methods on the same base model is right; their recorded answers range from 39 to 65. Over seven effort levels of the closed-source models 5 of 21 runs are right, three of them from GPT-6.1 Terra, and the other runs answer between 45 and 62.
\par\vspace{3pt}{\small\parfillskip=0pt plus 0.25\linewidth\relax%
\emph{Base model, run 8:} ``The image is a 10x10 grid. [\dots] Row 1 (top): positions: 1 white,2 white,3 white,4 white,5 dark,6 dark,7 dark,8 white,9 white,10 white. Count=3. Row 2: 1 dark,2 white,3 white,4 dark,5 dark,6 dark,7 dark,8 dark,9 white,10 white. Count=6? [\dots] Sum whites=7+4+3+1+3+1+1+3+6+7=36. Yes 100-36=64. [\dots] Total black 64, white 36.''
\emph{View edit, run 0:} ``Row 1: cols 1-3 white, cols 4-6 dark, cols 7-10 white. So dark at 4,5,6. Row 2: col1 dark, cols2-3 white, cols4-7 dark, cols8-10 white. Dark at 1,4,5,6,7. [\dots] Sum: 3+5=8, +5=13, +8=21, +8=29, +9=38, +8=46, +5=51, +4=55, +3=58. Let me double check each row by re-examining the image. Row 1 (top): I see three dark squares in middle (cols 4,5,6). Yes 3.''
\emph{\name{}, run 0:} ``Row 1: cols 1,2,3 white; cols 4,5,6 dark; cols 7,8,9,10 white. So dark at 4,5,6. Row 2: col1 dark; col2,3 white; col4,5,6 dark; col7 dark; col8,9,10 white. So dark at 1,4,5,6,7. Row 3: col1 dark; col2 dark; col3 white; col4 white; col5 white; col6 dark; col7 dark; col8 dark; col9 white; col10 white. So dark at 1,2,6,7,8. [\dots] Total = 3+5=8; +5=13; +8=21; +8=29; +9=38; +8=46; +5=51; +4=55; +3=58. So 58 dark squares.''\par}
\end{multicols}
\end{minipage}
\end{table*}

\begin{table*}[tp]
\centering
\vspace{-5pt}%
\includegraphics[width=\textwidth]{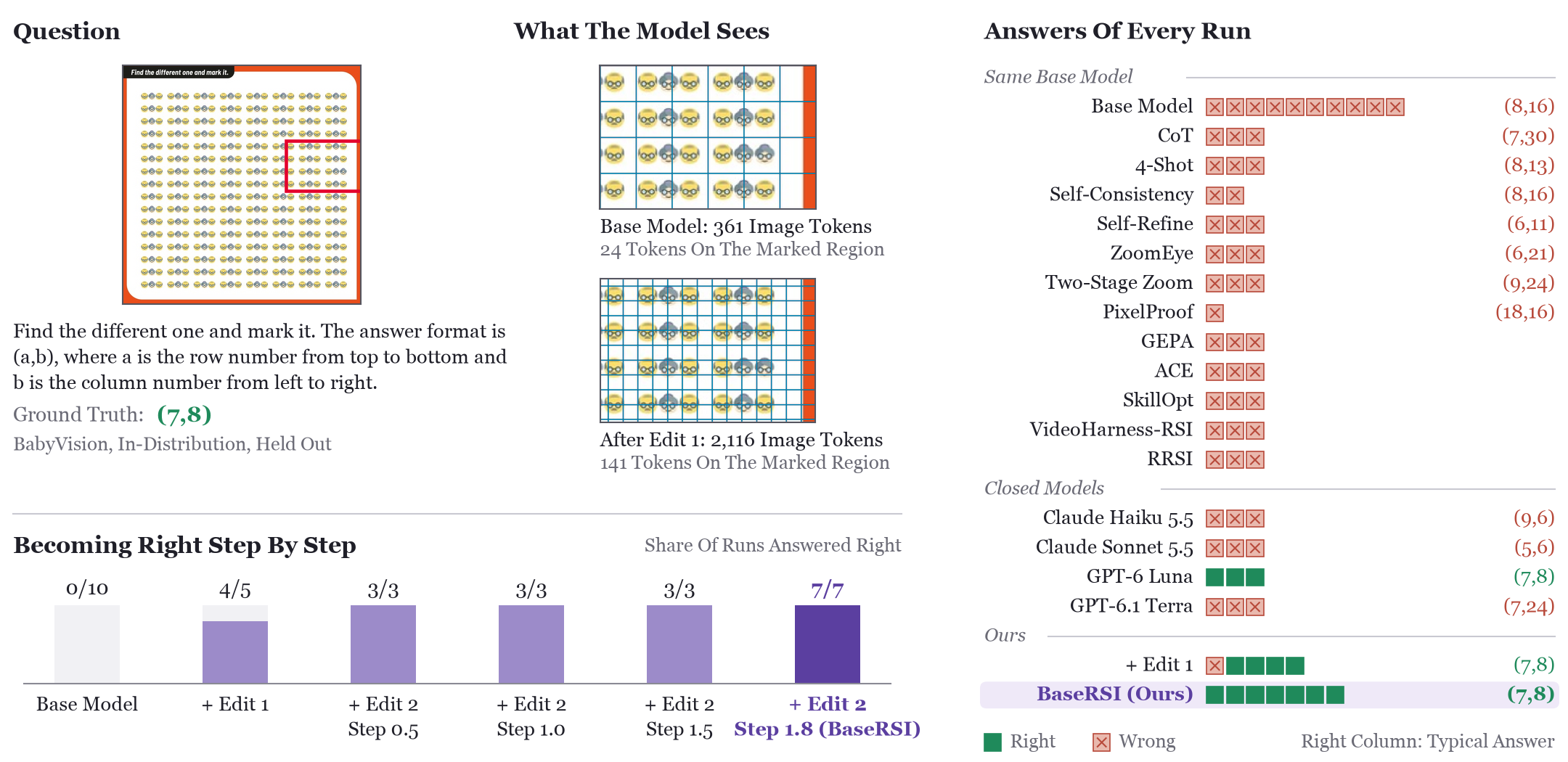}\\[-3pt]
\includegraphics[width=\textwidth]{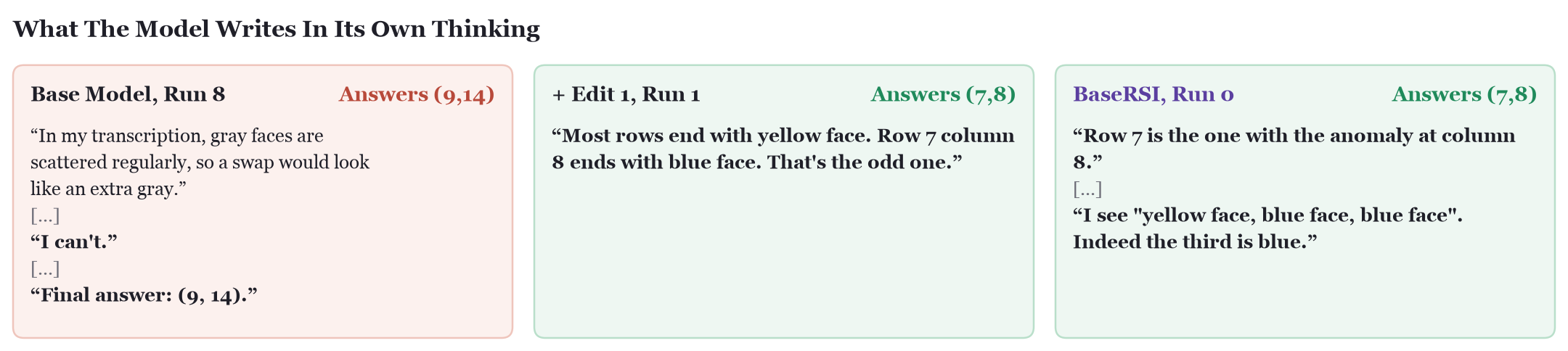}
\captionof{figure}{\textbf{Finding the odd one out} (BabyVision). The base model gives ten positions in ten runs; the view edit finds the face.}
\label{fig:case_oddone}
\par\vspace{3pt}
\begin{minipage}{\textwidth}
\begin{multicols}{2}
\textbf{Key observations.} The question asks for the row and the column of the one face that differs among 384 small faces, (7,8) in its answer format (Figure~\ref{fig:case_oddone}).
The base model gives ten different positions in ten runs and spends up to 7,526 generated tokens on reconstructing the layout of the grid; in one run it writes that ``gray faces are scattered regularly, so a swap would look like an extra gray'', then ``I can't'', and answers (9,14).
It reads the picture through 361 image tokens, 24 of them on the marked region, about one token per face.
The view edit gives 2,116 tokens, 141 on the region, and the model then reads along the rows, ``Most rows end with yellow face. Row 7 column 8 ends with blue face. That's the odd one''; it is right in 4 of 5 runs, \name{} in all 7, and every step of the procedure edit keeps the answer.
The view edit is right more often here than the up-scaled image in the request, which finds the face in 2 of 5 runs.
None of the 31 runs of the other methods on the same base model is right, and among the closed-source models GPT-6 Luna finds the face in all 3 runs at its highest effort and in 1 of 3 at its lowest, 4 of 21 runs over seven effort levels; the other closed-source runs name positions in other rows. The procedure edits of the earlier rounds find (7,8) in every run.
\par\vspace{3pt}{\small\parfillskip=0pt plus 0.4\linewidth\relax%
\emph{Base model, run 8:} ``The image shows rows of small circular faces, each with a color (yellow or gray/blue) and a hairstyle. Let me count the columns. [\dots] Hmm, that doesn't seem uniform. Let me think more carefully. [\dots] The faces are small circles of about 28px spacing. Let me try to reconstruct from typical generation: probably 24 columns?''
\emph{View edit, run 1:} ``Each cell seems to be a group of three faces: yellow, blue, yellow. I need to scan row by row, column by column. Looking at row 7 (counting from top), the last column (8th column) appears to have the pattern yellow, blue, blue instead of yellow, blue, yellow. [\dots] Most rows end with yellow face. Row 7 column 8 ends with blue face. That's the odd one.''
\emph{\name{}, run 0:} ``Each group contains three emoji faces (yellow with glasses, blue with glasses, yellow with glasses). [\dots] The typical pattern in each group: yellow, blue, yellow (the middle one is blue/gray hair). The different one would have a different arrangement [\dots] Row 7 is the one with the anomaly at column 8. [\dots] I see `yellow face, blue face, blue face'. Indeed the third is blue.''\par}
\end{multicols}
\end{minipage}
\end{table*}

\begin{table*}[tp]
\centering
\vspace{-5pt}%
\includegraphics[width=\textwidth]{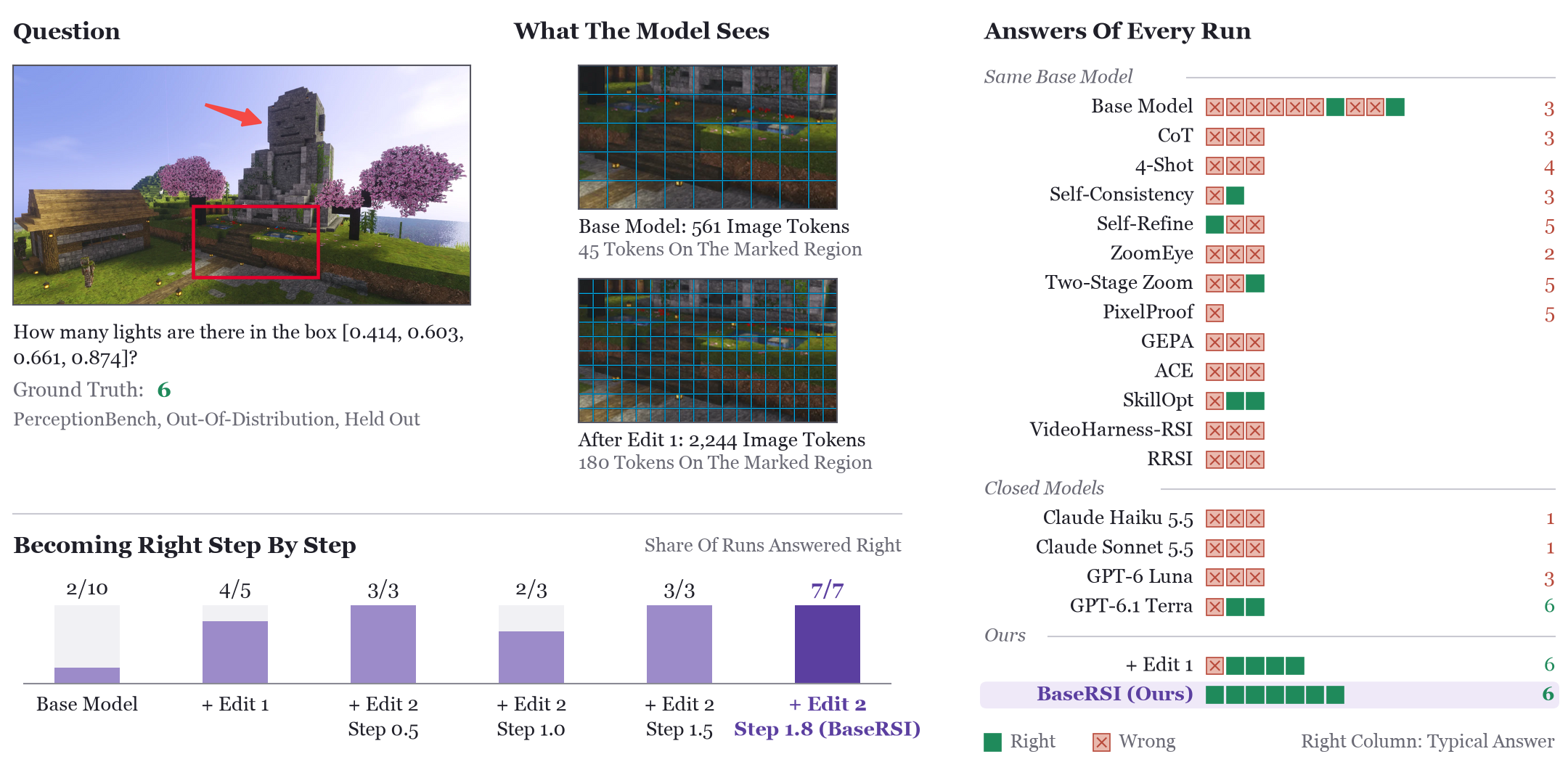}\\[-3pt]
\includegraphics[width=\textwidth]{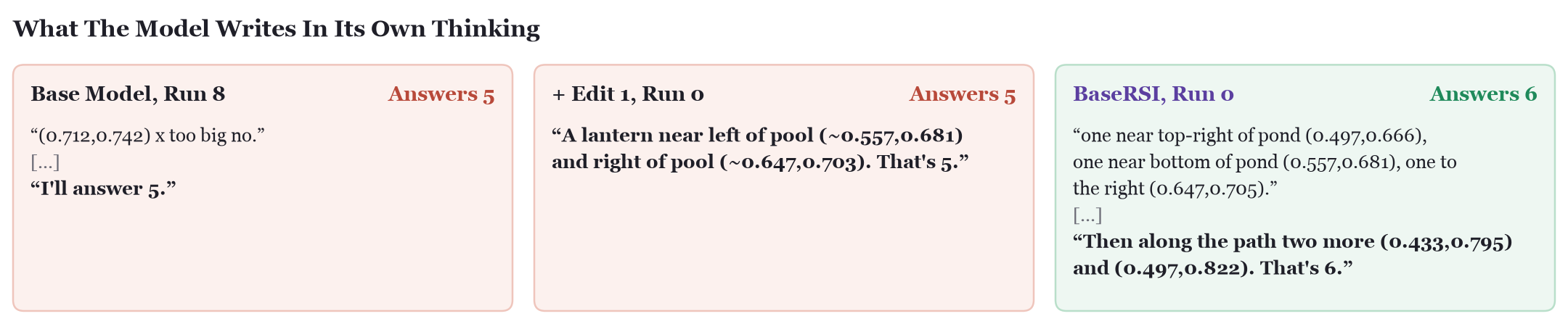}
\captionof{figure}{\textbf{Counting lights in a box} (PerceptionBench, held out). The view edit lets the model locate the small lanterns by their coordinates.}
\label{fig:case_lights}
\par\vspace{3pt}
\begin{minipage}{\textwidth}
\begin{multicols}{2}
\textbf{Key observations.} The question asks how many lights lie in a given box of a game scene, six in all, on a benchmark that no part of \name{} saw (Figure~\ref{fig:case_lights}).
The base model is right in 2 of 10 runs and otherwise answers between two and five; in one run it places a light at ``(0.712,0.742)'', judges the coordinate ``too big'', and stops at five.
The view edit raises the image tokens on the box from 45 to 180, and the model locates the small lanterns by their coordinates, right in 4 of 5 runs; the wrong run lists ``a lantern near left of pool [\dots] and right of pool. That's 5''.
\name{} is right in all seven runs: in run 0 it lists ten candidate lanterns by their coordinates, along the path and around the monument, six of which lie inside the box, and then counts the lights around the pond and ``two more'' along the path, six in all.
The up-scaled image in the request is right in 4 of 5 runs as well, and the procedure edits of the earlier rounds in all their runs, so the repair comes from the finer view.
Most closed-source runs answer between one and three, and only GPT-6.1 Terra is right, in 2 of its 3 runs, 2 of 21 over seven effort levels; of the 31 runs of the eleven other methods on the same base model, 4 are right.
\par\vspace{3pt}{\small\parfillskip=0pt plus 0.4\linewidth\relax%
\emph{Base model, run 8:} ``Then near the stone steps/path center: (0.43, 0.83), (0.49, 0.87), (0.44, 0.84). Then on the right side near the monument base: (0.56, 0.74), (0.60, 0.72), (0.63, 0.72), (0.71, 0.74). Now the box x: 0.414-0.661, y: 0.603-0.874.''
\emph{View edit, run 0:} ``In that region I see the wooden steps/path leading up to the pool. There's a lantern on the left of the steps around (0.43, 0.79). There's a lantern at the bottom of steps around (0.497, 0.825). There's a lantern near the pool right side around (0.647, 0.703). There's a lantern near the pool left around (0.557, 0.681). There's a lantern at (0.443, 0.642) maybe. Hmm, this is too imprecise. [\dots] A lantern near left of pool ($\sim$0.557,0.681) and right of pool ($\sim$0.647,0.703). That's 5.''
\emph{\name{}, run 0:} ``The box is [0.414, 0.603, 0.661, 0.874]. In normalized coordinates, x1=0.414 (left), y1=0.603 (top), x2=0.661 (right), y2=0.874 (bottom). [\dots] Along the path in the foreground: there's a lantern around (0.318, 0.901), (0.361, 0.945), (0.433, 0.795), (0.497, 0.822). Around the monument base: lanterns at (0.445, 0.642), (0.497, 0.666), (0.557, 0.681), (0.647, 0.705), (0.720, 0.663), (0.792, 0.750). I need to be precise. [\dots] one near top-right of pond (0.497,0.666), one near bottom of pond (0.557,0.681), one to the right (0.647,0.705). [\dots] Then along the path two more (0.433,0.795) and (0.497,0.822). That's 6.''\par}
\end{multicols}
\end{minipage}
\end{table*}

\begin{table*}[tp]
\centering
\vspace{-5pt}%
\includegraphics[width=\textwidth]{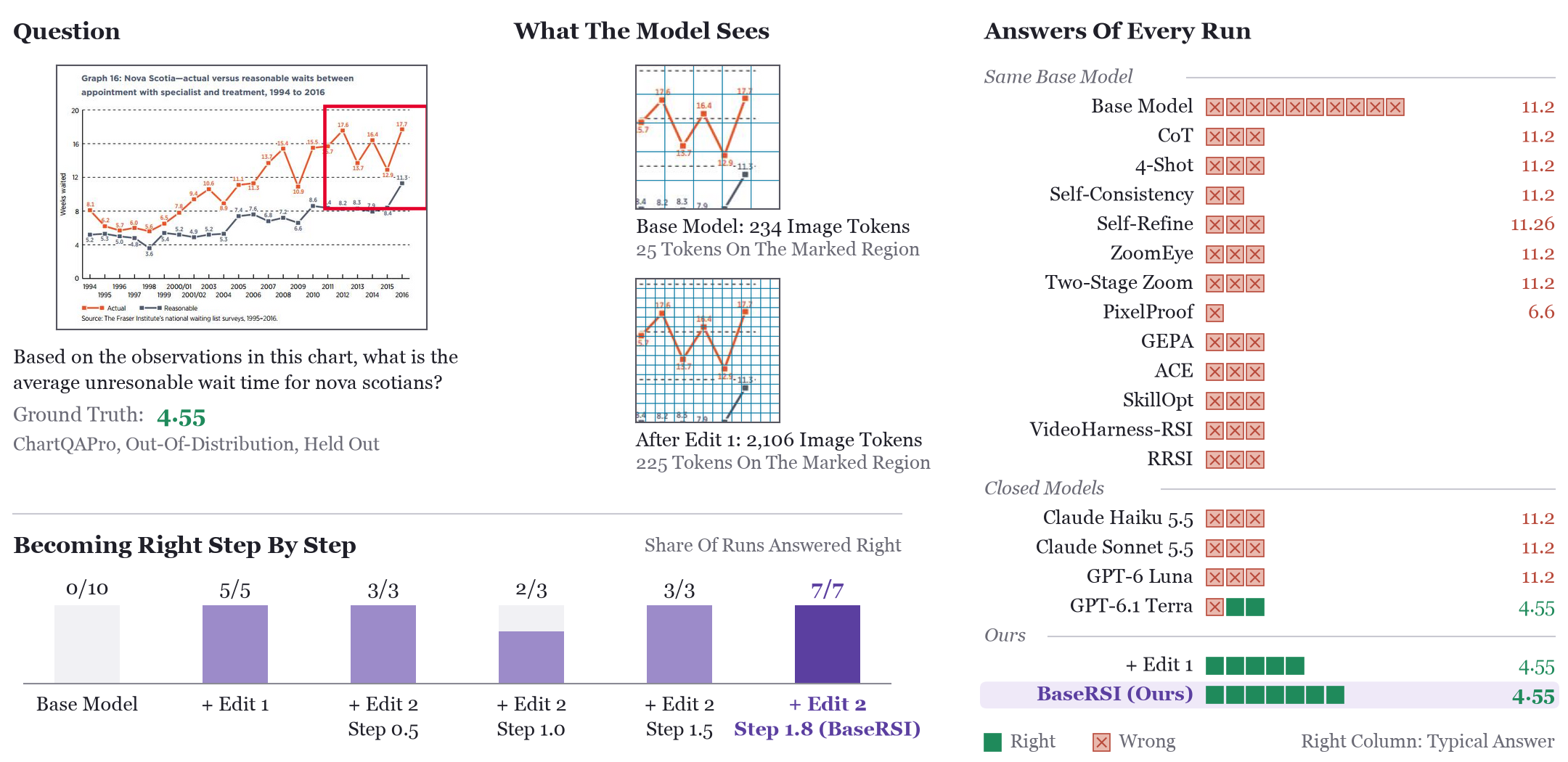}\\[-3pt]
\includegraphics[width=\textwidth]{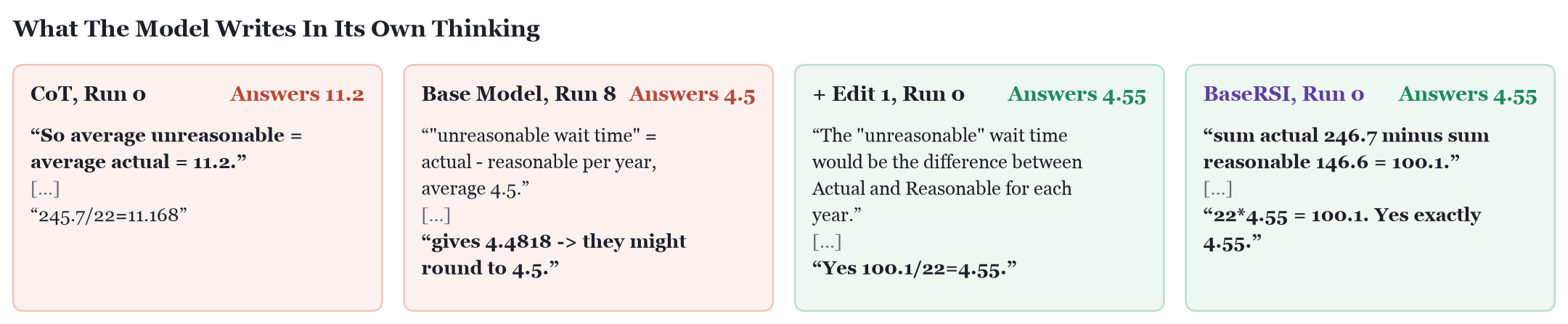}
\captionof{figure}{\textbf{Reading a chart} (ChartQAPro, held out). The base model and the other methods average the upper line; with the finer view the model averages the gap between the two lines, as the question asks, in all five runs.}
\label{fig:case_chart}
\par\vspace{3pt}
\begin{minipage}{\textwidth}
\begin{multicols}{2}
\textbf{Key observations.} The question asks for the average ``unreasonable'' wait time in a chart of the actual and the reasonable wait for each year, that is, the mean gap between its two lines, 4.55 weeks (Figure~\ref{fig:case_chart}).
Six of the ten runs of the base model answer about 11.2, the mean of the upper line, and the chain-of-thought baseline makes this reading explicit: ``average unreasonable = average actual = 11.2''.
The base model has 25 image tokens on the marked region of the labels, and the view edit 225 of 2,106; the edited model reads every label and writes that ``the `unreasonable' wait time would be the difference between Actual and Reasonable for each year'', reaching 4.55 in all five runs.
\name{} adds the two lines over the 22 years, ``sum actual 246.7 minus sum reasonable 146.6 = 100.1'', and checks ``22*4.55 = 100.1''.
All 31 runs of the other methods on the same base model are wrong, almost all with 11.2, and over seven effort levels of the closed-source models 3 of 21 runs are right, two of them from GPT-6.1 Terra, and 17 answer 11.2. The up-scaled image in the request is right in 4 of 5 runs, so here too the finer view decides how the question is read.
\par\vspace{3pt}{\small\parfillskip=0pt plus 0.4\linewidth\relax%
\emph{Base model, run 8:} ``The question asks for `average unresonable wait time' - likely meaning the actual wait time that exceeds reasonable, or perhaps they want the average of the `Actual' values? Hmm. [\dots] Let's list Actual values per year (1994-2016): 1994: 8.1, 1995: 6.2, 1996: 5.7, 1997: 6.0, 1998: 5.6, 1999: 5.5 [\dots]''
\emph{View edit, run 0:} ``The `unreasonable' wait time would be the difference between Actual and Reasonable for each year. [\dots] Actual (orange): 8.1, 6.2, 5.7, 6.0, 5.6, 6.5, 7.8, 9.4 [\dots] Reasonable (blue): 5.2, 5.3, 5.0, 4.8, 3.6, 5.4, 5.2, 4.9 [\dots] Alternatively, maybe they mean average of Actual minus average of Reasonable, which equals average of differences anyway. [\dots] Yes 100.1/22=4.55.''
\emph{\name{}, run 0:} ``The unreasonable wait time would be the difference between actual and reasonable for each year, then averaged. [\dots] Actual (red): 1994: 8.1, 1995: 6.2, 1996: 5.7 [\dots] Reasonable (blue): 1994: 5.2, 1995: 5.3, 1996: 5.0 [\dots] Now compute differences (actual - reasonable) for each: 1994: 8.1-5.2=2.9, 1995: 6.2-5.3=0.9, 1996: 5.7-5.0=0.7 [\dots]''\par}
\end{multicols}
\end{minipage}
\end{table*}

\end{document}